\documentclass[]{preprint}

\setcopyright{cc}
\acmDOI{10.1613/jair.1.19490}

\acmVolume{85}
\acmArticle{34}
\acmMonth{03}
\acmYear{2026}

\RequirePackage[
  datamodel=acmdatamodel,
  style=acmauthoryear,
  backend=biber,
  giveninits=true,
  uniquename=init
  ]{biblatex}

\usepackage{amsmath,amsfonts}
\usepackage{xcolor}
\usepackage{colortbl}
\usepackage{graphicx}
\usepackage{caption}
\usepackage{subcaption}
\usepackage{booktabs}
\usepackage{algorithm}
\usepackage{algpseudocode}
\usepackage{listings}
\usepackage{enumitem}
\usepackage{cleveref}
\usepackage{tikz}
\usetikzlibrary{trees, positioning, calc}
\usepackage[edges]{forest}
\usepackage{tabularx}
\usepackage{threeparttable}
\usepackage{longtable}
\usepackage{xspace}

\crefname{figure}{Figure}{Figures}
\Crefname{figure}{Figure}{Figures}
\crefname{section}{Section}{Sections}
\Crefname{section}{Section}{Sections}
\crefname{equation}{Eq.}{Eqs.}
\Crefname{equation}{Eq.}{Eqs.}
\crefname{appendix}{Appendix}{Appendix}
\Crefname{appendix}{Appendix}{Appendix}
\crefname{table}{Table}{Table}
\Crefname{table}{Table}{Table}

\newcommand{\numagents}{87\xspace} 
\newcommand{\numdatasets}{33\xspace} 

\newcommand{\uproman}[1]{\uppercase\expandafter{\romannumeral#1}}

\setlist[description]{
  itemsep=3pt,
  leftmargin=10pt,
  labelindent=10pt,
  font=\normalfont\bfseries, 
  labelsep=1ex, 
  style=sameline,
  itemindent=!,
}

\let\citet\textcite      
\let\citep\parencite     
\NewDocumentCommand{\citepeg}{m}{%
  \parencite[e.g.,][]{#1}%
}

\begin{document}

\title[Agents for Computer Use]{A Comprehensive Survey of Agents for Computer Use: Foundations, Challenges, and Future Directions}


\author{Pascal J. Sager}
\authornote{Corresponding author.}
\orcid{0000-0002-8084-2317}
\email{sage@zhaw.ch}
\affiliation{%
  \institution{Zurich University of Applied Sciences}
  \city{Winterthur}
  \country{Switzerland}
}
\affiliation{%
  \institution{University of Zurich}
  \city{Zurich}
  \country{Switzerland}
}
\affiliation{%
  \institution{ETH AI Center}
  \city{Zurich}
  \country{Switzerland}
}

\author{Benjamin Meyer}
\orcid{0009-0006-5609-2700}
\email{mebr@zhaw.ch}
\affiliation{%
  \institution{Zurich University of Applied Sciences}
  \city{Winterthur}
  \country{Switzerland}
}
\affiliation{%
  \institution{University of Zurich}
  \city{Zurich}
  \country{Switzerland}
}

\author{Peng Yan}
\orcid{0009-0006-0236-4707}
\email{yanp@zhaw.ch}
\affiliation{%
  \institution{Zurich University of Applied Sciences}
  \city{Winterthur}
  \country{Switzerland}
}
\affiliation{%
  \institution{University of Zurich}
  \city{Zurich}
  \country{Switzerland}
}

\author{Rebekka von Wartburg-Kottler}
\orcid{0009-0004-3506-444X}
\email{rebekka@kottler.ch}
\affiliation{%
  \institution{Zurich University of Applied Sciences}
  \city{Winterthur}
  \country{Switzerland}
}

\author{Layan Etaiwi}
\orcid{0000-0001-9250-7578}
\email{mashael.etaiwi@polymtl.ca}
\affiliation{%
  \institution{Polytechnique Montreal}
  \city{Montreal}
  \country{Canada}
}

\author{Aref Enayati}
\orcid{0009-0003-5313-1055}
\email{enay@zhaw.ch}
\affiliation{%
  \institution{Zurich University of Applied Sciences}
  \city{Winterthur}
  \country{Switzerland}
}
\affiliation{%
  \institution{University of Fribourg}
  \city{Fribourg}
  \country{Switzerland}
}

\author{Gabriel Nobel}
\orcid{0009-0002-9741-2521}
\email{nobel98@bluewin.ch}
\affiliation{%
  \institution{Zurich University of Applied Sciences}
  \city{Winterthur}
  \country{Switzerland}
}

\author{Ahmed Abdulkadir}
\orcid{0000-0003-4679-8081}
\email{abdk@zhaw.ch}
\affiliation{%
  \institution{Zurich University of Applied Sciences}
  \city{Winterthur}
  \country{Switzerland}
}

\author{Benjamin F. Grewe}
\orcid{0000-0001-8560-2120}
\email{bgrewe@ethz.ch}
\affiliation{%
  \institution{University of Zurich}
  \city{Zurich}
  \country{Switzerland}
}
\affiliation{%
  \institution{ETH AI Center}
  \city{Zurich}
  \country{Switzerland}
}
\affiliation{%
  \institution{AlpineAI AG}
  \city{Davos}
  \country{Switzerland}
}

\author{Thilo Stadelmann}
\orcid{0000-0002-3784-0420}
\email{stdm@zhaw.ch}
\affiliation{%
  \institution{Zurich University of Applied Sciences}
  \city{Winterthur}
  \country{Switzerland}
}
\affiliation{%
  \institution{European Centre for Living Technology}
  \city{Venice}
  \country{Italy}
}
\affiliation{%
  \institution{AlpineAI AG}
  \city{Davos}
  \country{Switzerland}
}

\renewcommand{\shortauthors}{Sager, Meyer, Yan, von Wartburg-Kottler, Etaiwi, Enayati, Nobel, Abdulkadir, Grewe \& Stadelmann}
\begin{abstract}
    {\bf Background:} 
    Agents for computer use (ACUs) are systems that execute complex tasks on digital devices -- such as personal computers or mobile phones -- given instructions in natural language. These agents automate tasks by controlling software through low-level actions like mouse clicks and touchscreen gestures. However, despite rapid progress, ACUs are not yet mature for everyday use.

    {\bf Objectives:}
    This survey examines the current state-of-the-art, identifies trends, and points out research gaps in the development of practical ACUs. The goal is to provide a comprehensive review and analysis that helps advance general-purpose, robust, and scalable agents for real-world computer use.

    {\bf Methods:}
    We introduce a multifaceted taxonomy of ACUs across three dimensions: (\uproman{1}) the \emph{domain perspective}, characterizing the contexts in which agents operate; (\uproman{2}) the \emph{interaction perspective}, describing observation modalities (e.g., screenshots, HTML) and action modalities (e.g., mouse, keyboard, code execution); and (\uproman{3}) the \emph{agent perspective}, detailing how agents perceive, reason, and learn. We review $\numagents$ original research papers about ACUs and $\numdatasets$ relevant datasets, covering both foundation model-based and specialized approaches.  
    
    {\bf Results:}
    Our taxonomy comprehensively structures state-of-the-art approaches and establishes the groundwork for guiding future ACU research. We found that the field is transitioning from specialized agents toward foundation-model-based agents, a shift from text to image-based observation space, and an increasing adoption of behavior cloning methodologies. Furthermore, we identify six key research gaps: insufficient generalization, inefficient learning, limited planning, low task complexity in benchmarks, non-standardized evaluation, and a disconnect between research and practical conditions.
    
    {\bf Conclusions:}
    To continue rapid improvements in the field, we recommend focusing on: (a) vision-based observations and low-level control to enhance generalization; (b) adaptive learning beyond static prompting; (c) effective planning and reasoning capabilities; (d) realistic, high-complexity benchmarks; (e) standardized evaluation criteria based on task success; and (f) aligning agent design with real-world deployment constraints. Collectively, our findings and proposed directions help develop more general-purpose agents for everyday digital tasks.
\end{abstract}


\received{11 June 2025}
\received[accepted]{24 February 2026}

\maketitle


\section{Introduction}\label{intro}
\begin{figure}[t]
\begin{subfigure}{0.49\linewidth}
\includegraphics[width=\linewidth]{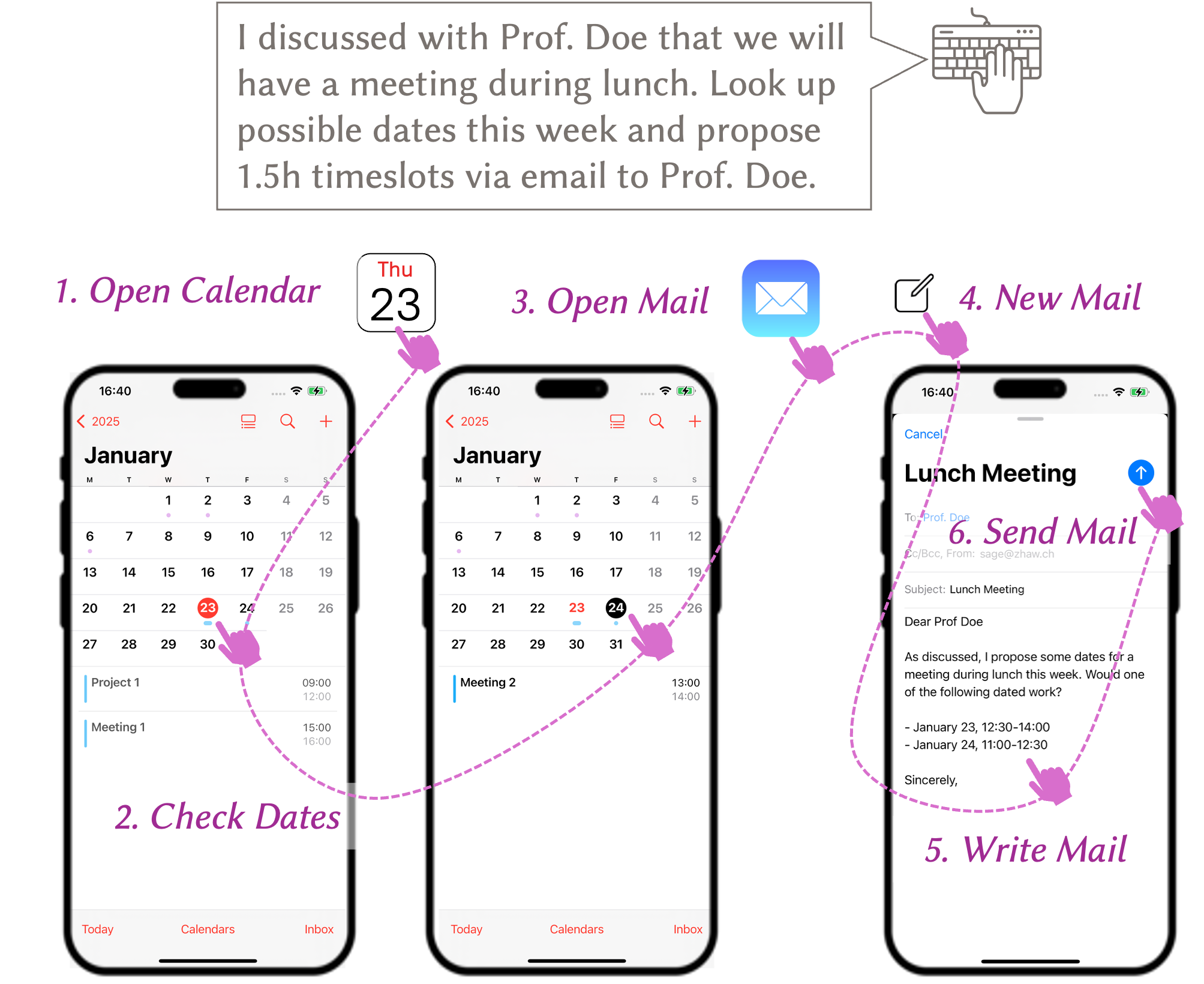}
\caption{Example of a computer use task.}
\label{fig:ACU-example}
\end{subfigure}
\hfill
\begin{subfigure}{0.49\linewidth}
\includegraphics[width=\linewidth]{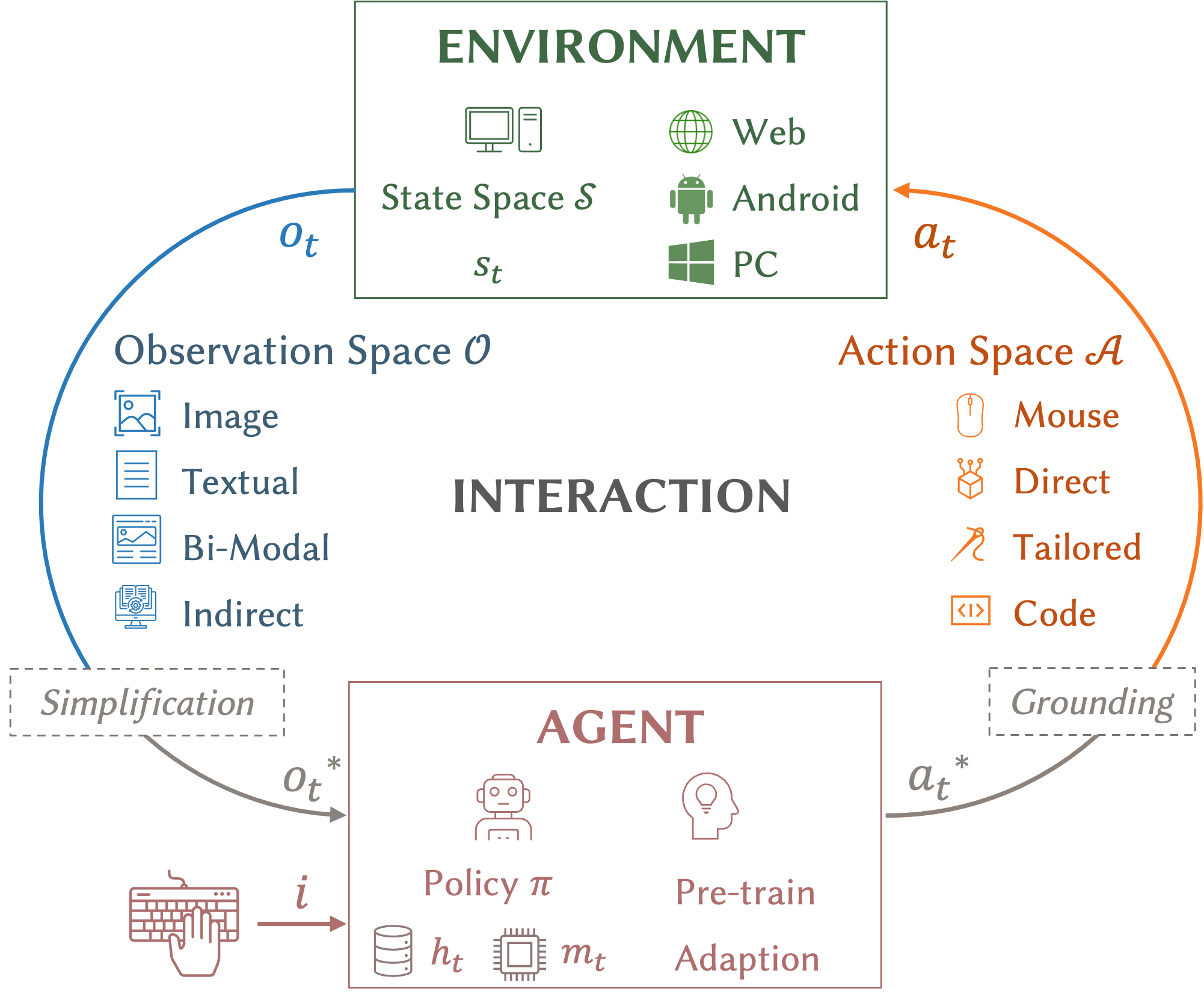}
\caption{Proposed taxonomy.}
\label{fig:ACU-taxonomy}
\end{subfigure}
\caption{
Overview: (a) An example of a task for an ACU: A user specifies a task (``propose meeting dates by email'') and the agent executes it.
(b) We structure the literature on ACUs based on three key \emph{perspectives} corresponding to the main differentiating aspects:
(1) The shared \emph{domain} properties across computer environments (e.g., Web, Android).
(2) The means of \emph{interaction} between the agent and the environment as manifested in the observation and action spaces.
(3) The \emph{agent} components: how an agent acts through a policy $\pi$ while tracking the past in memory and how an agent learns to act.}
\Description{Example of a typical task and conceptual taxonomy of ACU systems. In the left subfigure, a user instruction is shown in a text bubble at the top of the image: ``Look up possible dates this week and propose 1.5h timeslots via email to Prof. Doe.'' Below, a sequence of six steps is illustrated using three smartphone screenshots connected by arrows. The steps include opening a calendar app, checking available dates, switching to the email app, composing a message with proposed times, and sending the email. This visual sequence highlights the agent's execution of high-level user intent via low-level app interactions. In the right subfigure, a conceptual taxonomy diagram depicts the structure of ACU systems. At the top, the computer environment includes platforms such as Web, Android, and PC, along with their corresponding state spaces. Observation flows from environment to agent and is categorized into image-based, textual, bi-modal, and indirect observations. Action flows from agent to environment and is categorized into mouse interactions, direct input, tailored mechanisms, and code execution.}
\label{fig:intro}
\end{figure}

AI agents operate by perceiving their environment and selecting actions to achieve predefined goals \citep{mnih_playing_2013}.
This agent-based paradigm, popularized in the 1990s \citep{schmidhuber1990line, 10.1145/122344.122377, russell_artificial_2022}, has shown success across domains such as robotic control \citep{yang_data_2020}, game playing \citep{baker_video_2022}, and autonomous driving \citep{grigorescu_survey_2020}.
A growing class of agents extends this paradigm by allowing users to define goals in natural language \citep{ouyang_training_2022}. These instruction-based agents interpret textual instructions and autonomously act in complex environments to fulfill them.

One promising application of instruction-based agents is computer use, where agents control software through computer interfaces originally designed for humans. These agents automate tasks such as scheduling, browsing, or document editing by interacting with digital platforms via simulated inputs such as mouse clicks or touchscreen gestures.
For instance, a user could instruct an agent embedded in a smartphone to propose meeting dates and send them via email. The agent would then operate the phone through simulated touch actions to fulfill the request, as illustrated in \cref{fig:ACU-example}.
We refer to this class of agents as \textbf{agents for computer use (ACUs)}.

\begin{figure}
    \centering
    \includegraphics[width=0.6\linewidth]{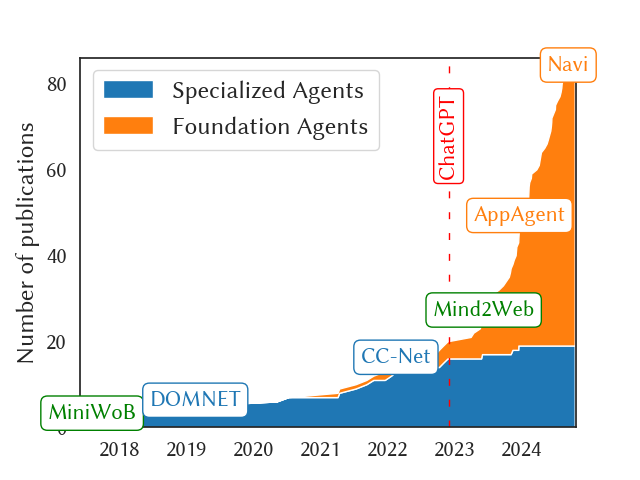}
    \caption{ACU publications over time. Boxes highlight seminal milestones. The advent of ChatGPT marks a shift from RL-focused agents to those primarily relying on foundation model reasoning.}
    \Description{
    Area chart showing the absolute number of publications per year from 2018 through 2024, separated by agent type. Differently colored areas distinguish specialized agents from foundation-model-based agents, with a legend identifying each. Key milestones, such as influential papers and dataset releases, are highlighted using boxed annotations along the timeline. A vertical red dashed line marks the launch of ChatGPT in late 2022, after which the number of publications on foundation-model-based agents rises sharply  (from about $20$ to over $80$ publications), overtaking specialized agents. The transition indicates a field-wide shift in research emphasis.}
    \label{fig:timeline-new}
\end{figure}

Early research on ACUs focused primarily on the learning methodology, particularly reinforcement learning (RL) techniques \citepeg{branavan_reinforcement_2009,jia_dom-q-net_2019,humphreys_data-driven_2022}.
Recently, a shift toward integrating foundation models, such as large language models (LLMs) and vision-language models (VLMs) (see \cref{sec:conceptional_view}), has accelerated progress, significantly enhancing reasoning capabilities and enabling ACUs to tackle increasingly complex tasks \citep{wei_emergent_2022,kim_language_2023}.
This transition has stimulated research activity, reflected in a strong increase in publications in the field (see \cref{fig:timeline-new}).

Concurrently, commercial prototypes of instruction-based agents for computer use have begun to emerge \citepeg{anthropic2024computeruse, google2024project, venturebeat2025openai}. However, despite this momentum, ACUs remain limited in their generalization, robustness, and planning abilities, achieving almost six time lower success rates than humans due to the inability of handling dynamic UI changes, switching between multiple software applications, or errors in dense UI environments such as spreadsheets \citep{xie_osworld_2024}.
To assess both the opportunities and barriers, we conduct a broad survey of ACUs across domains, learning strategies, and modalities. This survey complements the existing body of literature with a comprehensive taxonomy grounded in established intelligent agent theory \citep{russell_artificial_2022,sutton_reinforcement_2018}, enabling a holistic and technology-agnostic analysis of the ACU landscape (see \cref{fig:ACU-taxonomy} for an overview).

Applying this taxonomical framework to current research, we identify several critical research gaps.
First, many agents rely heavily on structurally inconsistent observations (e.g., unrealistically sanitized HTML), limiting their ability to generalize and scale to real-world applications.
Second, current learning strategies are costly and inefficient, often relying on simulations, requiring substantial labeled data, or struggling to adapt to specific environments.
Third, agents have a very limited ability to plan and execute complex multi-step tasks reliably.
Fourth, existing benchmarks focus on real-world perception but lack sufficient task complexity to fully assess agent capabilities.
Fifth, non-standard evaluation practices hinder meaningful comparison across different publications.
Sixth, a mismatch exists between the assumptions regarding ACUs and their operational environments made during research and the actual conditions encountered during real-world deployment.

To address these challenges, we propose several directions.
Observing and acting on uniform visual inputs provides more robust generalization than observing inconsistent textual inputs, e.g., HTML content.
To overcome inefficiencies in learning, we highlight the need for cost-effective strategies and discuss promising directions for more scalable adaptation.
To improve planning capabilities, we suggest exploring advances in reasoning models and integrating robust planning algorithms.
To close the gap in benchmarking ACUs, we advocate for the development of datasets that better capture task complexity alongside real-world perception.
To enable meaningful evaluation, we assert that the task success rate following standardized measurement practices should become the norm for comparing agent capabilities.
Finally, to bridge the gap between research assumptions and real-world conditions, we identify several key discrepancies and propose research directions to address them.

This survey aims to foster advancements in the field of ACUs by providing a principled and comprehensive overview of the domain. Specifically, our contributions are as follows: (1) The introduction of a comprehensive taxonomy for ACUs, (2) the classification of $\numagents$ ACUs and $\numdatasets$ datasets within this framework, (3) the identification of six critical research gaps, and (4) the proposal of strategic directions to address these challenges.

\subsection{Relation to Other Surveys}
In contrast to existing surveys, our review examines ACUs from a technology-agnostic perspective and introduces a unifying framework that bridges diverse domains, methodologies, and technologies.
This broader scope allows us to introduce a novel, unifying taxonomy for ACUs that is compatible across a wide range of agent types ---something previous work could not realize due to their limited scope. Specifically, existing surveys have the following limitations:

\begin{description}
    \item[Limitation in learning strategies:] \citet{zhang_large_2024} and \citet{wang_gui_2024} focus only on computer use for foundation-model-based agents, without discussing other learning frameworks such as pure reinforcement learning as the core principle of design. In contrast, 
    other surveys \citepeg{arulkumaran_deep_2017, moerland_model-based_2023} focus only on general reinforcement learning-based agents.

    \item[Limited scope within computer use:] 
    \citet{wu_foundations_2024} discuss only mobile agents, neglecting other computer domains.
    While their survey provides a comprehensive review of some aspects of computer use, it focuses on a specific subpart of the field. To discuss future research directions comprehensively, it is important to analyze the field as a whole.
    \item[Lack of computer use specificity:]
    Other reviews \citepeg{arulkumaran_deep_2017, wang_survey_2024} provide a comprehensive overview of agents based on a specific technology but do not focus on the domain of computer use and all its intricacies.
    \item[Adjacent research areas with limited relevance for agentic computer use:]
    Another set of surveys \citepeg{yu_vision-based_2023, li_gui_2023} concentrate on related topics, such as GUI testing, but do not cover agent-based interactions.
    Other reviews \citepeg{syed_robotic_2020, asatiani_robotic_2020} focus on robotic process automation using scripted software robots (also called agents) to automate predefined workflows.
\end{description}

\citet{gao_generalist_2024} provide a valuable overview with a similar scope. In contrast, we go considerably deeper into key aspects, offering a more comprehensive analysis and novel insights, based on a taxonomy built upon existing intelligent agent theory \citep{russell_artificial_2022,sutton_reinforcement_2018} that is useful to find related work down to intricate questions of agent design. We posit that a comprehensive and in-depth review of the field is essential to systematically uncover the limitations of current agents and identify how diverse technologies and methodologies can inform one another to advance the state of the art.

\subsection{Survey Methodology}
\begin{figure}[ht]
    \centering
    \begin{tikzpicture}[
        node distance=0.8cm and 0.8cm,
        font=\footnotesize,
        process/.style={rectangle, minimum width=2.2cm, minimum height=1cm, text centered, draw=black, fill=blue!5, rounded corners},
        criteria/.style={rectangle, minimum width=2.2cm, minimum height=0.8cm, text centered, draw=black, dashed, fill=yellow!5, rounded corners, font=\scriptsize},
        finalnode/.style={trapezium, shape border rotate=270, trapezium angle=60, minimum height=1.8cm, minimum width=1.1cm, text centered, draw=black, fill=green!5, inner sep=1pt, align=center},
        arrow/.style={thick, ->, >=stealth}
    ]
    
    \node (sources) [process, text width=2.5cm] {Initial collection};
    
    \node (screen) [process, right=of sources, text width=2.5cm] {Publication selection};
    
    \node (criteria) [criteria, above=0.6cm of screen, fill=white, text width=2.5cm] {Selection Criteria}; 
    
    \node (snowball) [process, below=0.6cm of screen, text width=2.5cm] {Snowballing};
    
    \node (final) [finalnode, right=0.6cm of screen] {Final\\Set}; 
    
    \node (agents) [process, right=0.8cm of final, yshift=0.6cm, minimum width=2.2cm, fill=orange!5, minimum height=0.7cm] {Agents ($N{=}\numagents$)};
    \node (data) [process, right=0.8cm of final, yshift=-0.6cm, minimum width=2.2cm, fill=orange!5, minimum height=0.7cm] {Datasets ($N{=}\numdatasets$)};
    
    \node (synthesis) [process, right=0.5cm of agents, yshift=-0.6cm, text width=2.5cm, minimum height=1.9cm]
    {Data Analysis and Synthesis};
    
    \draw [arrow] (sources) -- (screen);
    \draw [arrow] (criteria) -- (screen); 
    \draw [arrow] (screen) -- (final);
    
    \draw [arrow] (screen.south) -- (snowball.north);
    \draw [arrow] (snowball.west) -- ++(-0.3,0) |- (screen.west);
    
    \draw [arrow] (final.east |- agents.west) |- (agents.west);
    \draw [arrow] (final.east |- data.west) |- (data.west);
    
    \draw [arrow] (agents.east) -- (synthesis.west |- agents.east);
    \draw [arrow] (data.east) -- (synthesis.west |- data.east);

    \end{tikzpicture}
    \caption{Visual overview of the survey methodology phases, detailing the selection logic and synthesis process.}
    \Description{
    A horizontal flowchart illustrating the four-phase survey methodology. The process flows from left to right. It begins with Phase 1: Initial Collection, which leads into Phase 2: Publication Selection. Phase 2 inputs include specific Selection Criteria (Deep Learning, Computer Use, Common Applications) and an iterative feedback loop with Phase 3: Snowballing. The output of Phase 2 is a Final Set of papers (N papers). This final set branches into two categories: Agents (N agents) and Datasets (N datasets). Both categories feed into the final step on the right, Phase 4: Data Analysis and Synthesis.
    }
    \label{fig:methodology}
\end{figure}

The field of ACUs is fragmented and lacks a unified terminology, making a classic systematic review infeasible. We therefore employed an iterative, semi-structured collection process combining expert knowledge with snowball sampling techniques. As visualized in \cref{fig:methodology}, our methodology comprised the following phases:

\begin{description}
    \item[Initial collection:] Based on domain expertise and exploratory keyword-based searches, we compiled an initial set of candidate publications. Given the rapid evolution of this domain, we included both peer-reviewed and preprint works to reflect the state-of-the-art. 
    We conducted a semi-structured search using Google Scholar and Semantic Scholar, combining keywords such as ``AI agent'', ``LLM/LVM agent'', and ``computer use''. We only collect the papers from the last $7$ years (2018-2024).
    \item[Publication selection:] After initial collection, we filtered the literature and included publications after carefully reviewing their titles, abstracts, and additional parts of their content for fit, using the criteria catalog described below. The selection was conducted by two researchers independently, followed by a consensus discussion. Since some publications are preprints, we evaluated them by using domain knowledge and checked their consistency with emerging trends to mitigate quality concerns.
    \item[Snowballing:] We used backward snowballing for adding references and forward snowballing for determining saturation: For each selected publication, we analyzed its references (backward), and for randomly selected works, we analyzed citations (forward) and checked whether additional relevant works could be identified. We iteratively repeated this process until no additional relevant papers emerged, ensuring thematic saturation.
    \item[Data Analysis and Synthesis:] To develop the taxonomy, we employed an inductive approach. We extracted key attributes from the final set of papers (e.g., input modalities, action spaces, evaluation metrics) and grouped them into high-level dimensions. This process was iterative; as new properties emerged during data extraction, the taxonomic dimensions were refined to ensure they covered both agent architectures and dataset characteristics comprehensively.
\end{description}

The selection criteria for the collection process for both agent and dataset papers are defined as follows:

\begin{description}
    \item[Deep learning focus:]
        We only included agents utilizing deep learning for computer use, excluding traditional rule-based systems.
    \item[Computer use focus:]
        We distinguish between passive advice and active execution. We exclude instruction-based agents (chatbots) that access external tools but only provide text advice or instructions for a human to follow \citepeg{yang_gpt4tools_2023,tang_toolalpaca_2023,li_towards_2024,guo_stabletoolbench_2024,qin_toolllm_2024}. 
        In contrast, we include agents that generate executable code (e.g., Python/Selenium scripts). Although these agents output text (code), this code acts as a dynamic action space that is executed by an interpreter to directly manipulate the interface, thereby satisfying our definition of autonomous computer use.
    \item[Common computer applications focus:]
        We exclude agents playing video games \citepeg{baker_video_2022, zhu_ghost_2023}, controlling server facilities \citepeg{ran_deepee_2019, fulpagare_optimal_2022}, coding agents \citepeg{ross_programmers_2023, qian_chatdev_2024} or software testing \citepeg{koroglu_qbe_2018, degott_learning_2019, pan_reinforcement_2020}.
        We only include datasets that provide instructions and require agents to fulfill these instructions through computer interactions.
\end{description}

The final set of publications was curated through a multi-stage screening process and agreement through team discussion to ensure comprehensive coverage of the field. Our selection criteria are prioritized by scholarly impact (high citations or benchmark leadership), representation of key application domains (e.g., Web, Android), and diverse coverage across taxonomic dimensions outlined in \Cref{fig:framework}.

From the identified publications, we classify $\numagents$ as ACU agents and $\numdatasets$ as computer use datasets. We deliberately separate these two categories in our analysis because they serve distinct roles in the ecosystem: agents represent the \textit{methodological solutions} and architectures, while datasets provide the \textit{evaluation environments} and benchmarks. Analyzing them separately allows us to derive specific taxonomies for architectural properties (e.g., learning paradigms, see Appendix \ref{appendix_A1}--\ref{appendix_A2}) versus environmental properties (e.g., observation spaces, see Appendix \ref{appendix_A3}).

Nevertheless, we note several limitations: (i) In our review, about 1/3 of the cited works are preprints. While they reflect emerging trends, they may lack validation and introduce concerns about quality. Readers should interpret these works cautiously; (ii) keyword search and manual screening, even when combined with snowballing, might have overlooked relevant papers; (iii) the focus on deep learning excludes traditional machine learning and rule-based approaches; and (iv) selection involves subjective judgment. To mitigate these limitations, we took several steps. For the inclusion of preprints, we cross-referenced findings with peer-reviewed literature when possible. To address potential gaps from keyword search and manual screening, we iteratively refined our search strategy and complemented it with citation tracking. Although our focus excluded traditional machine learning and rule-based approaches, we clearly defined this scope upfront to maintain consistency. Finally, to reduce subjective bias in selection, we employed a consensus-based review process, where disagreements were discussed and resolved collectively by multiple reviewers.

\subsection{Survey Structure}

Due to the developing nature of this field, individual ACUs that stand for important strands do not yet stand out; rather, many agents only employ certain aspects of what contributes to the full picture of ACUs. Hence, most subsequent chapters of this review put individual elements of the taxonomy at the center rather than individual ACUs, giving representative exemplary or specific ACUs as references for each aspect. A notable exception will be \cref{sec:learning_strategy}, where individual agents are most prominently portrayed, as it discusses their core development paradigm. Otherwise, a structuring of the field by agents can be found in the tables in the Appendix~\ref{sec:appendix_tables}. 

The survey is structured as follows: In \cref{sec:field}, we formalize the problem of agents for computer use and introduce respective terminology as a precursor to introducing the perspectives of the proposed taxonomy. Then, we look into each perspective in detail in the three subsequent chapters:
In \cref{sec:domain_view}, we discuss the composition of commonly used domains (\emph{domain perspective});
in \cref{sec:interaction_view}, we analyze the interaction between the agent and the environment through the observation and action space (\emph{interaction perspective}); 
in \cref{sec:conceptional_view}, we dissect the components of an agent, how an agent acts, and how an agent learns to act (\emph{agent perspective}). 
Then, in \cref{sec:datasets}, we summarize existing datasets used to train or evaluate agents, and we examine metrics and methodologies used to evaluate an agent's performance in \cref{sec:evaluation_agent}.
Finally, we conclude by summarizing our findings and providing directions for future research in \cref{sec:discussion_conclusion}.

To complement the main text, the Appendix provides several in-depth analyses:
In \cref{sec:appendix_trends}, we examine trends and distributions in the literature, including common choices of observation modalities, action spaces, and learning strategies;
\Cref{sec:img_vs_text,sec:code_generation_examples} contrast different observations and provide examples of code-based actions;
\Cref{sec:domain_view:nature,sec:considerations_production} discuss challenges in deploying ACUs, such as mismatches between environment properties assumed in research and found in real-world settings;
\Cref{sec:appendix_tables} presents a structured overview of existing agents and datasets, classifying them according to our proposed taxonomy.

\section{The Field of ACUs}
\label{sec:field}

This section formalizes ACUs and outlines our taxonomy perspectives.

\subsection{Definitions}\label{sec:problem_statement}
In the following, we describe ACUs using well-established intelligent agent notation \citep{russell_artificial_2022,sutton_reinforcement_2018} to provide a consistent basis for discussion in the upcoming sections. Human users interact with ACUs by issuing a text-based instruction $i$, which the agent must fulfill through actions in a computer environment. To illustrate this interaction model, \cref{fig:ACU-taxonomy} visualizes the key components of an ACU and its interface with the environment, including how it perceives observations and selects actions.

At each time $t$, the computer environment is in a state $s_t \in \mathcal{S}$. 
The ACU receives only a partial view, called an \emph{observation} $o_t \in \mathcal{O}$. 
$\mathcal{S}$ denotes the state and $\mathcal{O}$ the observation space, respectively.
For example, $o_t$ could be a screenshot of the current screen, only showing the foreground application, whereas $s_t$ would encompass all running computer processes.
Based on $o_t$ and instruction $i$, the ACU selects an \emph{action} $a_t \in \mathcal{A}$ (action space), such as a mouse click, keypress, or a higher-level command \citepeg{shi_world_2017, wang_officebench_2024}.

In practice, ACUs often \emph{simplify} observations, denoted $o_t \rightarrow o_t^*$,  to reduce complexity by, for example, downscaling or cropping UI screenshots \citepeg{chen_webvln_2024}.
Besides using simplified observation, ACUs can also predict abstract actions.
Such actions must be converted in a \emph{grounding} process $a_t^* \rightarrow a_t$ from abstract actions $a_t^*$ into executable actions $a_t \in \mathcal{A}$.
Grounding is typically applied when a large language model (LLM) is used for planning, requiring the agent to convert high-level descriptions such as \texttt{click submit button} into executable commands such as \texttt{click(x,y)}, where \texttt{x} and \texttt{y} are screen coordinates of the submit button \citepeg{gao_assistgui_2024}.

The ACU’s behavior is defined by a (typically stochastic) policy $\pi$. In its simplest form, the policy determines the action solely based on the current observation $o_t$ and the instruction $i$:
\begin{equation}
a_t \sim \pi(\ \cdot\ |\ o_t,\ i\ ) \label{eq:simple-policy}
\end{equation}
Interacting over several steps yields a trajectory $\tau = \bigl((o_0,a_0), (o_1, a_1), ... \bigr)$, ending when the instruction is fulfilled or a step limit is reached. 
Effective computer control often requires remembering previous observations $(o_0, ..., o_{t-1})$, making adding a memory component to the policy essential.

\subsection{A Comprehensive Taxonomy}\label{sec:agent_dimensions}

\definecolor{domain_color}{RGB}{144,193,157} 
\definecolor{domain_color_bg}{RGB}{214,239,220} 

\definecolor{observation_space_color}{RGB}{110,164,220} 
\definecolor{observation_space_color_bg}{RGB}{219,235,250} 

\definecolor{action_space_color}{RGB}{239,156,91} 
\definecolor{action_space_color_bg}{RGB}{254,230,209} 

\definecolor{learning_strategy_color}{RGB}{200,90,120} 
\definecolor{learning_strategy_color_bg}{RGB}{245,210,220} 
\definecolor{learning_strategy_color_General_Pre_training}{RGB}{170,70,130}
\definecolor{learning_strategy_color_General_Pre_training_bg}{RGB}{235,210,230}
\definecolor{learning_strategy_color_Environment_Adaption}{RGB}{190,65,115}
\definecolor{learning_strategy_color_Environment_Adaption_bg}{RGB}{240,200,220}
\definecolor{learning_strategy_color_Episodic_improvement}{RGB}{210,100,130}
\definecolor{learning_strategy_color_Episodic_improvement_bg}{RGB}{250,225,230}

\definecolor{policy_color}{RGB}{220,120,110}
\definecolor{policy_color_bg}{RGB}{248,220,215}

\definecolor{agent_color_Reinforce_Agent}{RGB}{125,175,210}
\definecolor{agent_color_Reinforce_Agent_bg}{RGB}{210,230,245}
\definecolor{agent_color_Foundation_Agent}{RGB}{250,170,100}
\definecolor{agent_color_Foundation_Agent_bg}{RGB}{255,230,200}
\definecolor{agent_color}{RGB}{230,200,200}
\definecolor{agent_color_bg}{RGB}{245,235,235}

\definecolor{interaction_color}{RGB}{220,220,220}
\definecolor{interaction_color_bg}{RGB}{245,245,245}

\begin{figure}[p]
    \begin{forest}
        forked edges,
        for tree={
            grow=east, 
            draw, 
            rounded corners, 
            node options={align=center, font=\scriptsize}, 
            edge={thick}, 
            s sep=1mm, 
            l sep=4mm, 
            minimum width=2.5cm, 
            anchor=west,
            fork sep=2mm,
            inner sep=2pt,
            parent anchor=east,  
            child anchor=west     
        },
        [Domain Persp. \\ \textit{(\Cref{sec:domain_view}, p.~\pageref{sec:domain_view})}, fill=domain_color_bg
            [Android \\ \textit{(\Cref{sec:domain_view}, p.~\pageref{sec:domain_view})}, fill=domain_color]
            [Web \\ \textit{(\Cref{sec:domain_view}, p.~\pageref{sec:domain_view})}, fill=domain_color]
            [Personal Computer \\ \textit{(\Cref{sec:domain_view}, p.~\pageref{sec:domain_view})}, fill=domain_color]
        ]
    \end{forest}
    \par\vskip 4mm
    \begin{forest}
        forked edges,
        for tree={
            grow=east, 
            draw, 
            rounded corners, 
            node options={align=center, font=\scriptsize}, 
            edge={thick}, 
            s sep=1mm, 
            l sep=4mm, 
            minimum width=2.5cm, 
            anchor=west,
            fork sep=2mm,
            inner sep=2pt,
            parent anchor=east,  
            child anchor=west     
        },
        [Interaction Perspective \\ \textit{(\Cref{sec:interaction_view}, p.~\pageref{sec:interaction_view})}, fill=interaction_color_bg
            [Observation Spaces \\ \textit{(\Cref{sec:obervation_space}, p.~\pageref{sec:obervation_space})}, fill=observation_space_color_bg
                [Image \\ \textit{(\Cref{sec:obervation_space_image}, p.~\pageref{sec:obervation_space_image})}, fill=observation_space_color_bg]
                [Text \\ \textit{(\Cref{sec:obervation_space_image_text}, p.~\pageref{sec:obervation_space_image_text})}, fill=observation_space_color_bg]
                [Bi-Modal \\ \textit{(\Cref{sec:obervation_space_bimodal}, p.~\pageref{sec:obervation_space_bimodal})}, fill=observation_space_color_bg]
                [Indirect \\ \textit{(\Cref{sec:obervation_space_indirect}, p.~\pageref{sec:obervation_space_indirect})}, fill=observation_space_color_bg]
            ]
            [Action Spaces \\ \textit{(\Cref{sec:action_space}, p.~\pageref{sec:action_space})}, fill=action_space_color_bg
                [Mouse \& Keyboard \\ \textit{(\Cref{sec:action_space_mouse}, p.~\pageref{sec:action_space_mouse})}, fill=action_space_color_bg]
                [Direct UI Access \\ \textit{(\Cref{sec:action_space_direct}, p.~\pageref{sec:action_space_direct})}, fill=action_space_color_bg]
                [Task-Tailored Actions \\ \textit{(\Cref{sec:action_space_task}, p.~\pageref{sec:action_space_task})}, fill=action_space_color_bg]
                [Executable Code \\ \textit{(\Cref{sec:action_space_code}, p.~\pageref{sec:action_space_code})}, fill=action_space_color_bg]
                [Action Grounding \\ \textit{(\Cref{sec:action_space_grounding}, p.~\pageref{sec:action_space_grounding})}, fill=action_space_color_bg]
            ]
        ]
    \end{forest}
    \par\vskip 4mm
    \begin{forest}
        forked edges,
        for tree={
            grow=east, 
            draw, 
            rounded corners, 
            node options={align=center, font=\scriptsize}, 
            edge={thick}, 
            s sep=1mm, 
            l sep=4mm, 
            minimum width=2.5cm, 
            anchor=west,
            fork sep=2mm,
            inner sep=2pt,
            parent anchor=east,  
            child anchor=west     
        },
        [Agent Perspective \\ \textit{(\Cref{sec:conceptional_view}, p.~\pageref{sec:conceptional_view})}, fill=agent_color_bg
                [Foundation Agents \\ \textit{(\Cref{sec:foundation_agent}, p.~\pageref{sec:foundation_agent})}, fill=agent_color_bg]
                [Specialized Agents \\ \textit{(\Cref{sec:reinforce_agent}, p.~\pageref{sec:reinforce_agent})}, fill=agent_color_bg]
                [Agent Policy \\ \textit{(\Cref{sec:agent_policy}, p.~\pageref{sec:agent_policy})}, fill=policy_color_bg
                    [Memoryless \\ \textit{(\Cref{sec:agent_policy_memoryless}, p.~\pageref{sec:agent_policy_memoryless})}, fill=policy_color_bg]
                    [History-Based \\ \textit{(\Cref{sec:agent_policy_history}, p.~\pageref{sec:agent_policy_history})}, fill=policy_color_bg]
                    [State-Based \\ \textit{(\Cref{sec:agent_policy_state}, p.~\pageref{sec:agent_policy_state})}, fill=policy_color_bg]
                    [Mixed \\ \textit{(\Cref{sec:agent_policy_mixed}, p.~\pageref{sec:agent_policy_mixed})}, fill=policy_color_bg]
                ]
                [Learning Strategy \\ \textit{(\Cref{sec:learning_strategy}, p.~\pageref{sec:learning_strategy})}, fill=learning_strategy_color_bg
                    [General Pre-Training \\ \textit{(\Cref{sec:learning_strategy_pretraining}, p.~\pageref{sec:learning_strategy_pretraining})}, fill=learning_strategy_color_General_Pre_training_bg]
                    [Environment Learning \\ \textit{(\Cref{sec:environment_adaption}, p.~\pageref{sec:environment_adaption})}, fill=learning_strategy_color_Environment_Adaption_bg
                        [Reinforc. Learning \\ \textit{(\Cref{sec:environment_adaption_rl}, p.~\pageref{sec:environment_adaption_rl})}, fill=learning_strategy_color_Environment_Adaption_bg]
                        [Behavioral Cloning \\ \textit{(\Cref{sec:environment_adaption_bc}, p.~\pageref{sec:environment_adaption_bc})}, fill=learning_strategy_color_Environment_Adaption_bg]
                        [Long-Term Memory \\ \textit{(\Cref{sec:environment_adaption_ltm}, p.~\pageref{sec:environment_adaption_ltm})}, fill=learning_strategy_color_Environment_Adaption_bg]
                    ]
                    [Episodic Improvement \\ \textit{(\Cref{sec:episodic_improvement}, p.~\pageref{sec:episodic_improvement})}, fill=learning_strategy_color_Episodic_improvement_bg
                        [Instruction Tuning \\ \textit{(\Cref{sec:episodic_improvement_instruction}, p.~\pageref{sec:episodic_improvement_instruction})}, fill=learning_strategy_color_Episodic_improvement_bg]
                        [Demonstrations \\ \textit{(\Cref{sec:episodic_improvement_demostrations}, p.~\pageref{sec:episodic_improvement_demostrations})}, fill=learning_strategy_color_Episodic_improvement_bg]
                        [Planning \\ \textit{(\Cref{sec:episodic_planning}, p.~\pageref{sec:episodic_planning})}, fill=learning_strategy_color_Episodic_improvement_bg]
                    ]
                ]
            ]
    \end{forest}
    \caption{The taxonomy of instruction-based ACUs is structured by three main perspectives and their components. The respective colors will be used throughout this paper to help easily associate content with each component.}
    \Description{
    Taxonomy overview of instruction-based agents for computer use (ACUs) structured by three perspectives. The figure presents a hierarchical taxonomy that organizes ACUs from three complementary perspectives: domain, interaction, and agent. The domain perspective categorizes ACUs by target environment, including Android, web, and personal computers. The interaction perspective decomposes the agent–environment interface into observation and action spaces. Observation spaces include modalities such as image, text, bi-modal, and indirect signals. Action spaces encompass various control modalities, including mouse and keyboard input, direct user interface access, task-specific actions, executable code, and action grounding methods. The agent perspective outlines the internal components of ACUs. It distinguishes between foundation agents that rely on pre-trained foundation models and specialized agents with custom policies. It also breaks down agent policy types into memoryless, history-based, state-based, and mixed. Finally, it details learning strategies in terms of general pre-training, environment-specific learning (via reinforcement learning, behavioral cloning, and long-term memory), and episodic improvement (including instruction tuning, demonstrations, and planning). This taxonomy serves as a conceptual scaffold for the survey and introduces a novel organizational framework.
    }
    \label{fig:framework}
\end{figure}
\Cref{fig:framework} introduces our proposed taxonomy, which is organized around three complementary perspectives (each of which is explored in detail in the following sections).
The \emph{domain perspective} (\Cref{sec:domain_view}) focuses on the properties and interfaces of computer environments. It identifies commonalities in observation and action types across domains.
The \emph{interaction perspective} (agent $\leftrightarrow$ environment) (\Cref{sec:interaction_view}) describes how agents interact with their environments. It formalizes the observation space $\mathcal{O}$ and action space $\mathcal{A}$ used by ACUs and discusses techniques for simplifying observations and action grounding.
The \emph{agent perspective} (\Cref{sec:conceptional_view}) examines the internal structure of an ACU. We distinguish two main agent designs, identify three typical learning phases, and outline core components for acting, memory, and planning.

Our taxonomy builds on the agent-environment framework central to intelligent agent theory \citep{russell_artificial_2022,sutton_reinforcement_2018} and adopts well-established, domain-agnostic concepts, such as observation spaces and policies, whenever possible.
ACU-specific characteristics are categorized based on determining overarching patterns and concepts across the ACU literature.


\section{Domain Perspective}\label{sec:domain_view}
The most common domains in the literature are Web, Android, and personal computers.
Although these domains often overlap in terms of functionality, such as when users access web browsers on Android devices or manage emails via web interfaces on desktop computers, existing research typically distinguishes them based on their primary interaction environment.
Each domain presents a distinct interaction interface, such as HTML-based pages in the Web, touch screens in Android, and window-based GUIs in desktop environments.
To bridge these differences, we propose to group these interfaces into common types of observations and actions. 
These shared abstractions allow us to define a unifying perspective across domains, providing a foundation for transferable methods and cross-domain generalization.
We categorize domain-specific kinds of observations into the following types:
    \begin{description}
        \item[Image screen representation:] 
           Observations in the form of screenshots—either full screen, partial, or multi-view pixel images—common in all domains \citepeg{niu_screenagent_2024, song_visiontasker_2024, zhang_ufo_2024}.
        \item[Textual screen representation:] 
            Structured textual representations of the screen such as HTML markup in the Web domain, the Android view hierarchy, or the UI automation tree in Windows. They allow agents to interact with interfaces at a semantic level \citepeg{kim_language_2023, wen_autodroid_2024, zhang_ufo_2024}.
        \item[Indirect representation:] 
            Non-visual observations providing contextual or system-level information beyond what is currently rendered on screen, such as file system contents or network state \citepeg{song_restgpt_2023,wu_os-copilot_2024,guo_pptc_2024}.
    \end{description}
Representative examples of each observation type across domains are summarized in Table~\ref{tab:domain_os}.

\begin{table*}
    \caption{Our classification of \textbf{observation types} across the different domains, along with relevant examples for each domain.}
    \label{tab:domain_os}
    \begin{tabular}{>{\raggedright\arraybackslash}p{4cm} >{\raggedright\arraybackslash}p{3.4cm} >{\raggedright\arraybackslash}p{3.5cm} >{\raggedright\arraybackslash}p{3.4cm}}
    \toprule
    \footnotesize\textbf{Observation types} & \footnotesize \textbf{Web} & \footnotesize \textbf{Android} & \footnotesize \textbf{Personal computer} \\
    \midrule
    \footnotesize \textbf{Image screen representation} & {\footnotesize Website \citepeg{niu_screenagent_2024}, browser window \citep{zhou_webarena_2024}} & {\footnotesize Phone screen \citepeg{song_visiontasker_2024}} & {\footnotesize Foreground application \citep{zhang_ufo_2024}, computer screen} \\
    \footnotesize \textbf{Textual screen representation} & {\footnotesize HTML \citepeg{kim_language_2023}, accessibility tree \citep{zhou_webarena_2024}} & {\footnotesize Android view hierarchy \citepeg{wen_autodroid_2024}, accessibility tree \citepeg{li_uinav_2024}} & {\footnotesize UI automation tree \citepeg{zhang_ufo_2024}} \\
    \footnotesize \textbf{Indirect representation} & {\footnotesize Network traffic \citepeg{song_restgpt_2023}} & {\footnotesize -} & {\footnotesize Read files \citepeg{guo_pptc_2024}} \\
    \bottomrule
    \end{tabular}
\end{table*}

Similarly, we group domain-specific action types into:
    \begin{description}
        \item[Mouse/touch and keyboard:] 
            Low-level screen coordinate-based actions such as moving the cursor, tapping on coordinates, or typing text using the keyboard. These simulate typical human input across platforms \citepeg{humphreys_data-driven_2022, wang_mobile-agent_2024, rahman_v-zen_2024}.
        \item[Direct UI access:]
            Actions targeted at specific UI elements using structured identifiers (like HTML tags or accessibility IDs) \citepeg{gur_understanding_2023, zhang_appagent_2023, branavan_reinforcement_2009}.
        \item[Task-tailored actions:]
            High-level actions that encapsulate multi-step behaviors, such as ``go to home screen'' or ``send email,'' into single commands tailored to specific tasks \citepeg{nakano_webgpt_2022, bonatti_windows_2024, wang_officebench_2024}.
        \item[Executable code:] 
            Agents may interact programmatically with their environments by generating code (e.g., Python, Bash, JavaScript) and executing it with a corresponding interpreter \citepeg{sun_adaplanner_2023, gur_real-world_2024, deng_mobile-bench_2024}.
    \end{description}
Table~\ref{tab:domain_as} provides cited examples for each action type and domain.

\begin{table*}
    \centering
    \caption{Our classification of \textbf{action types} across the different domains, along with relevant examples for each domain.}
    \label{tab:domain_as}
    \begin{tabular}{>{\raggedright\arraybackslash}p{3.cm} >{\raggedright\arraybackslash}p{3.7cm} >{\raggedright\arraybackslash}p{3.55cm} >{\raggedright\arraybackslash}p{3.55cm}}
    \toprule
   \footnotesize \textbf{Action types} & \footnotesize \textbf{Web} & \footnotesize \textbf{Android} & \footnotesize \textbf{Personal computer} \\
    \midrule
    \footnotesize \textbf{Mouse/touch/keyboard} & {\footnotesize Mouse/touch/keyboard \citepeg{humphreys_data-driven_2022}} & {\footnotesize Touch and keyboard \citepeg{wang_mobile-agent_2024}} & {\footnotesize Mouse and keyboard \citepeg{rahman_v-zen_2024}} \\
    \footnotesize \textbf{Direct UI access} & {\footnotesize HTML elements \citepeg{gur_understanding_2023}} & {\footnotesize Android elements \citepeg{zhang_appagent_2023}} & {\footnotesize Custom \citepeg{branavan_reinforcement_2009}, UI automation API \citepeg{zhang_ufo_2024}} \\
    \footnotesize \textbf{Task-tailored actions} & {\footnotesize Find on page \citep{nakano_webgpt_2022}} & {\footnotesize Go back \citep{zhang_appagent_2023}} & {\footnotesize Switch application \citep{bonatti_windows_2024}, send email \citep{wang_officebench_2024}} \\
    \footnotesize \textbf{Executable code} & {\footnotesize JavaScript, Python \citepeg{sun_adaplanner_2023}, Selenium web driver \citepeg{gur_real-world_2024}} & {\footnotesize Android debug bridge \citepeg{deng_mobile-bench_2024}} & {\footnotesize UI automation API \citepeg{wu_os-copilot_2024}, Bash \citepeg{song_mmac-copilot_2024}} \\
    \bottomrule
    \end{tabular}
\end{table*}

\subsection{Recommendations}
Our analysis of the domains of the reviewed agents (see Appendix~\Cref{fig:domain-dist}) reveals that most of the ACU literature focuses on the Web and Android domains.
In contrast, the personal computer domain, despite its significant practical relevance in workplace automation and productivity applications, remains underexplored: Only $10$ out of our $\numagents$ ACUs target desktop environments.
We recommend that \textbf{desktop environments should get more attention in research}, as desktop environments not only offer high potential for impactful automation, but also present unique research challenges, such as handling more complex applications, overlapping windows, and orchestrating inter-application workflows reliant on the shared, user-navigable file system.

\section{\texorpdfstring{Interaction Perspective (Agent $\leftrightarrow$ Environment)}{Interaction Perspective (Agent <-> Environment)}}\label{sec:interaction_view}

The interaction perspective examines how agents interact with environments through observation and action types, building on the cross-domain abstractions introduced in \cref{sec:domain_view}.

\subsection{Observation Spaces}\label{sec:obervation_space}

Observation spaces $\mathcal{O}$ of ACUs typically comprise image screen representations, textual screen representations, or indirect observations. 
Similar to the previous chapter, we classify the observation type used by the reviewed $\numagents$ ACUs (see Appendix \cref{tab:interaction_literature}).
Our analysis shows textual observations are the most common observation type with $35$ ACUs relying only on textual representations (see Appendix \cref{fig:observation-space-trend}), reflecting the influence of LLMs over the last years. 
However, we identify a trend towards image screen representation, with image observations even being the most common observation type in 2024 (see Appendix \cref{fig:observation-space-trend}), partly driven by advances in vision language models (VLM).

\subsubsection{Image Screen Representation}\label{sec:obervation_space_image}
Image-based observations (e.g., screenshots) are used across Web \citepeg{zheng_gpt-4vision_2024}, Android \citepeg{zhang_appagent_2023}, and desktop environments \citepeg{gao_assistgui_2024}.
Using screenshots aligns with human visual perception, offering broad applicability since most applications provide a graphical interface.

Besides capturing the entire screen, there exist different approaches for taking screenshots. Some approaches only use parts of the screen, such as the active application \citep{gao_assistgui_2024}, while others extend it beyond the visible viewpoint by rendering the entire application as an image \citep{chen_webvln_2024}, whereas humans have to scroll.

To reduce the computational load of processing large images, screenshot observations $o_t$ are typically simplified $o_t \rightarrow o_t^*$ by downsampling their resolution \citepeg{toyama_androidenv_2021,chen_webvln_2024}.
\citet{rahman_v-zen_2024} even combine high-resolution and low-resolution screenshots to have a compact view but still access image details if needed.

Another challenge is to feed the textual instruction $i$ into vision-only agents.
Typically, the instruction is either encoded separately and added in the embedding space \citepeg{baechler_screenai_2024} or visually rendered atop of each screenshot \citepeg{shaw_pixels_2023}.

According to our analysis in Appendix Section~\Cref{sec:domain_view:nature}, a common assumption within the field of ACUs is that the environment remains static between actions, resulting in the prevailing practice of capturing screenshots only after actions. However, real-world applications exhibit dynamic behavior, necessitating continuous monitoring and the capacity to react to asynchronous events (e.g., the arrival of a new email). This area is currently underexplored in ACU research.

\subsubsection{Textual Screen Representation}\label{sec:obervation_space_image_text}
Agents using textual screen representations operate across diverse platforms, including the Web (via HTML) \citepeg{kim_language_2023}, Android (via view hierarchies) \citepeg{shvo_appbuddy_2021}, and desktop systems (via the Windows UI automation tree) \citepeg{zhang_ufo_2024}.
However, not all textual representations are equally robust; HTML and Android view hierarchies tend to be well-structured and consistent, as they are generated through standardized frameworks that a majority of developers follow. In contrast, the Windows UI automation tree is often of lower quality (due to a variety of UI toolkits or legacy applications), leading to incomplete or semantically sparse UI descriptions. Therefore, textual representations are typically only used for Web, Android, and very specific desktop applications.

Textual representations, particularly HTML, are often verbose, as they include styling metadata in addition to content. Processing raw text \citepeg{kim_language_2023, assouel_unsolved_2023} is therefore generally restricted to artificial environments with minimal markup, such as the MiniWoB++ benchmark (see \cref{sec:datasets}). 
In more realistic environments, textual representations are typically simplified through a combination of the following strategies:

\begin{description}
    \item[Heuristic pruning:] 
        Selects only the most essential attributes such as \texttt{id}, \texttt{class}, or \texttt{name} for each element while removing others \citepeg{li_zero-shot_2023, tao_webwise_2023}.
    \item[Elements filtering:] 
        Keeps only specific elements, such as leaf elements \citepeg{gur_learning_2019}, or those considered most relevant to fulfill a given instruction $i$ \citepeg{deng_mind2web_2023, zheng_synapse_2024}.
    \item[Representation embedding:] 
        Uses an embedding model to compress HTML into a vector representation \citepeg{jia_dom-q-net_2019, gur_learning_2019, liu_reinforcement_2018}.
    \item[Text summarization:]
        Utilizes an auxiliary model to compress the HTML into an abstract text summary \citepeg{zheng_synapse_2024}.
\end{description}

A key advantage of using HTML is its alignment with the pretraining data of LLMs, enabling LLM-based agents to exhibit a general understanding of it. 
To leverage this alignment in the Android domain, \citet{wang_enabling_2023} propose to map the Android view hierarchy to simplified HTML, an approach later adopted by subsequent works \citepeg{deng_mobile-bench_2024}.

This mapping approach has also been extended to image-based agents, where screenshots are translated into textual representations to align with text foundation models. The process, applied in Web \citepeg{cho_caap_2024}, Android \citepeg{li_uinav_2024}, and desktop environments \citepeg{gao_assistgui_2024}, typically involves two steps: First, object detection is used to detect UI elements, and then an additional model is used to extract an element's properties, such as its text and type.

\subsubsection{Bi-Modal Screen Representation}\label{sec:obervation_space_bimodal}
Bi-modal observations combine both image and textual inputs, and have been explored across domains such as Web \citepeg{he_webvoyager_2024}, Android \citepeg{sun_meta-gui_2022}, and desktop systems \citepeg{zhang_ufo_2024}, aiming to unify complementary information streams.
Typically, the two modalities have modality-specific encoders that embed the two types of observations before they are combined in the embedding space \citepeg{furuta_multimodal_2024}.
The potential of bi-modal agents leveraging the advantages of both modalities is an open research question, as more information can also lead to distracting information overload through irrelevant content and has not yet been proven to be considerably more effective on current benchmarks.

\subsubsection{Indirect Observation}\label{sec:obervation_space_indirect}
Some agents do not directly observe screen representation but have actions or routines to collect information about the current computer state $s_t$ \citepeg{qin_toolllm_2024,kong_tptu-v2_2023,guo_stabletoolbench_2024}.
For example, \citet{guo_pptc_2024} use a content reader routine that at each time step $t$ reads information from a PowerPoint file as observation $o_t$.
\citet{song_restgpt_2023} execute a REST-API call as an action $a_t$, and use the API \emph{response} as the next observation $o_{t+1}$. 
Similarly, \citet{wang_officebench_2024} use task-tailored actions to directly read information from files, e.g., \texttt{read\_excel\_file}, or use application-specific actions, e.g., an action \texttt{list\_emails} in an email application.

\subsubsection{Recommendations}
ACU agents commonly rely on either image-based or textual screen observations, each offering distinct advantages. 
Textual representations encode rich semantic information such as element attributes and hierarchical structures.
For example, a form element might include a semantic identifier such as \texttt{email-sender}, or a table may be represented as a hierarchy of rows and cells.
However, we hypothesize that agent behaviors trained on such structured representations are often brittle when applied across diverse applications, such as different websites.
This brittleness arises from a dependency on \emph{optional} semantic information, which is frequently absent, incomplete, inconsistent, or ambiguous in real-world settings.
For example, a form might lack a descriptive \texttt{id} attribute, or tables may be implemented using non-standard constructs. As a result, agents leveraging this information may develop scenario-specific heuristics (shortcuts, cp. \citet{geirhos2020shortcut}) that do not generalize well beyond the training data.

In contrast, image-based screen representations tend to exhibit greater consistency across scenarios due to widely adopted design conventions.
This consistency suggests that image-based observations can support the development of more robust and generalizable agent behavior. 
Based on this reasoning, we argue that \textbf{versatile ACUs should rely on visual perception} to enhance generalization and applicability in all scenarios where humans also operate computers.
A detailed comparison of textual versus image-based screen representations is provided in Appendix~\Cref{sec:img_vs_text}.

Our analysis of datasets supports our suspicion of brittle behavior for text-based agents.
In the Web domain, MiniWoB++ \citep{shi_world_2017,liu_reinforcement_2018} provides an artificial environment with unrealistic uniform HTML representations across its tasks.
In such sanitized settings, textual agents achieve strong performance, as shown by \citet{humphreys_data-driven_2022}, where removing the textual modality of a bi-model input resulted in a 75\% drop in performance.
However, in benchmarks based on realistic websites like Mind2Web \citep{deng_mind2web_2023}, success rates for text-based agents fall below 10\% \citep{deng_mind2web_2023, gur_real-world_2024, furuta_multimodal_2024}, while image-based agents leveraging large vision models (e.g., GPT-4V) achieve significantly higher success rates of up to 38\% \citep{zheng_gpt-4vision_2024}, hinting on the importance of visual input in real-world settings.

A similar shift is observable in the Android domain: while early systems favored textual representations \citep{wang_enabling_2023, shvo_appbuddy_2021}, recent image-based agents such as \citep{zhang_appagent_2023} outperform them in more dynamic or complex applications.

\subsection{Action Spaces} \label{sec:action_space}

As introduced in \cref{sec:domain_view}, ACUs can utilize four main types of actions: mouse/touch and keyboard actions, direct UI access, task-tailored actions, and executable code. 
In the following, we discuss each of these action spaces, while Table~\ref{tab:interaction_literature} provides the corresponding details for each ACU.

\subsubsection{Mouse/Touch and Keyboard}\label{sec:action_space_mouse}

\begin{figure}[t]
    \centering
    \includegraphics[width=0.65\linewidth]{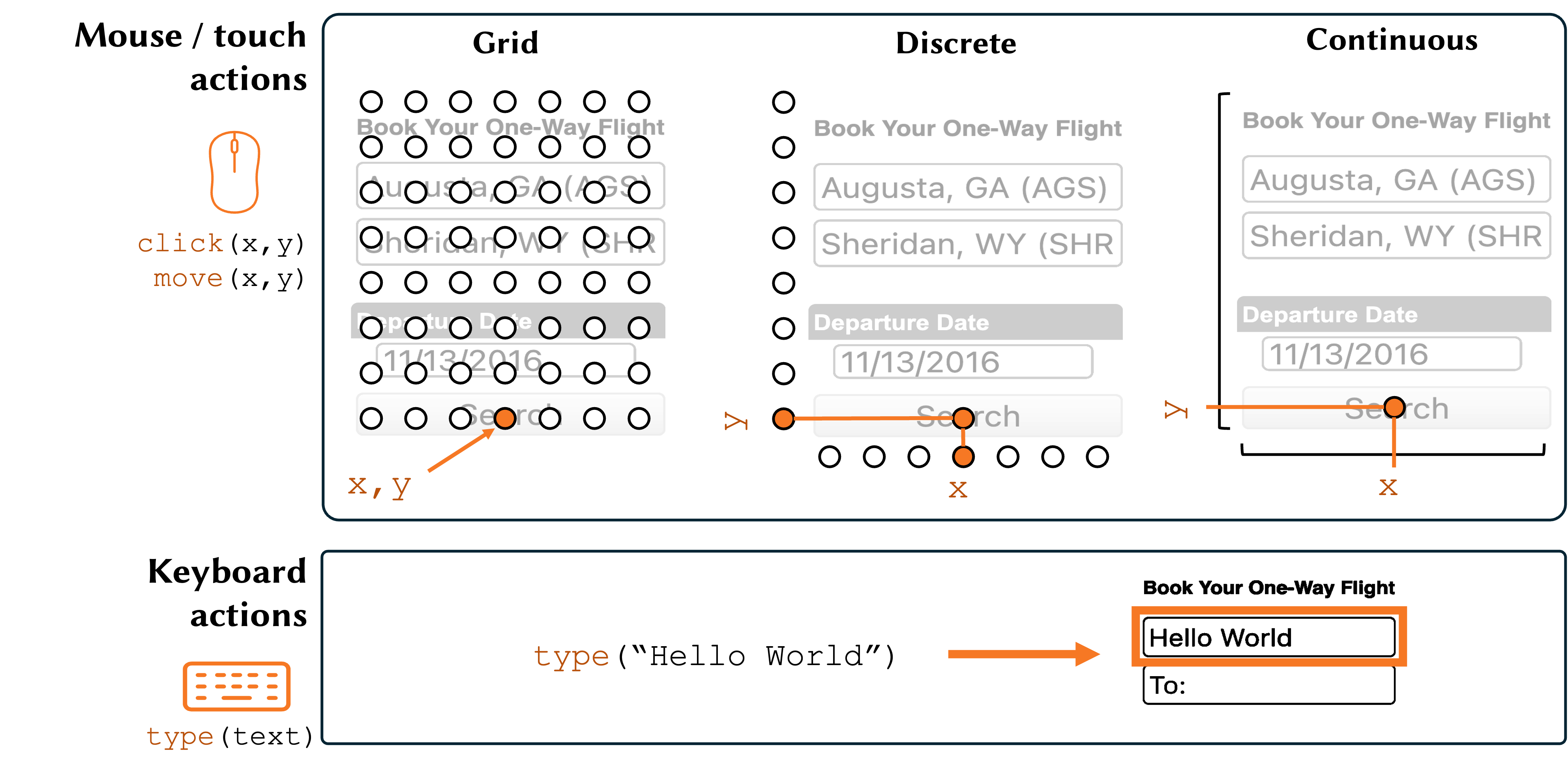}
    \caption{Common mouse and keyboard actions, highlighting coordinate prediction.}
    \Description{Visualization of mouse and keyboard action modalities in agent control interfaces.
    The figure clarifies differences between common screen-coordinate prediction modes in user interface control. In the top section, three types of mouse or touch actions are shown using illustrative overlays: (1) grid-based prediction constrains clicks to fixed grid points, (2) discrete prediction allows selection from independent horizontal and vertical positions, and (3) continuous prediction enables unconstrained coordinate selection. These modes are visualized over a web form to highlight spatial precision. The bottom section shows a keyboard input scenario, where an agent types into a text box within a form interface. This comparison illustrates how action types differ in terms of input structure, spatial for mouse/touch, symbolic for keyboard, and supports the distinction between discrete and continuous action spaces.}
    \label{fig:mouse-keyboard-action}
\end{figure}

Mouse, touch, and keyboard actions align closely with human interaction patterns, facilitating data collection and training \citep{humphreys_data-driven_2022}.
Both mouse actions (e.g., \texttt{click(x,y)}) and touch actions (e.g., \texttt{tap(x,y)}) require \emph{absolute} screen coordinates \texttt{(x,y)}, making them conceptually identical for ACUs\footnote{For humans, mouse actions are relative (to the current cursor position) and touch actions are absolute.}.
\Cref{fig:mouse-keyboard-action} summarizes approaches for predicting screen coordinates.
Some methods make discrete predictions, by either predicting a position on a low-resolution coordinate grid \citepeg{shi_world_2017,toyama_androidenv_2021}, predicting two interdependent discrete values for the \texttt{x} and \texttt{y} coordinates \citepeg{humphreys_data-driven_2022}, or generating discrete tokens through a text generation model \citepeg{hong_cogagent_2024}. Other approaches use continuous values by predicting two interdependent continuous coordinate values \citepeg{toyama_androidenv_2021}.
It remains unclear whether one prediction strategy is universally superior; instead, the choice usually depends on the agent architecture and task.
%

Keyboard actions (e.g., \texttt{type(text)}) are typically used to input text into a previously selected UI element.
While earlier methods relied on predefined text fragments \citepeg{humphreys_data-driven_2022} or extracted text from the instruction $i$ \citepeg{gur_learning_2019} as input, in most of the current systems, ACUs generate the text using a language model \citepeg{hong_cogagent_2024}, as this provides the required freedom to type diverse texts. 
%
%
Beyond typing, keyboard actions are frequently used for special commands, such as navigating via pressing arrow keys \citepeg{li_zero-shot_2023} or using shortcuts such as \texttt{select all}, \texttt{copy}, or \texttt{paste} \citepeg{cho_caap_2024}.

\subsubsection{Direct UI Access}\label{sec:action_space_direct}

\begin{figure}[b]
    \centering
    \includegraphics[width=0.6\linewidth]{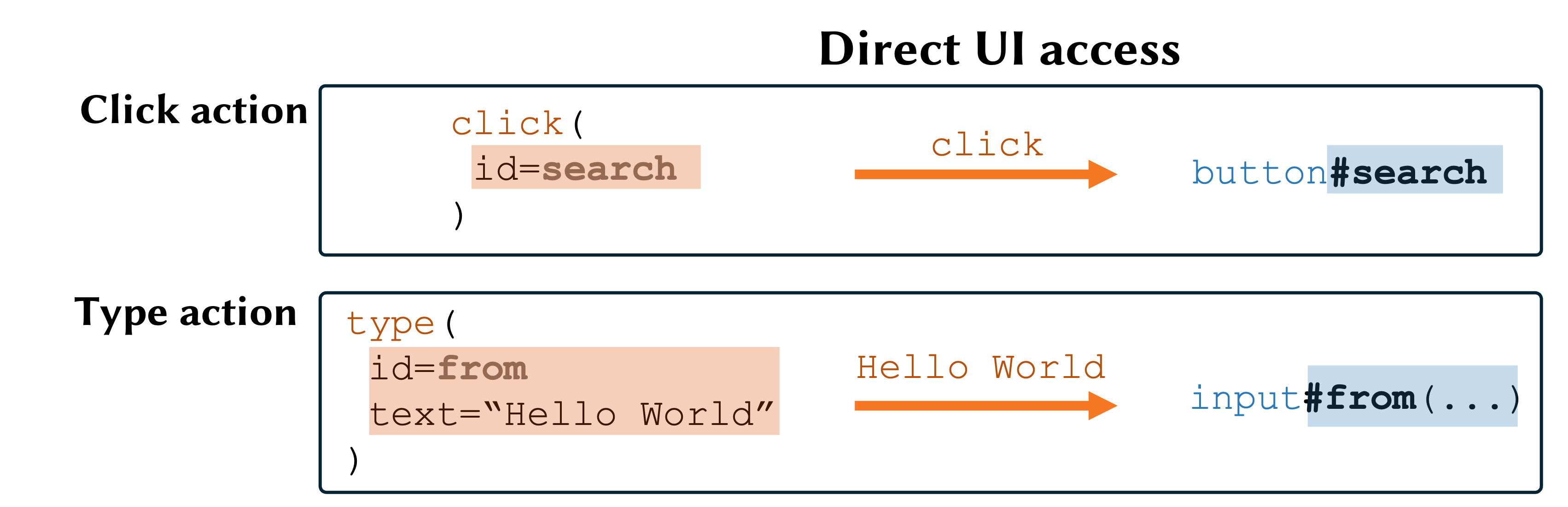}
    \caption{Common direct UI access actions, referencing the HTML element by its \texttt{id} attribute.}
    \Description{The figure illustrates examples of direct UI access targeting HTML elements via their bolded \texttt{id} attributes. The first example demonstrates a click action where the function call specifies the \texttt{id} as ``search'' to trigger a button click. The second example shows a type action, in which the function call targets the input field with \texttt{id} equals ``form'' to enter the text ``Hello World.'' Arrows visually connect each function call to its respective HTML element, emphasizing the explicit mapping between the code and interface components.}
    \label{fig:direct-action}
\end{figure}

Direct UI access actions such as \texttt{click(e)} or \texttt{type(e, text)} target a specific UI element \texttt{e} observed by the agent.
Agents typically identify these elements \texttt{e} by either predicting unique identifiers like HTML \texttt{id} tags \citepeg{li_zero-shot_2023}, XPath\footnote{https://www.w3.org/TR/xpath-31/} descriptions \citep{kim_language_2023} (for an example of predicting \texttt{click(id=search)} based on \texttt{\(<\)button id="search"\(>\)} see \cref{fig:direct-action}), or by scoring and selecting from all visible elements \citepeg{jia_dom-q-net_2019, li_uinav_2024}.

To simplify selection, agents may restrict referenceable elements to leaf nodes in the user interface tree \citepeg{liu_reinforcement_2018} or pre-filter them with an auxiliary model that can preselect potential candidate elements \citepeg{deng_mind2web_2023}. For specific tasks such as web navigation, agents may be limited by design to selecting hyperlinks only \citep{zaheer_learning_2022, chen_webvln_2024}.

After selecting an element, optional text input follows similar generation strategies as got keyboard input, including generating free-form text \citepeg{li_zero-shot_2023}, selecting predefined fragments \citepeg{shvo_appbuddy_2021}, or extracting text from the instruction \citepeg{jia_dom-q-net_2019}.

By classifying each reviewed ACU, we identify direct UI access as the most widely adopted action space in the literature (see Appendix~\cref{fig:action-space-dist}). We believe this dominance is due to its balance between generality and learnability. Unlike coordinate-based mouse or touch actions, which require fine-grained spatial reasoning and introduce high-dimensional prediction challenges, direct UI access operates over a lower-dimensional and semantically meaningful action space. By allowing agents to refer to structured element identifiers (e.g., \texttt{id}, XPath, or other selectors), this form of interaction abstracts away spatial complexity while still supporting a broad range of tasks.
However, as it requires an identifier for the UI elements, it is primarily compatible with text observations and, as discussed in the previous section, does therefore not scale well to the desktop domain.
Also, it typically only works well for simplified, well-structured UIs that are often not available for real-world use cases but common for earlier benchmarks (see \cref{sec:datasets}).

\subsubsection{Task-Tailored Actions}\label{sec:action_space_task}
Task-tailored actions are environment-specific commands for common operations.
For instance, \citet{wang_officebench_2024} define application-specific actions such as \texttt{create\_event} for a calendar application and \texttt{send\_email} for an email client.
These high-level actions reduce learning complexity as they typically correspond to an entire trajectory of clicking actions.
Nevertheless, the downside is that they require additional engineering as these subroutines are often hand-crafted \citepeg{wang_officebench_2024, tan_towards_2024}.

We consider most task-specific actions as a shortcut that might help to improve on narrowly designed benchmarks, but are not helpful for building general ACUs, especially when the actions are highly task-specific.
However, a few task-tailored actions demonstrate broader applicability and merit inclusion due to their capacity to generalize across tasks within a given domain.
For example, \citet{bonatti_windows_2024} define the action \texttt{open\_application}, which enables an agent to open and switch between applications on a Windows operating system. Similarly, \citet{nakano_webgpt_2022} define a \texttt{search} action, which allows the agent to navigate to specific text positions within a website. These actions exemplify a favorable trade-off between general-purpose utility and domain-relevant specialization, particularly when integrated with more comprehensive action spaces.

\subsubsection{Executable Code}\label{sec:action_space_code}
Agents may also generate code, executed by interpreters like Python or Bash.
Executable code varies in its structure and the level of abstraction provided by its application programming interface (API):

\begin{description}
    \item[Structure] of generated code: \hfill
        \begin{description}
            \item[Straight-line code] consists of a sequence of statements without control flow \citepeg{tao_webwise_2023}. It is akin to predicting a single or multiple actions.
            \item[Control-flow code] includes control flow mechanisms such as conditional statements (e.g., \texttt{if}), loops (e.g., \texttt{for}), and function definitions. Complex code can represent the agent's entire execution plan \citep{sun_adaplanner_2023}, where the agent dynamically adjusts its plan based on precondition checks failing.
        \end{description}
    \item[API abstraction level] utilized by generated code: \hfill
        \begin{description}
            \item[General-purpose API:]
                Some agents use an API with functions akin to general-purpose actions like clicking elements or screen coordinates. For example, \citet{gur_real-world_2024} use the Selenium web driver API\footnote{\url{https://www.selenium.dev/documentation/webdriver/}} to provide such low-level interactions.
            \item[Task-tailored API:]
                Other agents use an API of hand-engineered functions tailored to tasks. For example, \citet{guo_pptc_2024} define functions like \texttt{insert\_rounded\_rectangle(...)} for their PowerPoint agent.
        \end{description}
\end{description}

See Appendix \cref{fig:action-space-code} for both code structure and API abstraction examples.
Executable code is generated using either general foundation models \citepeg{guo_pptc_2024} or specialized models \citepeg{gur_real-world_2024}. Foundation models often come pre-trained on well-established APIs like Selenium web driver, while hand-engineered functions are typically introduced through contextual prompts or, alternatively, by using an API selector to first retrieve relevant functions \citep{song_restgpt_2023}.

An open question is the benefits of using executable code over other action types.
\citet{chen_program_2023} and \citet{gao_pal_2023} found that producing straight-line code instead of predicting actions as strings can reduce hallucinations when using GPT-3. 
However, \citet{assouel_unsolved_2023} suggested that this advantage disappears when using GPT-4, indicating that the benefits of executable code over other action types diminish with more advanced models.

\subsubsection{Action Grounding}\label{sec:action_space_grounding}
Action grounding  $a_t^* \rightarrow a_t$ refers to the process of converting an abstract action, such as \texttt{click submit button}, into an executable action, such as \texttt{click(e)}, where \texttt{e} represents the specific UI element. Grounding is typically required when a text foundation model generates an abstract, text-based plan that must be transformed into a sequence of executable actions \citepeg{gao_assistgui_2024, kim_language_2023}.
Two common strategies include:

\begin{description}
    \item[Prediction-based grounding:] 
        Given an abstract action, a grounding model predicts the corresponding UI element. For example, for the abstract action \texttt{navigate to settings}, \citet{li_mapping_2020} predict \texttt{click(e)} where \texttt{e} refers to the settings app icon.
    \item[Rule-based grounding:] 
        A rule-based module matches an abstract action $a_t^*$ to the actionable UI element. For example, \citet{song_visiontasker_2024} use text matching rules to achieve this mapping, whereas \citet{lee_explore_2023} first predict abstract \emph{template} actions containing placeholders (e.g., \texttt{click(text=``[contact\_name]'')}), followed by rule-based grounding substituting the placeholders with context-specific values derived from the user instruction $i$.
\end{description}

Grounding is not limited to textual models but is also used in vision-based agents.
These vision-based agents rely on grounding because current vision models struggle to predict screen coordinates accurately. Several strategies for grounding in vision models have been explored and discussed by \citet{zheng_gpt-4vision_2024}.
The most successful one is \emph{set-of-mark} prompting \citep{yang_set--mark_2023}, where actionable elements are annotated with bounding boxes and unique identifiers, enabling the agent to access them directly using the identifier instead of relying on coordinate prediction.
However, this prompting strategy requires identifying the actionable elements, which is done by either using an additional textual screen representation with positional data \citepeg{zheng_gpt-4vision_2024,li_appagent_2024,zhang_appagent_2023} or extracting them from the screenshot via a specialized model \citepeg{lu_omniparser_2024}. 
Although the latter approach offers flexibility, it is often imprecise, leading to suboptimal performance \citep{bonatti_windows_2024}.

Despite the success of set-of-mark prompting, we posit that this grounding step may be a temporary workaround, designed to compensate for the limitations of current vision foundation models, which have not yet been trained sufficiently to predict screen coordinates directly. Recent studies suggest that learning visual grounding via coordinate prediction is feasible and straightforward \citep{dardouri_visual_2024, cheng_seeclick_2024} and may eventually render the need for set-of-mark prompting unnecessary.

\subsubsection{Recommendations}
\begin{figure}[t]
    \centering\includegraphics[width=1\linewidth]{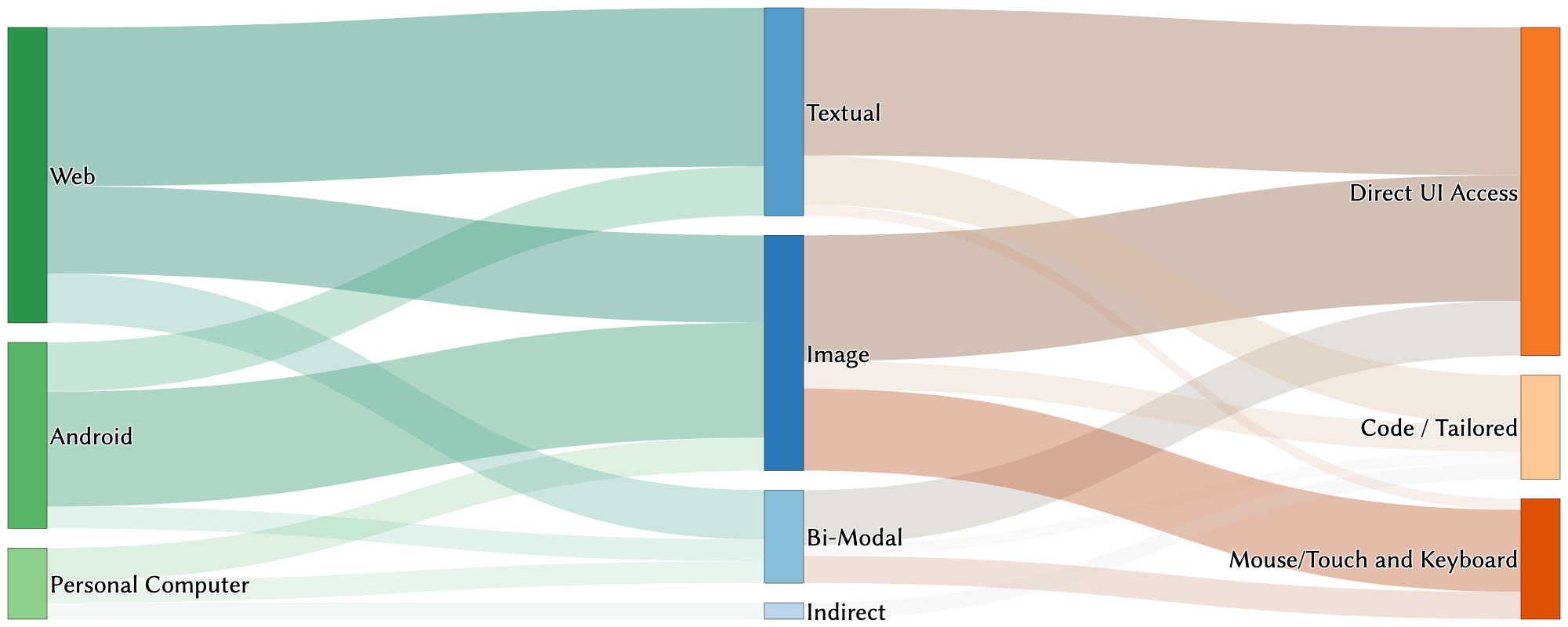}
    \caption{Sankey diagram showing the connections between domains (left) and observation spaces (middle), and between observation and action spaces (right). The path width represents the number of papers within this survey associated with the corresponding topic.}
    \Description{A Sankey diagram with three columns depicting the relationships among surveyed research papers across domains, observation types, and action types. The left column enumerates the domains: Web, Android, and personal computer. The middle column presents observation types categorized as Image, textual, bi-modal, and indirect. The right column displays action types, including direct user interface (UI) access, mouse/touch and keyboard input, and code/tailored actions. Flows connect domains to observation types and subsequently to action types, with flow widths proportional to the number of papers associated with each connection. The strongest flows from domains to observations are from Web to textual data, Android to image, and Web to image. Among observation-to-action connections, the most prominent flows are from textual and image observations to direct UI access, with notable flows from image observations to mouse/touch and keyboard actions. This visualization provides a quantitative summary of prevailing research focuses within the surveyed literature.}
    \label{fig:domain-os-as-sankey}
\end{figure}

Different action types demand distinct observational inputs.
Coordinate-based actions (e.g., mouse clicks) depend on spatial information, whereas direct UI actions (e.g., button presses) must be able to reference UI elements. 
Consequently, coordinate-based actions naturally align with visual input, while element-based actions align with text-based input.

Nevertheless, our analysis shows many deviations from this expected alignment through modality bridging.
For instance, many vision-based agents employ direct UI access actions (see \cref{fig:domain-os-as-sankey}).
While such strategies can be effective, we argue that they are short-term workarounds tailored to existing technological constraints, introducing unnecessary long-term architectural complexity.

Historically, during the dominance of text-only LLMs, many ACUs converted screenshots into textual representations to maintain compatibility \citepeg{song_visiontasker_2024,wen_autodroid_2024}.  More recently, techniques such as set-of-mark prompting have been adopted to compensate for shortcomings in precise coordinate prediction \citepeg{zheng_gpt-4vision_2024,zhang_appagent_2023}.
These workarounds are likely to diminish in relevance as vision foundation models improve in spatial and semantic grounding.

Expanding on the premise that image-based observations offer a more coherent and spatially continuous representation of the user interface, we propose that \textbf{versatile ACUs should rely on mouse, touch, and keyboard actions} as these actions align naturally with visual observations.
Moreover, these actions can still be combined with higher-level subroutines (e.g., application switching) that abstract common interaction patterns into single actions.

\section{Agent Perspective}\label{sec:conceptional_view}

While previous sections described the agent's external environment and interactions, this section examines the internal structure of ACUs.
Here, we focus on two prevalent agent types: \emph{Foundation agents} (based on foundation models) and \emph{specialized agents} (based on domain-specific design).

\begin{description}
    \item[Foundation Agent:] \label{sec:foundation_agent}
    A foundation agent \citepeg{zheng_gpt-4vision_2024} uses a \emph{general pre-trained} foundation model (such as an LLM or VLM) as its policy $\pi$. While currently dominated by LLMs and VLMs, this category encompasses any architecture leveraging broad pre-training for zero-shot or few-shot transfer, including vision-language-action (VLA) models or diffusion models. These agents employ the model's broad knowledge and in-context learning capabilities for \emph{episodic improvement} (see \cref{sec:learning_strategy_pretraining,sec:episodic_improvement}).
    For example, a text-based foundation model receives a textual observation $o_t$ alongside a prompt specifying its role as agent, a description of available UI actions $\mathcal{A}$, and the instruction $i$. The model then \emph{generates} an action $a_t$ that is executed in the environment.
    \item[Specialized Agent:] \label{sec:reinforce_agent}
    A specialized agent \citepeg{humphreys_data-driven_2022} employs a \emph{custom} network architecture as its policy $\pi$, which \emph{predicts} actions $a_t \in \mathcal{A}$ based on a given observation $o_t$ and instruction $i$, relying on the possibilities of the predefined output options.
    For example, the architecture might process an image $o_t$ and a text instruction $i$ as inputs through encoder networks and predict logits for each action type (such as \texttt{clicking}) alongside additional outputs (such as screen coordinates \texttt{(x,y)}). 
    Learning typically involves \emph{environment learning} techniques such as reinforcement learning (see \cref{sec:environment_adaption}).
\end{description}

Specialized agents work particularly well on narrow tasks and when the task conditioning of humans (the instruction) is limited (e.g., fill out a simple form given a user ID). In such cases, specialized agents often perform robustly and are computationally efficient due to their smaller parameter count \citepeg{humphreys_data-driven_2022}.
However, in multi-step tasks with diverse observation spaces or strong instruction-based conditioning, agents clearly benefit from general pre-training and reasoning techniques such as Chain-of-Thought (CoT) \citepeg{zhou_webarena_2024}. 

\Cref{table:two-main-agents} summarizes the key characteristics of the two common agent designs, highlighting their differences in architecture, action type, memory of information from past episodes, and learning strategy.

\begin{table}
    \caption{Properties of the two common ACU types.}
    \label{table:two-main-agents}
    \begin{tabular}{lllll}
    \toprule
   \textbf{ACU Agent Types } & \textbf{Architecture}   & \textbf{Action}  & \textbf{Memory}  & \textbf{Learning Strategy} \\ 
    \midrule
    Foundation agent & LLM / VLM               & Generation       & history-based  & General + Episodic \\
    Specialized agent  & Custom                  & Prediction       & state-based    & Environment learning  \\ 
    \bottomrule
    \end{tabular}
\end{table}

\subsection{Policy -- How to Act}\label{sec:agent_policy}
The policy $\pi$ defines how an agent selects actions \citep[Chapter~1.3]{sutton_reinforcement_2018}. For computer control, we distinguish three types of policies: Memoryless policies that act only on the current input; history-based policies that use explicit past observation and/or action sequences; and state-based policies that aggregate information about the past in an internal memory. 
Among them, history-based policies are the main research focus in the surveyed reviews, with almost $60$ out of the reviewed $87$ ACUs using some form of history in their policy (see Appendix \cref{fig:policy-dist} and \cref{tab:agent_literature}).

\subsubsection{Memoryless Policies}\label{sec:agent_policy_memoryless}
Memoryless policies \citepeg{chen_webvln_2024} ignore past observations and actions and act solely on the current observation $o_t$:
\begin{equation}
a_t \sim \pi(\ \cdot\ |\ o_t,\ i\ ) \label{eq:memoryless-policy}
\end{equation}

Memoryless policies are often insufficient for real-world control scenarios where context across time is critical and selecting an appropriate next action $a_t$ requires information about past observations $o_{t-n}, ..., o_{t-1}$ and/or actions $a_{t-n}, ..., a_{t-1}$. 
For instance, in the context of purchasing multiple items from an online store, an agent must remember which items were already added to the shopping cart.
Still, since memoryless can be sufficient for specific tasks, their simplicity is sometimes leveraged in model design \citepeg{shvo_appbuddy_2021}.

\subsubsection{History-based Policies}\label{sec:agent_policy_history}
History-based policies track the past by adding observations and actions in a continuously growing sequence, called history $h_t=\left(o_{0},\hdots,o_{t-1},a_{0},\hdots,a_{t-1}\right)$ \citepeg{zheng_gpt-4vision_2024}.
For example, a vision-only agent's history consists of all the screenshots it perceived and the actions it performed during an episode. When predicting the next action $a_t$, the agent retrieves relevant information from its history $h_t$:
\begin{equation}
a_t \sim \pi(\ \cdot\ |\ o_t,\ i,\ h_t\ ) \label{eq:history-based-policy}
\end{equation}

Foundation agents commonly follow this pattern, as foundation models typically come with a context window to track past information, a specific instance of a policy history.
A key challenge with history-based approaches lies in the high dimensionality of observations $o_t$ (often screenshots or long textual descriptions). They either do not fit into the limited context windows of foundation models or, if they fit, they are computationally very expensive due to the large amount of required tokens.
Therefore, the history $h_t$ is often approximated as $h_t^*$. Common simplifications include:

\begin{description}
    \item[Actions only:] Keep only the past actions $h_t^*=\left(a_0,\hdots,a_{t-1}\right)$ and discard the observations \citepeg{zheng_gpt-4vision_2024}. In extreme cases, retain only the last action $h_t^*=\left(a_{t-1}\right)$ \citepeg{gao_assistgui_2024}.
    \item[Selective observations] Retain certain previous observations $o_{<t}$, such as keeping the last two screenshots with all actions as $h_t^*=\left(o_{t-2},o_{t-1},a_0,\hdots,a_{t-1}\right)$ \citep{furuta_multimodal_2024}.
    \item[Embedded summaries:] Create embeddings of the last observations $h_t^*=\left(\tilde{o}_{t-4},\hdots,\tilde{o}_{t-1},a_0,\hdots,a_{t-1}\right)$ \citep{lu_gui_2024}.
    \item[Text summaries:] Summarize past observations into text, enabling to keep the entire summarized observation history alongside raw actions $h_t^*=\left(\tilde{o}_{0},\hdots,\tilde{o}_{t-1},a_0,\hdots, a_{t-1}\right)$ \citep{zheng_synapse_2024}.
\end{description}

These strategies reduce token usage but risk omitting essential information, as the function for reducing information is not optimized for the given task. While suitable for simple GUIs, they are likely to limit performance in tasks requiring longer-term reasoning.

\subsubsection{State-based Policies}\label{sec:agent_policy_state}
In contrast, state-based policies rely on a compact internal memory $m_t$, often referred to as a Markov state, that is of fixed dimensionality and used during the action selection process \citepeg{humphreys_data-driven_2022}:
\begin{equation}
a_t \sim \pi(\ \cdot\ |\ o_t,\ i,\ m_t\ ) \label{eq:state-based-policy}
\end{equation}
This internal state $m_t$ is updated at each time step via a deterministic state-update function, typically of the form $m_{t+1}=f_m(o_{t},m_{t})$.
In practice, $f_m$ is commonly implemented as a learnable function, such as a recurrent neural network, enabling the agent to track task-relevant aspects of the history in a compressed representation for improved decision-making \citepeg{humphreys_data-driven_2022}.

While foundation models typically employ history-based policies, specialized agents often rely on state-based policies.
A notable exception is \citet{zhang_appagent_2023}, who propose a foundation agent with a state-based policy using an external \emph{text-based} state $m_t$. The foundation model not only generates the next action $a_t$ but also the next state $m_{t+1}$ given the current state $m_{t}$ and observation $o_t$, effectively operating as both policy and state-update function.

\subsubsection{Mixed Policies}\label{sec:agent_policy_mixed}
Mixed policies are hybrid approaches that combine history and state.
For example, \citet{bonatti_windows_2024} prompt their internal foundation model with the past actions and the last observation $h_t^*=\left(o_{t-1},a_0,\hdots, a_{t-1}\right)$, while keeping an external text-based state $m_t$.
Similarly, \citet{iki_berts_2022} feed the current observation $o_t$, the last action $h_t^*=\left(a_{t-1}\right)$, and an external text-based state $m_{t}$ into a fine-tuned model to predict the next action $a_t$ as well as state $m_{t+1}$.


\subsubsection{Recommendations}
Most state-of-the-art approaches leverage foundation models that use history-based policies.
Nevertheless, processing the full, unfiltered episode history at every decision step is computationally inefficient and, in many cases, infeasible. This limitation necessitates some form of history simplification for history-based policies. 
A common approach involves retaining only past actions while discarding observations. Although effective for current benchmarks, this method deliberately omits observation information that is essential for solving more complex tasks requiring reasoning over temporally distributed observations.

Other history simplification strategies are either manually crafted \citepeg{cho_caap_2024} or based on generic summarization models \citepeg{zheng_synapse_2024}, both lacking adaptability to environments.
We posit that the ability to act effectively in complex environments inherently involves the capacity to learn what historical information is relevant and what can be safely ignored. Therefore, we argue that \textbf{history simplification should be a learnable component of the agent}, integral to achieving robust and generalizable behavior for complex tasks. The literature on \emph{world models} is herein closely related (cp. \citet{ha2018world}, \citet{lecun_path_2022}, \citet{hafner2025mastering}).

Notably, specialized agents employ state-based policies, wherein past information is compressed into a Markovian state via a learnable state-update function.
This function can be interpreted as a form of learned history simplification.
This conceptual link between specialized and foundation agents, revealed by our taxonomy, may inspire future history simplification research for foundation agents.

\subsection{Learning Strategy - How to Learn to Act}\label{sec:learning_strategy}

An agent’s learning strategy can involve up to three stages (not all ACUs utilize every stage):

\begin{description}
    \item[General pre-training:] 
        The agent acquires broad, environment-agnostic knowledge. Examples include foundation models learning general-purpose capabilities or vision backbones learning image representations.
    \item[Environment learning:] 
        The agent learns to adapt to a specific computer environment. This involves explicit parameter (weight) updates or implicit methods, such as storing environment experiences for later retrieval.
    \item[Episodic improvement:] 
        The agent refines its performance within the current episode through methods such as instruction tuning or few-shot learning \citep{brown_language_2020}. Unlike the previous steps, this step does not result in persistent learning, as changes are discarded post-episode.
\end{description}

\begin{figure}[t]
    \centering
    \includegraphics[width=0.95\linewidth]{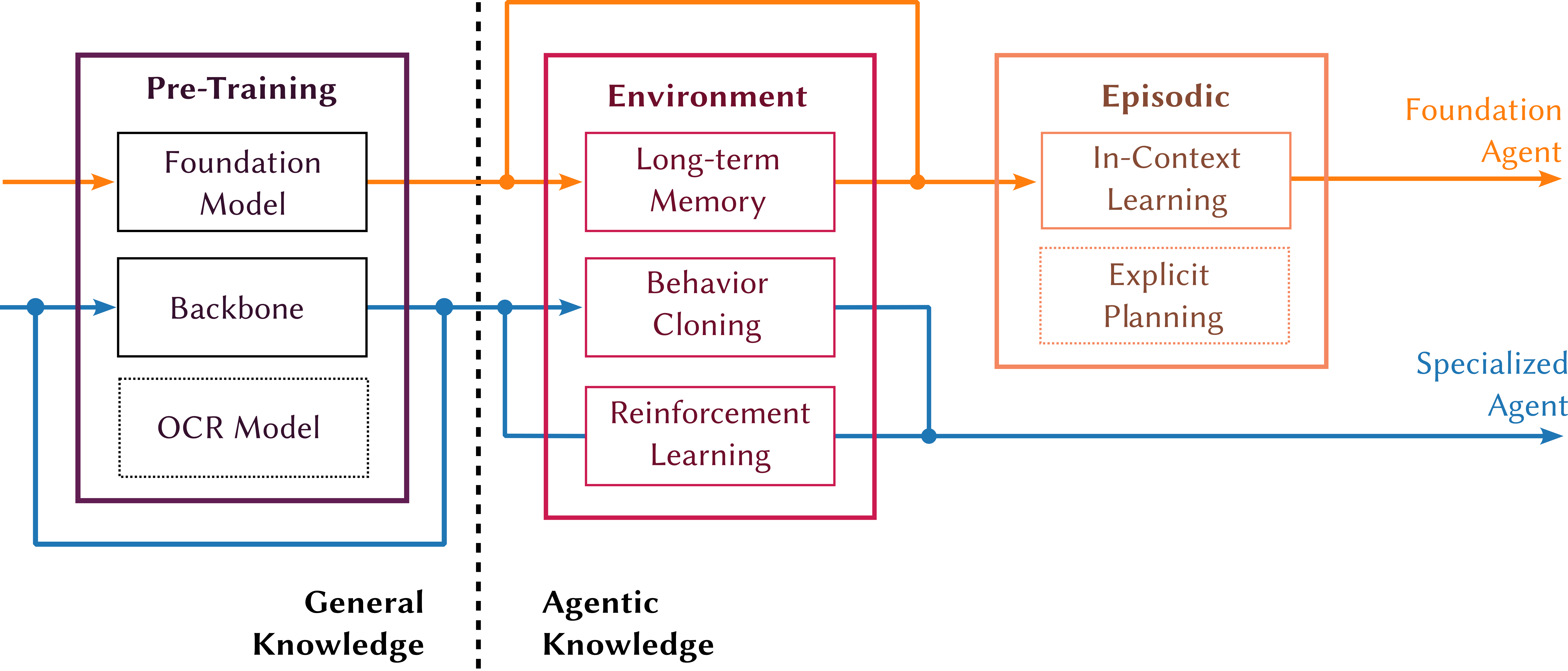}
    \caption{Overview of learning steps and strategies: \emph{Pre-training} involves acquiring broad, environment-agnostic knowledge. \emph{Environment learning} and \emph{episodic improvement} hone a ACU's agentic capabilities. A combination of these steps defines a learning strategy. ACUs typically follow one of two strategies: (1) \emph{Specialized agents} (blue)  start from scratch or use a pre-trained backbone, learn to act in a specific environment through behavioral cloning (BC) or reinforcement learning (RL) ; (2) \emph{foundation agents} (orange) begin with a general-purpose foundation model, optionally storing successful episodes for future demonstration retrieval, and employ in-context learning.}
    \Description{The diagram depicts the learning strategies and typical training pathways of agents for computer use, organized into three main stages arranged from left to right: pre-training, environment learning, and episodic improvement. The pre-training stage includes the foundation model, backbone, and OCR model components and corresponds to general knowledge acquisition. The environment learning stage involves long-term memory, behavior cloning, and reinforcement learning, which together form agentic knowledge. The episodic improvement stage consists of in-context learning and explicit planning, aimed at refining agent performance in specific tasks. Two main learning strategies are illustrated by color-coded flow paths: foundation agents begin with the foundation model, optionally leverage components in the environment stage, and utilize in-context learning during episodic improvement, while specialized agents start from the backbone model or from scratch and rely on behavior cloning or reinforcement learning during environment learning. A dashed vertical line separates the pre-training phase from the subsequent stages, highlighting the transition from general to agentic knowledge. The figure emphasizes that foundation and specialized agents generally follow distinct and limited combinations of these training strategies.}

    \label{fig:learning-flow}
\end{figure}

\Cref{fig:learning-flow} illustrates how these steps sequentially combine into learning strategies for both foundation and specialized agents.
Our analysis of learning strategies shows that the combination of general pre-training with prompting is the most common strategy (see Appendix~\Cref{fig:learning-strategy-dist}).
The following sections explore each learning step in detail, emphasizing current practices and highlighting exceptions.

\subsubsection{Leveraging General Pre-Training} \label{sec:learning_strategy_pretraining}
General pre-training serves as an initialization stage for an agent. This initial knowledge can be modified and adapted through \emph{environment learning} (e.g., fine-tuning, see \cref{sec:environment_adaption}) or preserved and utilized via \emph{episodic improvement} (e.g., prompting, see \cref{sec:episodic_improvement}).

Foundation agents primarily rely on the former approach. They leverage foundation models with broad knowledge and in-context learning capabilities \citep{brown_language_2020}. These capabilities can eliminate the need for environment-specific fine-tuning, allowing agents to operate in computer environments using only the foundation model's broad knowledge and instructions provided through prompts to adapt to specific environments \citepeg{kim_language_2023}.
For example, GPT-4 \citep{openai_gpt-4_2024}, when prompted as a web agent, can complete tasks such as filling out forms or navigating website links \citep{zheng_gpt-4vision_2024}.

In contrast, specialized agents are either trained from scratch \citepeg{humphreys_data-driven_2022} or initialized with a pre-trained backbone (e.g., an image encoder) to accelerate learning the observation space \citep{li_mug_2024}. These agents typically require additional fine-tuning to adapt to computer environments (see \cref{sec:environment_adaption}).
The foundation model or backbone choice depends on the observation space, action space, and specific task requirements. For example, 
    \citet{zheng_gpt-4vision_2024} use GPT-4 \citep{openai_gpt-4_2024} as a multi-modal foundation model for their bi-modal agent.
    \citet{gur_real-world_2024} employ a coding-proficient foundation model \citep{chung_scaling_2022} to generate executable code.
    \citet{shaw_pixels_2023} fine-tune a vision backbone for their vision-based agent.
    \citet{iki_berts_2022} fine-tune a text backbone for their text-based agent.
    \citet{song_visiontasker_2024} use pre-trained object detection and OCR models to convert screenshots into text-based observations for direct UI access actions.
    \citet{gur_real-world_2024} pre-train an LLM from scratch only on HTML data while utilizing an HTML-specific local and global attention mechanism.

\subsubsection{Environment Learning}\label{sec:environment_adaption}
Environment learning involves adaptation to computer environments through experience. Three main approaches are used: \emph{reinforcement learning}, \emph{behavioral cloning}, and \emph{long-term memory}. Among them, behavioral cloning is the most frequently used strategy (see Appendix~\Cref{fig:environment-adaption-dist}). 

Many foundation agents bypass the environment learning step, relying solely on their pre-trained, out-of-the-box capabilities by using prompting strategies. While these capabilities can be remarkably effective \citepeg{zheng_gpt-4vision_2024}, the absence of environment learning limits these agents, as they lack mechanisms to adapt or improve their performance within specific computer environments.

\paragraph{Reinforcement Learning} \label{sec:environment_adaption_rl}

In reinforcement learning (RL), an agent acts in an environment and learns to maximize a cumulative reward by trial and error \citep{sutton_reinforcement_2018}.
For computer use tasks, such environments are hand-crafted simulations, called \emph{controlled environments}, designed to mimic real-world computer settings while providing a reward signal for guidance.
RL has been implemented with various algorithms, including approximate policy iteration \citep{humphreys_data-driven_2022}, policy gradients \citep{shi_world_2017}, and bootstrapping with tree search \citep{shaw_pixels_2023}.

Agents in simpler environments may rely on brute-force exploration to learn directly from random behavior \citepeg{toyama_androidenv_2021, shvo_appbuddy_2021}.
However, in most computer environments, rewards are sparse as they are only given upon completing the assigned instruction $i$ \citepeg{shi_world_2017}, such as submitting a flight booking form after filling out \emph{all} details correctly.
Sparse rewards make learning from an initial random behavior often unsuccessful, as an agent is unlikely to predict a long action sequence by random chance \citep{humphreys_data-driven_2022}.
One strategy to mitigate sparse rewards is to begin by training an agent on human-labeled demonstrations (behavioral cloning), providing it with enough competence to start finding and learning from rewards \citepeg{shi_world_2017, humphreys_data-driven_2022}.
Relatedly, \citet{liu_reinforcement_2018} use human-labeled demonstrations to constrain the action space by defining sets of valid actions based on similarity to demonstrated actions, increasing the likelihood of reward discovery.

Without demonstrations, reward shaping \citep{ng1999policy} can artificially reduce sparsity by providing intermediate guidance, as shown by \citet{gur_learning_2019} and \citet{li_glider_2021}.
Alternatively, the task complexity can be adaptively adjusted. \citet{gur_environment_2021}, for instance, introduce a controlled environment that enables autonomous curriculum learning \citep{bengio2009curriculum} by automatically changing a task's complexity. Similarly, \citet{gur_learning_2019} employ curriculum learning by gradually moving an agent's starting point away from the goal state as it gains competence.

The key advantage of RL is its ability to autonomously explore environments and effectively navigate a dynamic dataset. However, RL's reliance on controlled environments limits its application to broad computer use tasks, as rewards must be defined and action consequences suppressed (i.e., ensure that actions within the simulation have no real-world consequences). The development of such environments can be very tedious, typically preventing current ACUs from acquiring broad knowledge by solely using RL.
Nevertheless, AndroidEnv \citep{toyama_androidenv_2021} combats this limitation by simulating a complete, virtual Android environment on top of which tasks can be configured by defining instructions and rewards.

\paragraph{Behavioral Cloning}\label{sec:environment_adaption_bc}

In behavioral cloning (BC) \citep{pomerleau_alvinn_1988}, an agent learns to mimic a shown behavior through supervised learning.
The shown behavior is usually a sequence of recorded observations and actions of a human completing a computer use task given an instruction \nolinebreak$i$.

Unlike RL, learning via BC does not require the agent to execute actions in the environment, making it applicable in \emph{uncontrolled} environments (trained based on observation-action pairs instead of a simulation). 
For example, \citet{zhang_you_2024} fine-tune a model on Android demonstrations from the work of \citet{rawles_android_2023}, while \citet{hong_cogagent_2024} combine Android demonstrations provided by \citet{rawles_android_2023} with Web demonstrations taken from the work of \citet{deng_mind2web_2023}.

BC methods vary in training strategies and data collection.
For example, \citet{gur_understanding_2023} train the entire model, \citet{hong_cogagent_2024} only update specific components, while \citet{li_effects_2024} use low-rank adaption \citep{hu_lora_2021} to fine-tune a foundation model.
Datasets are typically human-labeled \citepeg{humphreys_data-driven_2022}, but autonomous data collection methods also exist. For instance, \citet{furuta_multimodal_2024} use rejection sampling to identify successful trajectories from another agent's actions in a controlled environment, leveraging the environment’s rewards for validation. Similarly, \citet{lai_autowebglm_2024} iteratively collect successful demonstrations for improving their agent.

While BC can be used independently to train ACUs, it can also be used as a pre-trained step, with the ACU being fine-tuned afterward using RL to improve performance.
Typically, RL further enhances the agent by exploring aspects missing from the behavioral data. For instance, \citet{humphreys_data-driven_2022} demonstrate that after training their agent on $2.4$ million human-labeled actions, RL increases the task success rate from approximately $30\%$ to over $95\%$. 
Nonetheless, some agents rely entirely on BC, which can suffice for simpler tasks \citepeg{gur_understanding_2023}.

\paragraph{Long-Term Memory} \label{sec:environment_adaption_ltm}

\begin{figure}[b]
  \begin{subfigure}{0.49\linewidth}
    \includegraphics[width=\linewidth]{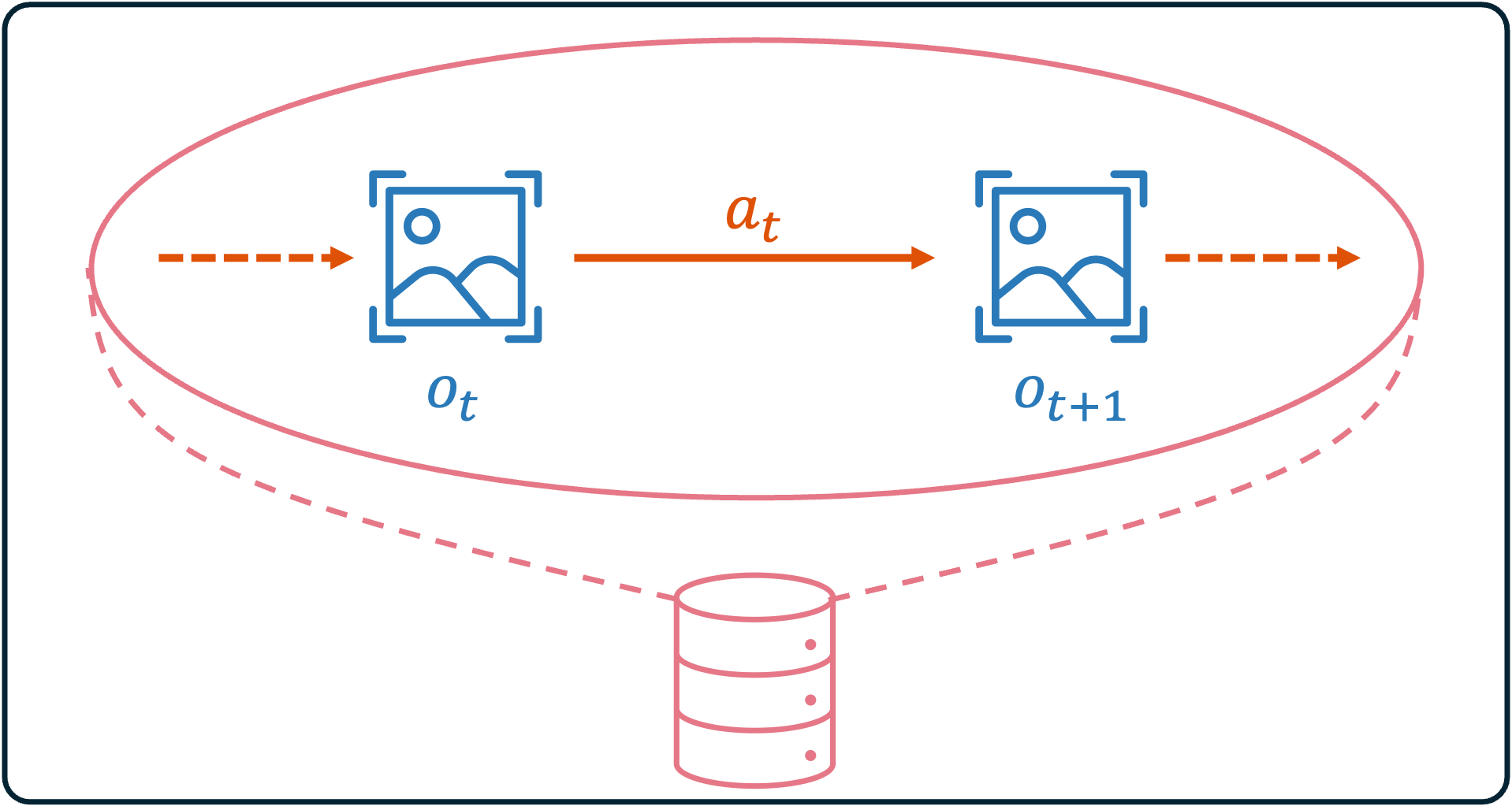}
    \caption{Memorize transitions $(o_t,a_t,o_{t+1})$}
    \label{fig:few_shot_what_env}
  \end{subfigure}
  \hfill
  \begin{subfigure}{0.49\linewidth}
    \includegraphics[width=\linewidth]{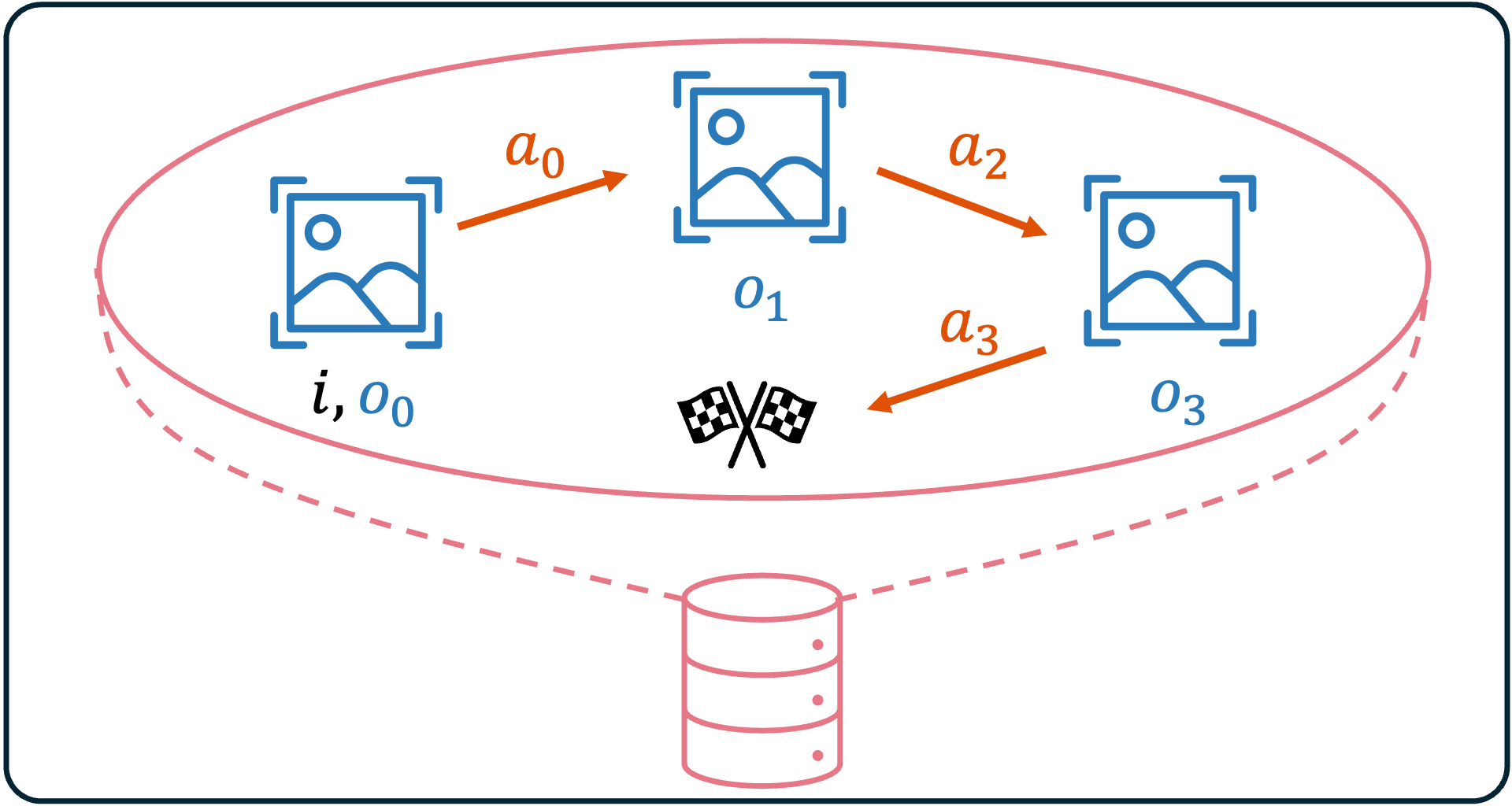}
    \caption{Memorize task demonstrations $(i,\tau)$}
    \label{fig:few_shot_what_act}
  \end{subfigure}
    \caption{Two kinds of experiences an agent can store in its long-term memory. (a) The agent memorizes the pre- and post-action observations as environment transitions $(o_t, a_t, o_{t+1})$. (b) The agent memorizes an instruction with a successful action-observation episode $(i, \tau)$}.
    \Description{This figure illustrates two conceptual approaches an agent can use to store experiences in long-term memory. The left panel shows storage of individual environment transitions as triplets consisting of the observation at time $t$, the action taken at time $t$, and the subsequent observation at time $t+1$. This emphasizes storing atomic transitions as discrete records. The right panel depicts storage of complete task trajectories, where an instruction is paired with the entire sequence of observations and actions from a successful episode. The figure uses a database symbol to represent the memory storage, highlighting that the left stores isolated transitions while the right stores full episodic trajectories.}
    \label{fig:few_shot_what}
\end{figure}

Foundation models exhibit strong few-shot learning capabilities \citep{brown_language_2020}, enabling foundation agents to enhance action prediction by incorporating successful demonstrations as examples directly into their context (see \cref{sec:episodic_improvement}).
This paradigm is also known as in-context learning (ICL).
Long-term memory extends ICL by first allowing the agent to execute and store successful trajectories in an external memory, and then later retrieve previous trajectories from this memory as examples of how a specific (sub-)task can be solved.
Importantly, since the examples are collected by the agent itself rather than provided by a human, the agent learns to improve its capabilities over time and adapts \emph{autonomously} to an environment (we discuss human prompt designs in the next section).
\cref{fig:few_shot_what} illustrates the two main types of experiences:

\begin{description}
    \item[Environment transitions:] 
        The agent memorizes environment transitions as triples ($o_t$, $a_t$, $o_{t+1}$), where $a_t$ represents the action taken, and $o_t,o_{t+1}$ capture the pre- and post-action observations, respectively. For example, the agent might store the consequence of its actions, such as ``clicking on the calculator app ($a_t$) on the home screen ($o_t$) opens the calculator app ($o_{t+1}$).'' \citet{wen_autodroid_2024} collect such transitions for Android apps in an offline phase by random exploration. They describe and summarize these transitions using an LLM, enabling the agent to enrich actionable elements with outcome information. For example, a \texttt{more options} button could be annotated to reveal specific hidden menu items, informing the agent what to expect if this button is clicked. Autonomous transition memories can also be combined with human demonstrations, as shown by \citet{zhang_appagent_2023} and \citet{li_appagent_2024}.
    \item[Task demonstrations:] 
        The agent memorizes task demonstrations by storing a tuple ($i$, $\tau_i$) containing the instruction $i$ and a successful demonstration $\tau_i=(o_{0},a_{0},\hdots,o_{t},a_{t})$ of solving $i$. Since only successful attempts are informative for the agent, the agent must have a mechanism to filter \emph{successful} trajectories. 
        A common approach is to use a controlled environment's feedback and only to store trajectories that yield a high reward \citepeg{tao_webwise_2023}.
        To manage memory constraints, trajectories $\tau_i$ are typically simplified to $\tau_i^*=(\tilde{o}_0,a_0,\hdots,\tilde{o}_t,a_t)$ (where $\tilde{o}_i$ is a simplification of $o_i$) before being stored. 
        For instance, \citet{deng_multi-turn_2024} only keep the actions $\tau_i^*=(a_0,\hdots,a_t)$, while \citet{sun_adaplanner_2023} store the complete executable program that solves $i$.
        These simplifications mirror history simplifications ($h_t \rightarrow h_t^*$), as the history $h_t$ is a (partial) trajectory.
        An alternative approach for discovering successful trajectories is programming by demonstration. Here, a human supervises the agent, intervenes if necessary, and demonstrates the correct solution for $i$, enabling online learning. \citet{song_visiontasker_2024} propose this method to summarize the corrected behavior for future retrieval.
\end{description}

A limitation of memory-based approaches is their reliance on storing specific instances rather than learning abstract generalizations. To mitigate this, \citet{lee_explore_2023} organize memories into a graph where observations are nodes, actions are edges, and both are generalized to unify related experiences. For example, an action $a=$ \texttt{click(text=Bob)} is generalized to $a^*=$ \texttt{click(text=[contact name])}. When retrieving memories, the graph is searched, and parameterized actions are instantiated based on the current state $(o_t, i)$, grounding parameters like \texttt{[contact name]} to specific values.
However, it still remains challenging to map specific trajectory instances to general concepts and to later retrieve and adapt helpful general concepts for specific tasks.

\subsubsection{Episodic Improvement}\label{sec:episodic_improvement}
Episodic improvement refers to an agent's ability to enhance its performance within a single episode by reasoning over its current context, without retaining knowledge across episodes. This effectively trades test-time computing for improved task execution.

Foundation agents commonly achieve episodic improvement through in-context learning \citep{brown_language_2020}.
ICL encompasses techniques such as \emph{instruction tuning}, where guidance is provided to the model through the prompt, and few-shot learning, which gives examples of successful trajectories as \emph{demonstrations} to the agent.

In contrast, current specialized agents typically do not employ episodic improvement. However, analogous mechanisms exist, such as search-based planning in game-playing agents, which simulate future outcomes to guide action selection \citep{silver_mastering_go_2017}. Such approaches can be considered as a more traditional reasoning within agents.

\paragraph{In-Context Learning through Instruction Tuning} \label{sec:episodic_improvement_instruction}

Foundation models are often adapted to specific tasks through \emph{prompt engineering}.
Typically, these prompts are designed by humans to adapt a foundation model to specific environmental conditions \citepeg{zheng_gpt-4vision_2024} and can include guidance on valid actions, previous history, assumed roles, or intermediate reasoning steps.
\Cref{tab:in-context-learning} exemplifies some snippets taken from the (much longer) prompts in the literature (for more details, refer to Table 6 in Zheng at al.~\citep{zheng_gpt-4vision_2024}). 

While most prompts are human-authored, some methods automate prompt construction.
For example, \citet{sun_adaplanner_2023} uses a second model as a planner to autonomously generate prompts for the agent.
This strategy, known as \emph{self-prompting}, involves using multiple instances of the foundation model, each fulfilling different roles and interacting with one another through iterative prompting \citepeg{song_mmac-copilot_2024}.

With the rise of vision-language models, \emph{visual prompt engineering} has emerged. This includes techniques such as extending screenshots to incorporate user instructions \citep{lee_pix2struct_2023}, overlaying bounding boxes on actionable UI elements \citepeg{bonatti_windows_2024}, and adding unique identifiers for visual grounding \citepeg{zhang_ufo_2024}.

\begin{table}

    \caption{Example snippets of actual prompts.}
    \label{tab:in-context-learning}
     {\footnotesize
    \begin{tabular}{>{\raggedright\arraybackslash}p{3.5cm} >{\raggedright\arraybackslash}p{10cm}}
        \toprule
        \textbf{Category} & \textbf{Prompt Snippet} \\
        \midrule
        Action Generation & \texttt{[...] you can \textbf{click an object} by referring to its id, such as 'click id=..., [...]'} \citep{li_zero-shot_2023} \\ 
        Provide history $h_t$ &  \texttt{\textbf{Previous} Actions: \{PREVIOUS ACTIONS\}} \citep{zheng_gpt-4vision_2024} \\ 
        Prescribe a role & \texttt{Imagine that you are \textbf{imitating humans} doing web navigation [...]} \citep{zheng_gpt-4vision_2024} \\ 
        Elicit intermediate thoughts & \texttt{[...] \textbf{think about} what the current webpage is [...] analyze each step of the previous action history [...] based on your analysis [...] decide on the following action [...]} \citep{zheng_gpt-4vision_2024} \\ 
        Provide general guidelines & \texttt{To be successful [...] only issue a \textbf{valid action} [...] only issue \textbf{one action} [...]} \citep{zheng_gpt-4vision_2024} \\
        \bottomrule
    \end{tabular}
    }
\end{table}

\paragraph{In-Context Learning through Demonstrations}\label{sec:episodic_improvement_demostrations}

\begin{figure}[b]
    \centering
 \begin{subfigure}{0.49\linewidth}
    \centering
    \includegraphics[width=0.7\linewidth]{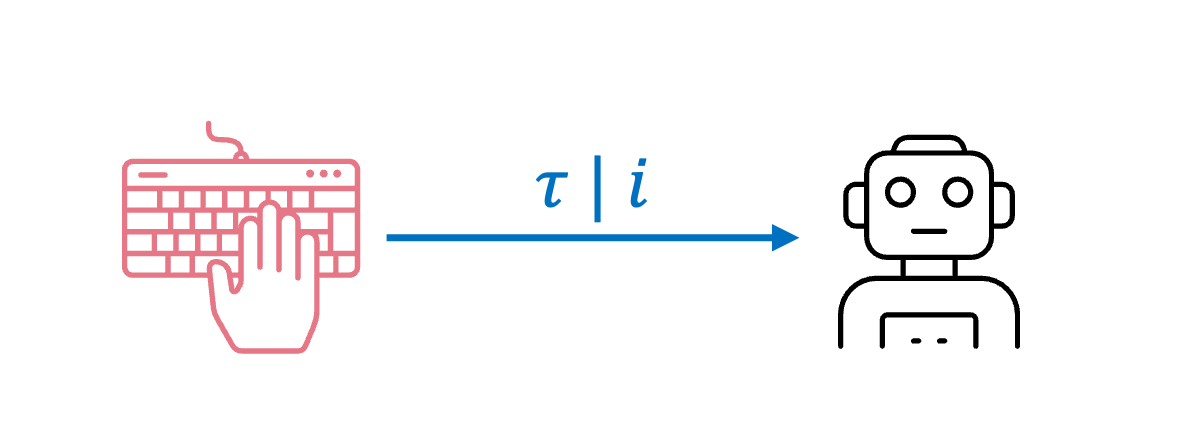}
    \caption{Human-crafted.}
    \label{fig:few_shot_1}
  \end{subfigure}
  \hfill
  \begin{subfigure}{0.49\linewidth}
    \centering
    \includegraphics[width=0.7\linewidth]{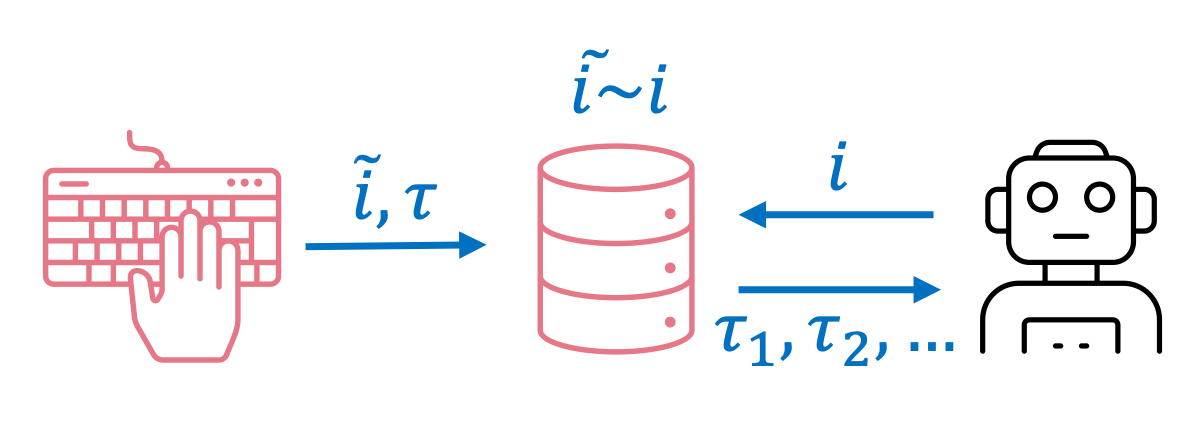}
    \caption{Semantic search.}
    \label{fig:few_shot_2}
  \end{subfigure}
  \hfill
  \begin{subfigure}{0.49\linewidth}
  \centering
    \includegraphics[width=0.7\linewidth]{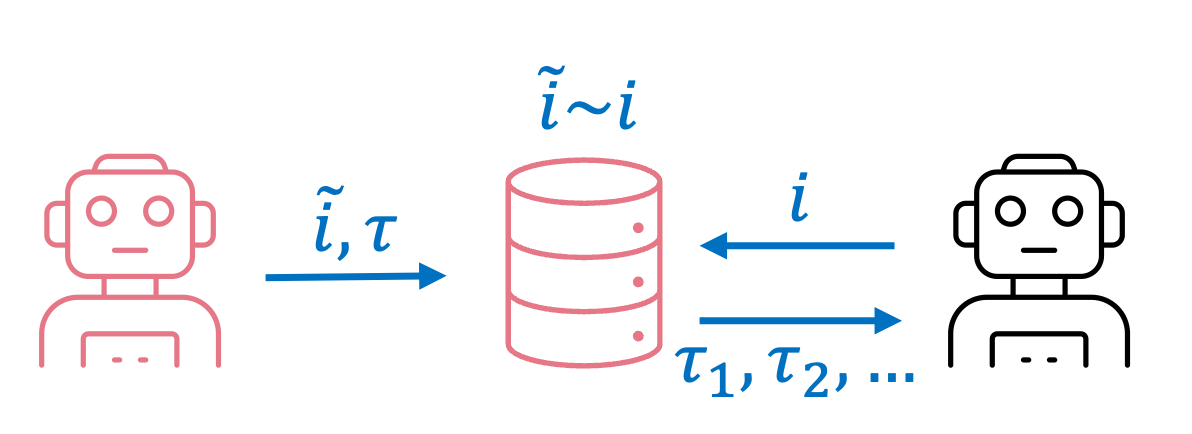}
    \caption{Auxiliary model.}
    \label{fig:few_shot_3}
  \end{subfigure}
  \hfill
  \begin{subfigure}{0.49\linewidth}
  \centering
    \includegraphics[width=0.7\linewidth]{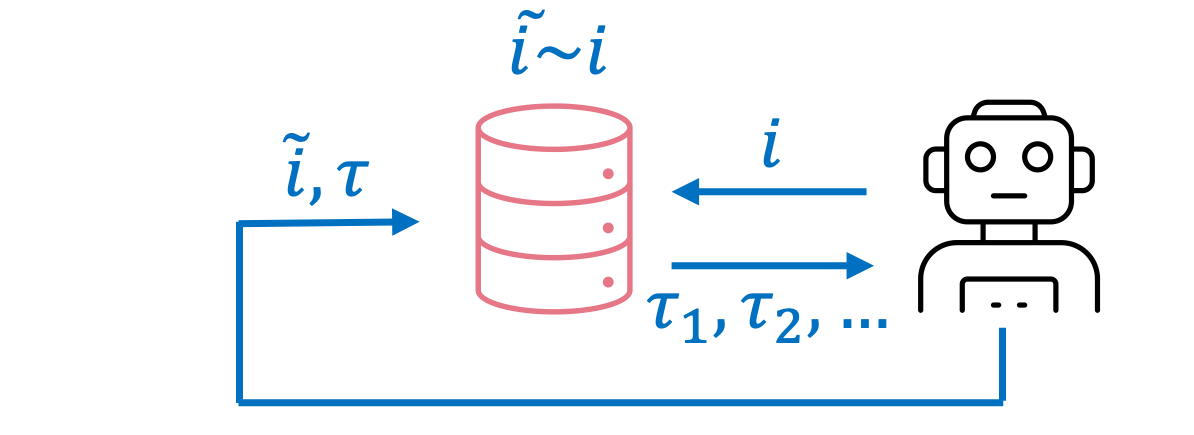}
    \caption{Agent collected.}
    \label{fig:few_shot_4}
  \end{subfigure}
    \caption{The four most common few-shot learning strategies for ACUs.}
    \Description{
    This figure presents four common few-shot learning strategies for ACUs, each visualized through schematic subfigures involving a human (keyboard icon), an ACU (robot icon), and a demonstration store (cylinder icon). Arrows denote the direction and type of information flow, such as instructions $i$ and task demonstrations $\tau$, highlighting distinctions in data provenance and retrieval mechanisms. In (a) \emph{human-crafted}, the human directly provides demonstrations $\tau$ conditioned on $i$ to the agent. In (b) \emph{semantic search}, the human contributes $\tau$ and $i$ to a database, which the agent then queries bidirectionally to retrieve demonstrations. In (c) \emph{auxiliary model}, an auxiliary agent, rather than a human, populates the database with $\tau$ and $i$, while the main agent retrieves data as in (b), visually distinguishing the source of data via the origin of the arrows. In (d) \emph{agent collected}, the ACU both generates and retrieves demonstrations, forming a closed-loop interaction absent in other strategies and not explicitly detailed in the main text. These diagrams emphasize that although the textual descriptions may appear similar, the strategies differ fundamentally in terms of data origin and control flow.
}
    \label{fig:few_shot}
\end{figure}

Few-shot learning enhances agent performance by providing example trajectories, ${\tau_1, \tau_2, \hdots}$, which demonstrate successful task execution.
\Cref{fig:few_shot} illustrates four common techniques for collecting and providing demonstrations to the agent. These common sourcing strategies include:

\begin{description}
    \item[Human-crafted:] 
        For a given class of tasks, a fixed set of human-crafted demonstrations $\{\tau_1, \tau_2, \hdots\}$ is provided to the foundation model \citepeg{kim_language_2023}.
    \item[Semantic retrieval:] 
        Based on the semantic similarity of the instruction $i$ compared to previous instructions, an agent retrieves human-crafted demonstrations from a database \citepeg{cho_caap_2024}.
    \item[Auxiliary model:] 
        A secondary agent is first used to generate a large set of demonstrations, after which the agent retrieves those demonstrations that are semantically relevant to the current instruction $i$.
    \item[Agent-collected:] 
        The agent autonomously collects its own demonstrations, referred to as \textit{long-term memory}, by searching through its past experiences (\cref{sec:environment_adaption_ltm}).
\end{description}

Given the limitations of context length, a provided trajectory $\tau = \bigl((o_0,a_0), (o_1, a_1), ... \bigr)$ is typically compressed $\tau \rightarrow \tau^*$,  analogous to history simplification ($h_t \rightarrow h_t^*$, \cref{sec:agent_policy_history}).

In addition to the trajectory, a demonstration may include rationales for each action taken \citepeg{cho_caap_2024}. These rationales, inspired by chain-of-thought prompting \citep{liu_chain_2023}, can aid the agent when making similar decisions. Such reasoning can be written by humans \citepeg{wang_enabling_2023} or generated autonomously by another model \citepeg{cho_caap_2024, sodhi_heap_2023}.

\paragraph{Episodic Improvement through Planning} \label{sec:episodic_planning}

ACUs are goal-driven and often require planning to fulfill complex instructions $i$ \citep[Chapter~2.4]{russell_artificial_2022}. Most specialized agents perform implicit planning in their latent space, a process \citet{li_zero-shot_2023} called \emph{iterative planning}, where future states or action consequences are not explicitly constructed.

Agents based on foundation models typically generate explicit plans in text form. One common method is \emph{chain-of-thought} prompting \citep{liu_chain_2023}, which guides the model to produce intermediate reasoning steps before deciding on an action, improving the agent's performance \citepeg{rawles_android_2023, zhang_android_2024}. Another method involves decomposing an instruction into sequential sub-tasks, such as breaking down the task \texttt{Book an economy class flight from Hangzhou to Beijing} into steps like \texttt{Open the Alipay app} and \texttt{Input ``Hangzhou'' as the departure city} \citep{guan_intelligent_2023}. 

Plans can be refined iteratively.
For instance, after initial prompting, agents may either follow their initial plan rigidly \citepeg{kim_language_2023} or adapt it based on new observations \citepeg{sun_adaplanner_2023}.
Another refinement \citep{kim_language_2023} is done by prompting their foundation model to critique and refine its generated plans recursively. Although this can yield minor improvements, \citep{kambhampati_can_2024} argues that the benefits of self-critiquing may be limited.

These prompt-based planning strategies are considered informal planning, as they are based on the text output of foundation models, as opposed to internally simulating various action trajectories before deciding on one.
In contrast, formal planning can be implemented based on a search algorithm.
For instance, \citet{koh_tree_2024} simulate actions in a controlled environment and search through potential future states (observations) to better inform decision-making for the next action.
This approach shows significant performance gains, with task success rates improving by 50\% at a search depth of $5$. 
Building on this, \citet{chae_web_2024} fine-tune a model to predict the effects of actions on current observations, allowing for better decision-making without relying on an external simulator.

\subsubsection{Recommendations}
The landscape of learning strategies for ACUs is notably diverse.
Historically, RL and BC dominated as the primary paradigms. More recently, the emergence of foundation agents has shifted attention toward prompt-based learning. Despite this evolution, our analysis reveals that the field has yet to converge on a unified framework (see Appendix~\cref{fig:learning-strategy-dist}).

To enable a technology-agnostic characterization, we categorized learning paradigms into three sequential steps: pre-training, environment learning, and episodic improvement.
Our analysis reveals \textbf{a research gap in an effective and practical environment learning paradigm for foundation agents}:
Long-term memory approaches, while practical, often suffer from poor generalization. Storing raw trajectories is less beneficial than capturing underlying concepts, which remains highly challenging within this approach.
In contrast, reinforcement learning (RL) and behavioral cloning (BC) provide strong learning signals and enable concept abstraction, but they are highly resource-intensive, requiring either high-fidelity simulation environments or curated and labeled datasets for effective fine-tuning.

To address this bottleneck, we recommend research into the direction of introducing a \textbf{self-supervised fine-tuning stage between general pre-training and resource-intensive environment learning}. This intermediate stage would align general-purpose foundation models more closely to computer use contexts --- analogous to the role of RLHF in aligning LLMs with human preferences \citep{ziegler_fine_2020} or GRPO in improving reasoning \citep{shao2024deepseekmathpushinglimitsmathematical}. Such an alignment stage would equip models with domain-specific inductive biases, enabling faster and more robust adaptation during subsequent environment learning phases \citep{ouyang_training_2022}.

Our analysis also identifies planning as a major limitation in current ACU architectures. 
LLMs exhibit limited long-horizon planning capabilities \citep{valmeekam2023planning}, and the dynamics of the environment are often unknown, which hinders direct adaptation of symbolic approaches.
Thus, we argue that \textbf{planning in ACUs remains an open and pressing research challenge}. However, we identify two promising research directions for addressing this gap:
First, recent developments in reasoning-oriented LLMs, such as OpenAI o1, demonstrate promising capabilities in planning and long-horizon decision-making \citep{valmeekam2024llms, tan_towards_2024}. 
Adapting these capabilities for ACUs—and demonstrating robust planning performance in dynamic digital environments—is a critical next step.
Second, hybrid systems that combine symbolic planning with learned models of perception and action outcomes, such as those proposed by \citet{koh_tree_2024}, offer a compelling alternative. These methods draw from classical planning algorithms \citep[Chapter~11]{russell_artificial_2022} while leveraging neural components for generalization and flexibility.
Although these approaches extend beyond the ACU domain, we argue that ACUs provide an ideal testbed due to their complexity and fully digital nature. The integration of neuro-symbolic methods with agentic foundation models may pave the way for more sophisticated, adaptive, and general-purpose computer use agents.

\section{Computer Use Datasets}\label{sec:datasets}
In this section, we focus on important computer use datasets and do not cover datasets used for general pre-training of foundation models or those only partially relevant for computer use, such as question answering \citepeg{hudson2019gqa} and tool usage datasets \citepeg{patil2023gorilla}.
\Cref{tab:overview_datasets} provides an overview of all considered computer use datasets and their key properties.

\begin{figure}[t]
    \centering
    \includegraphics[width=0.7\linewidth]{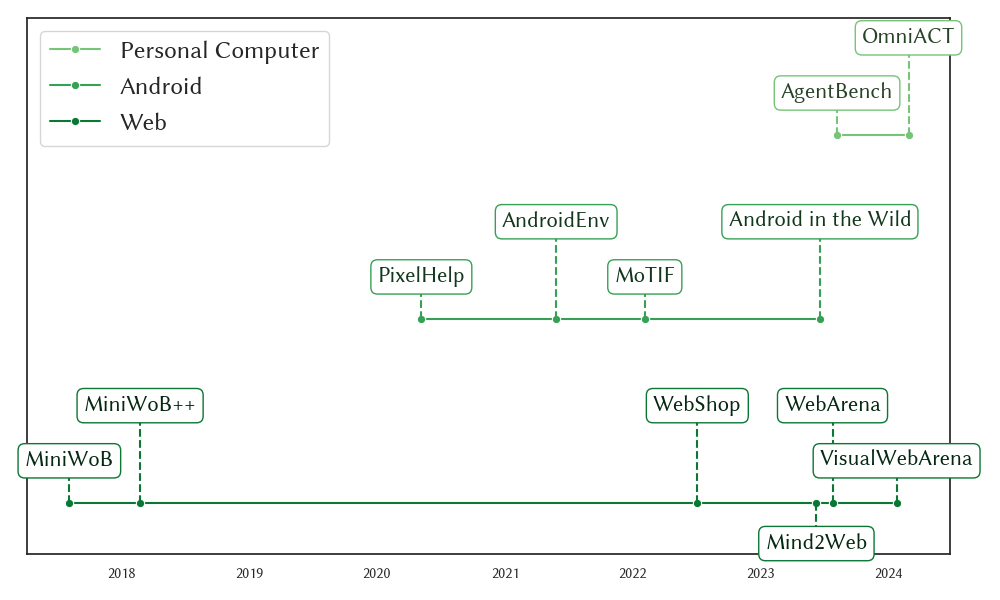}
    \caption{Development of datasets across domains over time. In general, complexity increases over time. For the \emph{Web domain}: MiniWoB \citep{shi_world_2017} and MiniWoB++ \citep{liu_reinforcement_2018} contain $100$ tasks on a simplified UI. WebShop \citep{yao_webshop_2022} is a more realistic single webshop application focusing on realistic product diversity. Mind2Web \citep{deng_mind2web_2023} contains $2350$ demonstrations across $137$ actual websites. WebArena \citep{zhou_webarena_2024} is a controlled environment of $4$ realistic web applications. VisualWebArena \citep{koh_visualwebarena_2024} extends WebArena with $910$ more visual tasks and an additional application. For the \emph{Android domain}: PixelHelp \citep{li_mapping_2020} contains step-by-step instructions across $4$ applications. AndroidEnv \citep{toyama_androidenv_2021} provides a framework to define custom tasks in Android applications. MoTIF \citep{avidan_dataset_2022} contains $756$ demonstrations across 125 applications. Android in the Wild \citep{rawles_android_2023} provides over $700,000$ demonstrations across $357$ applications. For the \emph{personal computer domain}: AgentBench \citep{liu_agentbench_2023-1} is a benchmark framework that spans operating systems, databases, web, and gaming tasks, including existing benchmarks like WebShop or Mind2Web. OmniACT \citep{kapoor_omniact_2024} contains $9802$ tasks labeled with straight-line code actions spanning $57$ applications on Windows, MacOS, Linux, and the Web.}
    \Description{The figure shows the chronological order of dataset releases for Web, Android, and Personal Computer domains over the years 2017 to 2024. It reveals that the Web domain saw earlier development activity, starting in 2017, while large-scale Android and Personal Computer datasets began emerging around 2020 and 2023, respectively. The timeline also indicates a concentration of new datasets in 2023 and 2024 across all domains, suggesting recent intensified interest. Although dataset complexity is not visually encoded, the temporal arrangement highlights the evolving landscape of task diversity and domain coverage over time.}
    \label{fig:dataset-timeline}
\end{figure}

To illustrate the evolution of these datasets, \Cref{fig:dataset-timeline} presents a timeline of their development across three major domains: Web, Android, and personal computers. The figure reflects a general trend toward increasing task complexity and realism over time, highlighting how research has shifted from simplified environments toward real-world applications and large-scale demonstrations.

\subsection{Dataset Types}

ACUs leverage two types of computer use datasets:

\begin{description}
    \item[Controlled Environments:] 
    A controlled environment is a simulated setting, meaning an agent can act freely without consequences, as the simulation can always be reset.
    These environments support reinforcement learning, given they provide an additional reward signal  \citepeg{humphreys_data-driven_2022}.
    Furthermore, they can be utilized to collect long-term memories in a safe simulation phase \citepeg{wen_autodroid_2024} and to plan at inference time by simulating potential actions \citep{koh_tree_2024}.
    \item[Offline Dataset: ]
    An offline dataset is collected by instructing humans on a computer task while recording observations and executed actions.
    The agent only sees the recorded interaction during training, meaning it never acts in the underlying environment, making training safe from consequences.
    Offline datasets can be utilized for few-shot learning \citepeg{deng_mind2web_2023} or fine-tuning an agent \citepeg{rahman_v-zen_2024} in an uncontrolled environment like a productive website. Furthermore, an offline dataset of a controlled environment can be used for initial behavioral cloning to combat sparse rewards \citep{humphreys_data-driven_2022}.
\end{description}

Both dataset types have distinct characteristics. 
Controlled environments are costly to create because they involve engineering simulations that mimic real-world behaviors, but the agent can explore all aspects of the environment autonomously.
In contrast, offline datasets can be recorded in any environment, but are incomplete as not every possible interaction is captured.
Furthermore, offline datasets only show a single trajectory to achieve an instruction, but maybe multiple ones exist.

\subsection{Domains, Observation and Action Spaces}
By analyzing the domains of our $\numdatasets$ reviewed datasets, we find that the majority of existing datasets are from the Web domain \citepeg{zhou_webarena_2024} and Android domain \citepeg{rawles_android_2023}, while the personal computer domain \citepeg{hong_cogagent_2024} receives less attention (see Appendix~\cref{fig:dataset-domain-dist}).

The types of observations and actions available in these datasets vary depending on the domain and data collection method. For observations, some datasets provide only image screen representations \citepeg{rawles_android_2023}, some only textual screen representations \citepeg{pasupat_mapping_2018}, while others offer both \citepeg{chen_websrc_2021}. Regarding actions, some datasets focus solely on mouse/touch and keyboard actions \citepeg{kapoor_omniact_2024}, some provide direct UI actions \citepeg{chen_webvln_2024}, while others focus on task-tailored actions \citepeg{liu_agentbench_2023}.
\Cref{tab:overview_datasets} provides an overview.
%
In many cases, additional observation and action types can be generated through post-processing efforts. For instance, HTML representations can be rendered through a web browser to provide image-based screen representations.

\subsection{Dataset Complexity}
Several factors, including the size of the state, observation, and action spaces, and the diversity of the tasks, influence the complexity of a computer use dataset. 

Controlled environments are often simplified and less diverse compared to offline datasets. 
For instance, in MiniWoB++ \citep{shi_world_2017}, all tasks are performed within a uniform, simplified website design with minimal graphical user interface (GUI) elements and clean HTML. 
Similarly, WebShop \citep{yao_webshop_2022} is limited to a single, simplified webshop application. 
While WebArena \citep{zhou_webarena_2024} offers more realistic web environments, it is limited to four tasks.

Offline datasets tend to feature more realistic observations, with the diversity depending on the variety of scenarios, such as how many websites were included.
For example, Mind2Web \citep{deng_mind2web_2023} records tasks from 137 websites across 31 categories, providing substantial diversity.
Similarly, Android in the Wild \citep{zhang_android_2024} records tasks spanning 357 Android apps or websites.

The complexity of tasks varies greatly across datasets. 
For example, MiniWoB++ \citep{shi_world_2017} includes 100 tasks with randomized text and an average of 3.6 actions per task, ranging from simple actions like clicking a button to more complex tasks like filling out a form to book a flight. 
WebShop \citep{yao_webshop_2022} offers 12,000 crowd-sourced instructions, all related to shopping, with an average of 11.3 actions per task. 
Mind2Web \citep{deng_mind2web_2023} provides 2,000 tasks averaging 7.3 actions, while WebArena \citep{zhou_webarena_2024} features 812 tasks, some requiring actions across applications, such as the task to create a Reddit account mirroring a GitLab profile.

Generally, the complexity of newer datasets increases as agents become more capable.
A straightforward way to do this is to make observations and tasks more diverse and challenging.
For example, WebArena \citep{zhou_webarena_2024} has a more realistic observation space than MiniWoB++ \citep{shi_world_2017}, and tasks require more actions to be achieved.
However, there are many other ways to increase complexity:
VisualWebArena \citep{koh_visualwebarena_2024} adds images as part of the instruction, such as asking an agent to create a post selling a product shown in an image.
AgentStudio \citep{zheng_agentstudio_2024} provides video-based observations, requiring agents to process dynamic, time-dependent information. 
MT-Mind2Web \citep{deng_multi-turn_2024} extends Mind2Web by introducing multi-turn tasks, where users give sequential instructions to the agent, requiring a more nuanced agent behavior.
MoTIF \citep{avidan_dataset_2022} introduces infeasible instructions in its offline dataset, challenging agents to recognize unachievable tasks.

\subsection{Recommendations}

A key limitation of current datasets lies in their insufficient trajectory complexity. They have limited structural and causal dependencies between actions in a task sequence. Complex tasks often require agents to execute actions in a specific causal order, where later actions depend on the outcomes of earlier ones. For instance, proposing a meeting time requires first retrieving the user's availability. In contrast, filling form fields can often be performed in any sequence. We recommend that \textbf{future ACU datasets increase trajectory complexity} by designing tasks that require longer, causally dependent action sequences.

Although trajectory complexity is crucial, observation, action, and task diversity should not be neglected, as broader observation and action types (e.g., dealing with various UI components) and more diverse tasks within and across applications are essential for accurately reflecting real-world usage and improving the generality of agents.

\section{Agent Evaluation}\label{sec:evaluation_agent}
\begin{figure}
    \centering
    \includegraphics[width=0.6\linewidth]{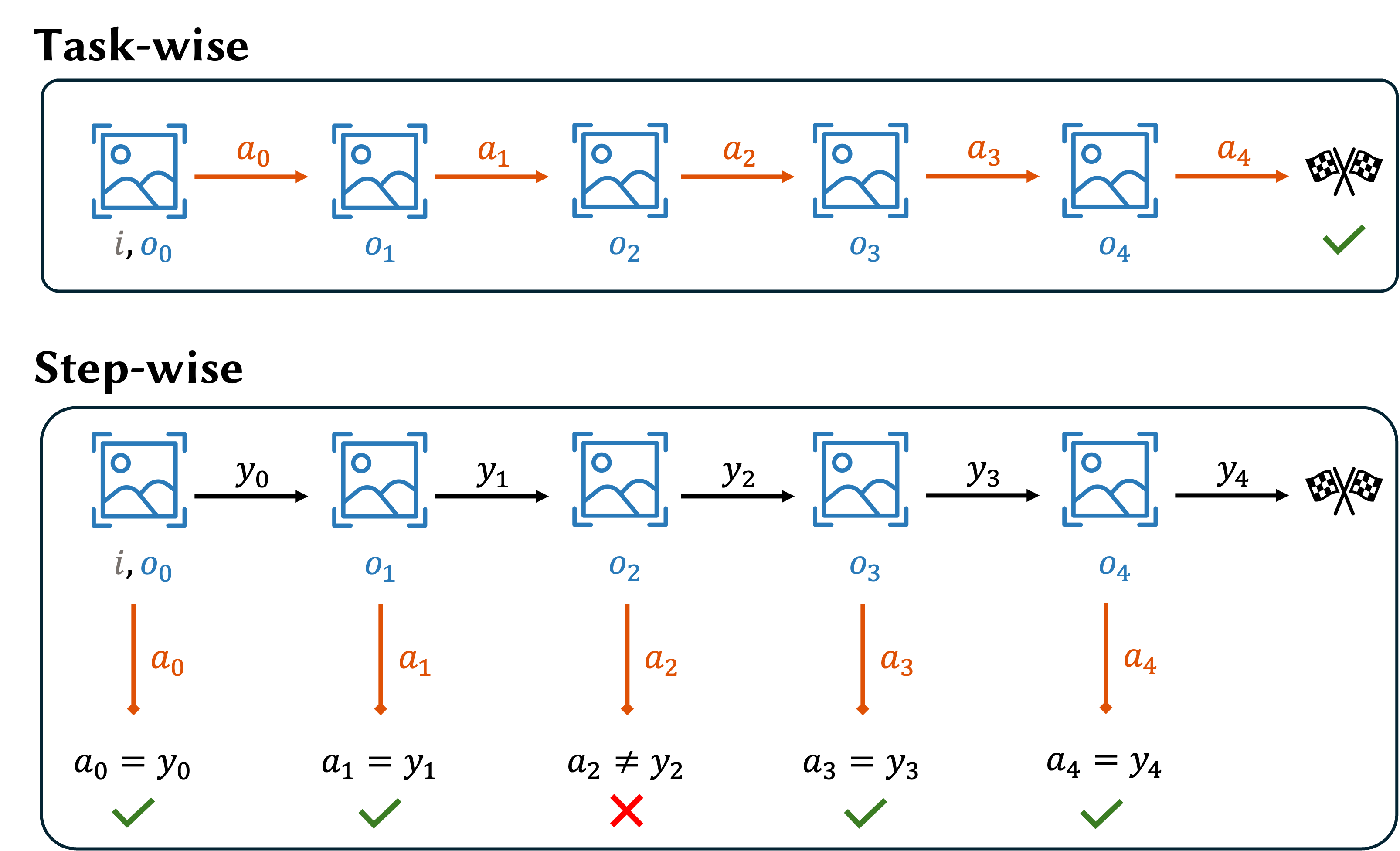}
    \caption{Task-level metrics measure performance across individual tasks (instructions). Step-level metrics measure performance across individual steps (actions).}
    \Description{The figure contrasts two evaluation approaches. In both cases, a sequence of four generic observations and a single instruction is shown. In the task-wise evaluation (upper part), transitions between observations result from the model's predicted actions, and evaluation is performed only at the final observation. In the step-wise evaluation (lower part), transitions follow ground-truth actions, and the performance metric is computed after each individual prediction. A green checkmark visually indicates when evaluation occurs. Specific content of observations and instructions is not depicted, as they serve only to illustrate the evaluation paradigms.}
    \label{fig:evaluation_types}
\end{figure}

Various evaluation metrics are used in the current literature. 
We identify three groups of evaluation metrics (see also \cref{fig:evaluation_types}): \emph{Task-level metrics}, \emph{step-level metrics}, and \emph{other metrics}.  

\subsection{Task-Level Metrics}
Task-level metrics focus on the overall effectiveness of an agent in achieving an instruction $i$.
\emph{Task success rate} is the most common task-level metric, which measures the overall success rate of completing an entire task \citep{deng_mind2web_2023,zhang_android_2024}.
For controlled environments, the environment state indicates successful task completion. 
For offline datasets, an agent predicting the full trajectory correctly counts as successful task completion, termed \emph{offline task success rate} (also called complete match \citepeg{li_mapping_2020}).

The offline task success rate underestimates the actual task success rate, as it only considers a single recorded trajectory, whereas alternative valid trajectories may exist. Consequently, it serves as a \emph{lower bound} on the true task success rate.
To obtain a more accurate estimate, the \emph{online task success rate} can be used. To measure online task success rate, the agent must be deployed in its original environment, typically the live websites from which the offline dataset was collected. 
Human evaluators then determine whether the agent successfully completes the task \citep{zheng_gpt-4vision_2024, song_restgpt_2023, li_sugilite_2017}. Notably, \citet{zheng_gpt-4vision_2024} report that their agent's success rate increased from 12\% to 36\% when evaluated online, highlighting the limitations of relying solely on offline trajectories.
However, the reproducibility of the online task success rate poses a challenge, due to potential changes in the online environment and the potential for error in human evaluation \citep{reason1990human}.

Other, less common task-level metrics exist, often providing a more nuanced assessment of the agent's capabilities.
\emph{Task progress} measures the average task completion progress, meaning how far the agent, on average, is to complete a task \citepeg{sodhi_heap_2023, zhang_android_2024}.
\emph{Average reward} captures the average reward obtained across episodes within a controlled environment \citepeg{jia_dom-q-net_2019}. 

\subsection{Step-Level Metrics}

Step-level metrics focus on the overall effectiveness of an agent in predicting actions (steps) across tasks.
\emph{Step success rate} is the most common step-level metric, which assesses the accuracy of action prediction \citepeg{deng_mind2web_2023}.
In the literature, step success rate is also called \emph{partial match} \citepeg{li_mapping_2020} or \emph{action accuracy} \citepeg{wen_autodroid_2024}.

Each step (action) is part of a trajectory (a sequence of multiple actions), which in turn represents a single task in the dataset (comprising multiple tasks). Consequently, step-level metrics must define how to average step scores both within their trajectory and across tasks, similar to other fields like multi-class classification, where metrics are averaged within classes and across samples \citep{grandini2020metrics}.
Two natural approaches for averaging exist:

\begin{description}
    \item[Macro averaging:] 
        Step scores are averaged first within their respective trajectory and then across tasks. As a result, each step score is weighted by the inverse of its corresponding trajectory's length.
    \item[Micro averaging:] 
        Step scores are averaged across all steps (of all trajectories). This assigns equal weight to each step score regardless of trajectory length.
\end{description}

For computer use, \emph{macro averaging} seems to be the prevailing approach, established by Mind2Web \citep{deng_mind2web_2023} and adopted by subsequent work \citepeg{zheng_synapse_2024}.

Other less common step-level metrics include the \emph{action F1} score \citepeg{li_mug_2024}, \emph{action recall} \citepeg{li_glider_2021}, or measuring only if parts of the action are correct, like the \emph{element accuracy} for direct UI access actions \citepeg{deng_mind2web_2023}.
Finally, all step-level metrics only exist for offline datasets, as controlled environments' rewards do not indicate the correctness of individual actions.

\subsection{Other Metrics}

Other metrics in the literature measure performance indicators other than an agent's capabilities.
\citet{song_restgpt_2023} evaluate agent \emph{efficiency} by measuring the number of API calls required to execute an instruction successfully, emphasizing minimal resource usage during task execution.
\citet{zhang_ufo_2024} incorporate a safeguard mechanism to seek user confirmation before executing critical actions (e.g., delete) to build a safer and more trustworthy agent. The \emph{safeguard rate} measures how accurately the agent identifies sensitive actions and requests user confirmation. 

\subsection{Recommendations}
A standardized evaluation protocol is currently absent in the ACU literature. A key challenge lies in the prevalent use of custom datasets or modified benchmarks to highlight specific strengths \citepeg{wang_mobile-agent_2024,wen_autodroid_2024}, which limits comparability across studies.
For instance, in the MiniWoB++ benchmark \citep{shi_world_2017,liu_reinforcement_2018}, different studies have adopted varying subsets of tasks, complicating cross-study comparisons. \citet{humphreys_data-driven_2022} evaluated agents on all 104\footnote{MiniWoB++ currently includes 100 tasks, excluding four that violate the static assumption.} tasks, while \citet{zheng_synapse_2024} used 64 tasks and \citet{kim_language_2023} selected 55. 
Despite these differences between ACU evaluations, average task performance is compared directly, undermining fair assessment. 
For example, \citet[Figure 3]{zheng_synapse_2024} and \citet[Figure 4 (b)]{kim_language_2023} compared their average task performance on a subset of tasks with an ACU evaluated on the full benchmark.

Moreover, no consensus exists on how to measure agent performance.
Among the available performance metrics, the task success rate best reflects practical agent capabilities, as it evaluates whether an agent completes a task in its entirety.
In contrast, step-level metrics measure the performance across individual steps and can be misleading when assessing actual agent competence, as a single error in a lengthy action trajectory may have substantial real-world consequences but only a minor influence on step-level metrics.
Accordingly, we recommend that for evaluating performance, \textbf{ACU publications should report the task success rate as the primary metric on established and complete benchmarks}.

While task success rate can be reliably measured in controlled environments, it is challenging for offline datasets. 
Reporting the offline task success rate provides only a lower bound and might underestimate agent performance; reporting the online task success rate is labor-intensive and suffers from limited reproducibility due to evolving online environment conditions and different evaluation procedures.
In such offline dataset settings, the step success rate can serve as a reproducible proxy that must be interpreted with caution:
First, it is a conservative estimate of actual step correctness, as alternative valid actions are often not captured in the dataset and penalized as errors.
Second, higher step-level accuracy does not necessarily translate to higher task-level performance.

For evaluations on offline dataset benchmarks, we recommend reporting both the step success rate as a reproducible proxy metric and, where possible, the online task success rate as a measure of actual agent performance. To improve comprehensibility and traceability, the online evaluation protocol should be thoroughly documented, including details such as the date of the evaluation and any strategies used to mitigate human error \citep{reason1990human}.

Finally, ACUs must be capable of predicting a deliberate \texttt{stop} action to signal task completion. This capability is critical both for practical deployment and for correctly identifying when a task has been completed \citepeg{wang_mobile-agent_2024}. We therefore recommend requiring a \texttt{stop} action in ACU evaluations whenever feasible and suggest that future benchmarks enforce this requirement. For example, controlled environments could reward agents only after they reach the goal state and explicitly issue the stop action.

%
\section{Conclusions}\label{sec:discussion_conclusion}
Agents for Computer Use (ACUs) represent a rapidly advancing frontier in AI, offering both significant research challenges and substantial practical impact.
While specialized designs remain viable for narrow, efficiency-critical tasks, the field is undergoing a paradigm shift toward foundation agents to enable the open-ended reasoning required for general computer use.
Despite the accelerated progress driven by foundation models, many core challenges remain unresolved.
This work identifies these challenges based on a unifying taxonomy that organizes ACU research across key concepts and establishes a shared vocabulary. Our taxonomy is structured around three complementary perspectives: The \emph{domain perspective}, which characterizes the computing environment; the \emph{interaction perspective}, which defines the observation and action spaces; and the \emph{agent perspective}, which concerns internal structure and learning dynamics. This framework bridges previously disconnected lines of work, from reinforcement learning to prompting-based agents, and provides a technology-agnostic basis for comparison and analysis.

By applying our taxonomy to $\numagents$ ACUs across $\numdatasets$ datasets, we uncover several fundamental limitations in the current landscape of ACU research. Specifically, we identify:
(1) reliance on structurally inconsistent input modalities that hinder generalization; 
(2) inefficient learning strategies; 
(3) limited capabilities in planning for executing complex, multi-step tasks successfully; 
(4) benchmarks that prioritize perception realism over task complexity; 
(5) inconsistent evaluation metrics that obstruct comparability; and 
(6) a disconnect between experimental assumptions and real-world deployment conditions. 

To overcome these limitations and advance the ACU field, we recommend:
(a) adopting image-based observation spaces to support consistent and robust perception;
(b) pursuing cost-efficient learning strategies that allow better scalability and adaptability;
(c) advancing policy architectures that support long-horizon reasoning and planning;
(d) constructing benchmarks that integrate both realistic perception and task complexity;
(e) standardizing evaluation metrics, especially success rates, to enable fair comparisons; and
(f) grounding research in realistic assumptions by closely examining deployment conditions.

While limitations (1)–(5) and recommendations (a)–(e) are discussed throughout the main text, the limitation (6) and the recommendation (f) concern real-world discrepancies that are not addressed in the current literature.
First, most ACU systems are built for idealized settings, assuming a deterministic, static, stationary, and episodic environment. 
However, real computing environments are dynamic, meaning agents must adapt their strategy based on changing perceptions due to other running processes, e.g., notifications obscuring the view. Furthermore, real-world environments are non-stationary, meaning an environment changes over time due to, e.g., application updates (see Appendix \cref{sec:domain_view:nature} for a detailed discussion on such environment discrepancies).
Second, there are unique privacy considerations: Traditional user education techniques fail, as users cannot control what an autonomous agent might observe and send to an ACU model provider (see Appendix \cref{sec:cp-technical-challenges} for a detailed discussion).
Third, safety considerations are systematically underexplored.
Current research focuses solely on full autonomy, but conditional autonomy can increase safety, such as an ACU handing back control to the user for critical decisions (see Appendix \cref{sec:safety} for a detailed discussion).

While our taxonomy offers a structured overview, it has several limitations. We did not include a comprehensive comparison of agent capabilities, as many ACUs support only a subset of benchmarks or tasks within benchmarks and report different metrics, making it infeasible. We also do not explore how specific design choices, such as the selection of a foundation model or RL algorithms, affect performance. The scope is limited to agents using text-based instructions. Although our taxonomy is compatible with dynamic observation-action loops (e.g., those involving video inputs), we focus on static interaction patterns and do not evaluate the taxonomy in dynamic settings such as AndroidWorld \citep{rawles_androidworld_2024}. Furthermore, we intentionally focus on fully disclosed contributions, which excludes some of the recent commercial systems. While we ground our framework in established concepts, some components, such as our definition of agent learning, are computer use specific; it remains open how the field will evolve with respect to them.

Nonetheless, our taxonomy and associated analysis offer a valuable foundation for organizing and advancing ACU research. By integrating diverse perspectives and identifying shared challenges, this work supports a more cohesive, forward-looking research agenda. We hope this work fosters the development of ACUs that are robust, adaptive, and ready for deployment in real-world computing environments, and that the structure and terminology we introduce bring coherence to this currently fragmented field.

\begin{acks}
The authors P. Sager and B. Meyer contributed equally to this work.
The work is funded in part by the Canton of Zurich, Switzerland, through the Digitalization Initiative of the Canton of Zurich (DIZH) Fellowship project `Stability of self-organizing net fragments as inductive bias for next-generation deep learning.'
\end{acks}


\printbibliography

\appendix

\newcommand{\T}{\checkmark}
\newcommand{\halfTT}{(\checkmark)}
\newcommand{\halfT}{\kern-4.5px(\checkmark)}
\renewcommand*{\arraystretch}{1.3} 
\section{Trends and distributions in ACU Literature}\label{sec:appendix_trends}

Based on the information that we collect from the reviewed literature (see details in \Cref{sec:appendix_tables}), we further analyze the statistics of important topics related to ACU. Specifically, we identify the frequencies, trends, and distributions to highlight key insights.
An interactive versions of the plots presented below are available on our project page at \url{https://sagerpascal.github.io/agents-for-computer-use}.

\begin{figure}[htbp]
    \centering
    \includegraphics[width=0.48\linewidth]{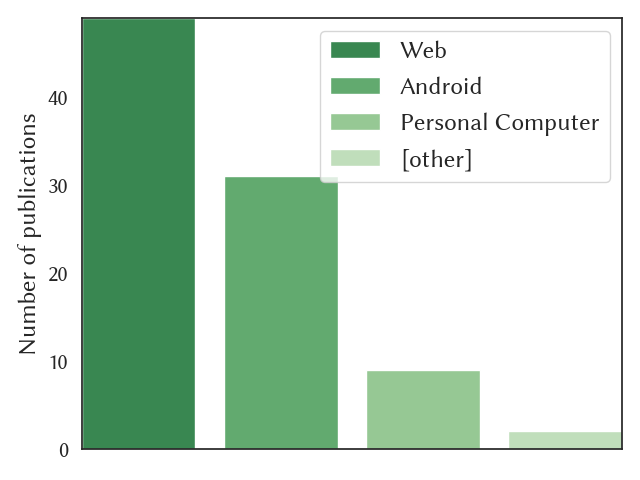}
    \caption{Number of ACU publications across domains, showing a strong preference for web-based platforms, followed by Android and personal computers.}
    \Description{A plot showing the number of agent publications with green bars. $49$ ACUs target web-based platforms, $31$ target Android, $9$ target personal computers, and $2$ target other domains. The figure highlights the concentration of research efforts on web and Android environments.}
    \label{fig:domain-dist}
\end{figure}

\begin{figure}[htbp]
    \centering
    \includegraphics[width=0.48\linewidth]{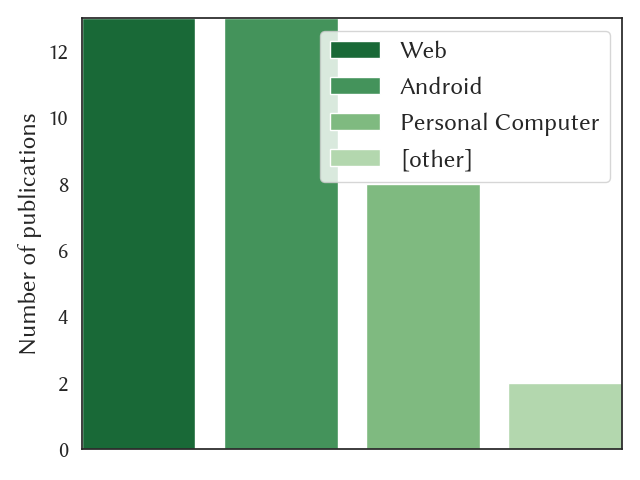}
    \caption{Dataset counts across domains. The distribution of datasets shows that most datasets analyzed in this study are from the Web and Android domains.}
    \Description{A plot showing the number of datasets publications with green bars. There are 13 datasets focusing on Web, $13$ datasets focusing on Android, $8$ datasets focusing on personal computers, and $2$ datasets focusing on other domains. This illustrates the alignment between dataset availability and research focus in ACU studies.}
    \label{fig:dataset-domain-dist}
\end{figure}

\begin{figure}[htbp]
    \centering
    \begin{minipage}[t]{0.48\textwidth}
        \centering
        \includegraphics[width=\linewidth]{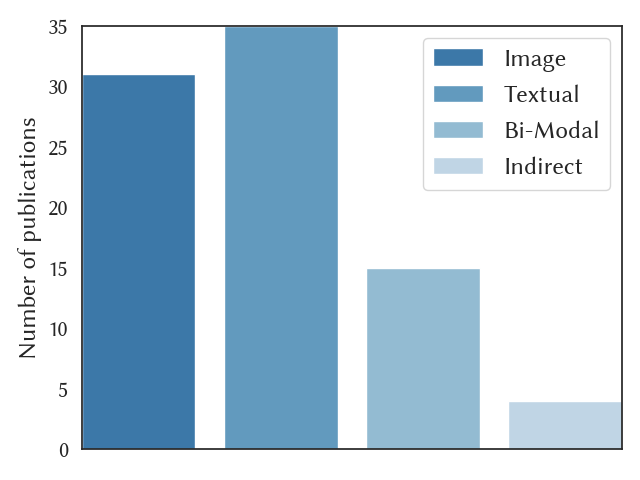}
    \end{minipage}
    \hfill
    \begin{minipage}[t]{0.48\textwidth}
        \centering
        \includegraphics[width=\linewidth]{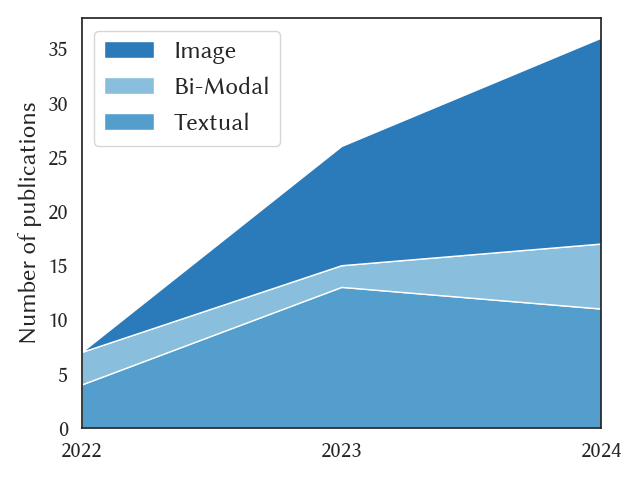}
    \end{minipage}
    \caption{(left) Frequencies over all years and (right) trends as a stacked area plot of publications of observation spaces. It shows rapid growth in image and bi-modal.}
    \Description{Two plots, both showing the number of publications for different observation spaces. The left subplot is a bar plot, showing that $35$ ACUs use textual observations, $31$ ACUs use image observations, $15$ ACUs use bi-modal observations, and $4$ ACUs use indirect observations. The right subplot is a stacked area plot illustrating trends in the number of publications using each modality from 2022 to 2024.  Image-based observation usage shows a pronounced increase over time, from $0$ in 2022 to $12$ in 2023 and $19$ in 2024. In contrast, textual and bi-modal modalities remain relatively stable, with textual at $3$, $10$, and $8$ publications, and bi-modal at $1$, $1$, and $2$ publications across the three years, respectively.}
    \label{fig:observation-space-trend}
\end{figure}

\begin{figure}[htbp]
    \centering
    \includegraphics[width=0.48\linewidth]{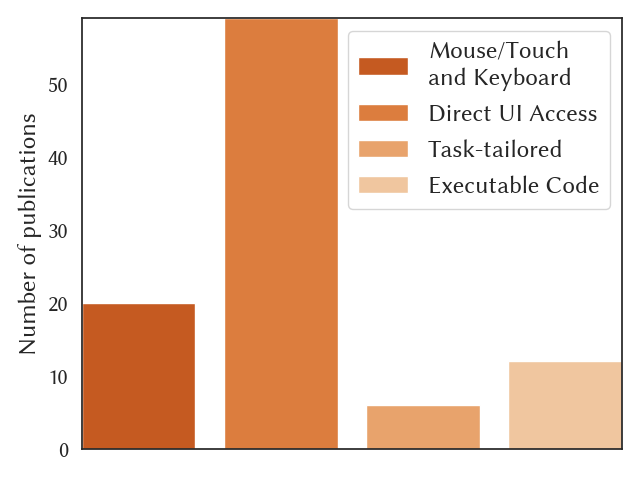}
    \caption{Frequencies of action spaces. Agents with multiple action types are counted once.}
    \Description{A bar plot showing the number of ACU publications per action space. The most common action type is direct user interface access ($59$ ACUs), followed by mouse/touch and keyboard actions ($20$ ACUs). Executable code ($12$ ACUs) and task-tailored actions ($6$ ACUs) are less frequent.}
    \label{fig:action-space-dist}
\end{figure}

\begin{figure}[htbp]
    \centering
    \includegraphics[width=0.48\linewidth]{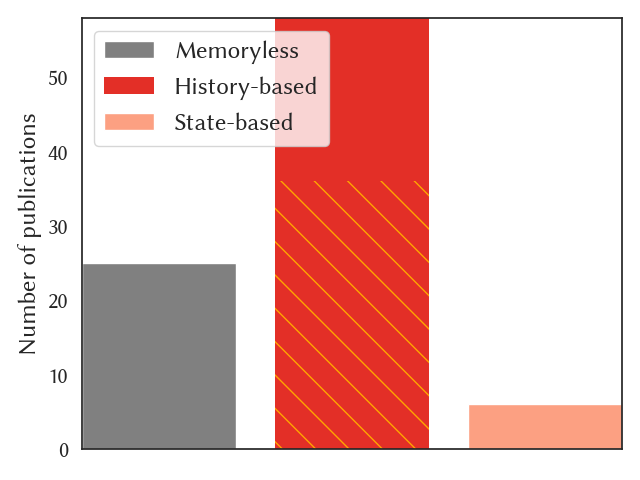}
    \caption{Frequencies of policy types. Orange strips indicate agents that only track past actions.}
    \Description{A bar plot showing the number of ACU publications for different policy types. There are $58$ ACUs using history-based policies (from which $36$ only track past actions), $25$ ACUs use memoryless policies, and $6$ ACUs use state-based policies.}
    \label{fig:policy-dist}
\end{figure}

\begin{figure}[htbp]
    \centering
    \includegraphics[width=0.48\linewidth]{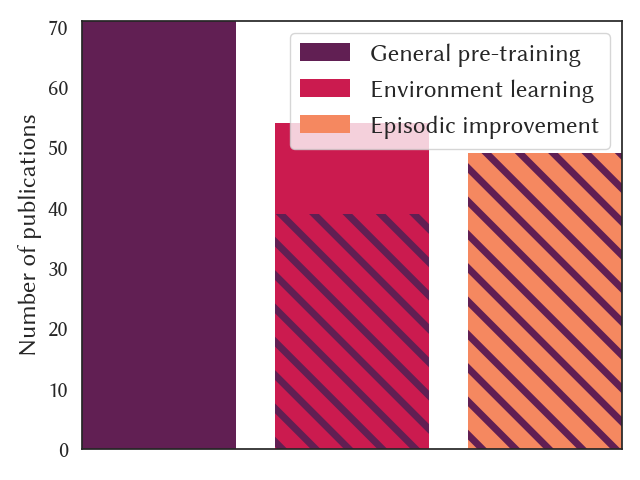}
    \caption{Frequencies of learning strategies. Purple stripes indicate initial pre-training.}
    \Description{A bar plot showing the number of ACU publications for different learning strategies. $70$ ACUs use general pre-training, $53$ ACUs use environment learning (from which $38$ are pre-trained), and $49$ ACUs use episodic improvement (from which all are pre-trained).}
    \label{fig:learning-strategy-dist}
\end{figure}

\begin{figure}[htbp]
    \centering
    \begin{minipage}[t]{0.48\textwidth}
        \centering
        \includegraphics[width=1.0\linewidth]{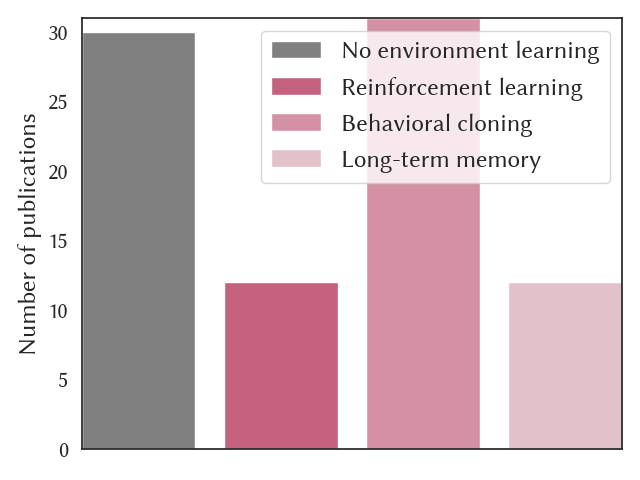}
    \end{minipage}%
    \hfill
    \begin{minipage}[t]{0.48\textwidth}
        \centering
        \includegraphics[width=1.0\linewidth]{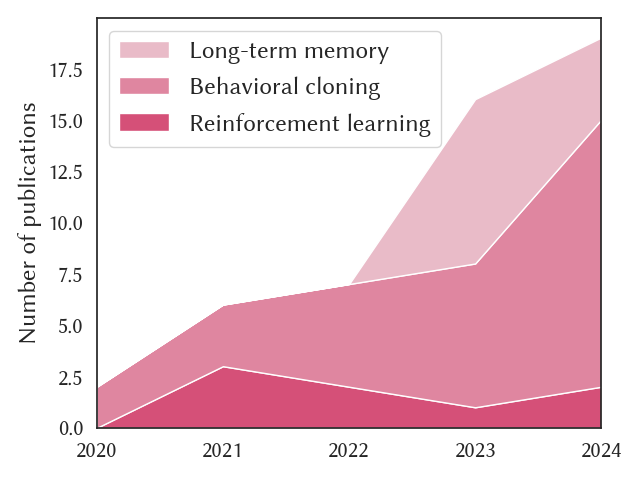}
    \end{minipage}
    \caption{
    (Left) Frequencies over all years and (right) usage trends as staked area plot of environment learning strategies over the last 5 years. It shows that behavioral cloning is the most adopted method.
    }
    \Description{Two plots, both showing the number of publications for different environment learning strategies. The left plot is a bar plot, showing that 31 ACUs use behavior cloning, 30 ACUs use no environment learning, 12 ACUs use reinforcement learning, and 12 ACUs use long-term memory.
    The right panel is a stacked area plot showing yearly trends from 2020 to 2024. Behavioral cloning has increased steadily, from $2$ publications in 2020, to $3$ in 2021, to $5$ in 2022, to $7$ in 2023, and to $12$ in 2024.
    The number of reinforcement learning publications is rather constant over time ($0$ publications in 2020, $3$ in 2021, $2$ in 2022, $1$ in 2023, and $2$ in 2024).
    Long-term memory appears for the first time in 2023 and is directly used by $8$ ACUs, while $4$ agents used it in 2024.
    }
    \label{fig:environment-adaption-dist}
\end{figure}

\section{Image vs. Textual Screen Representation} \label{sec:img_vs_text}
Most ACUs either use image or text observations, or a combination of them.
In the following, we provide a comparison between image and textual screen observations.
In \cref{fig:text-repr-advantages}, we illustrate that textual screen representations offer unique strengths, particularly in exposing hidden semantics and structural relationships.
However, these strengths are often undermined by practical drawbacks (see \cref{fig:text-repr-disadvantages}), such as verbosity and inconsistency, especially when deployed in real-world environments.
These findings are in line with the works of \cite{he_actionbert_2021,wang_enabling_2023,li_zero-shot_2023,zheng_gpt-4vision_2024,cheng_seeclick_2024} and can be summarized as follows:
\begin{figure}[htbp]
\centering
\includegraphics[width=0.6\linewidth]{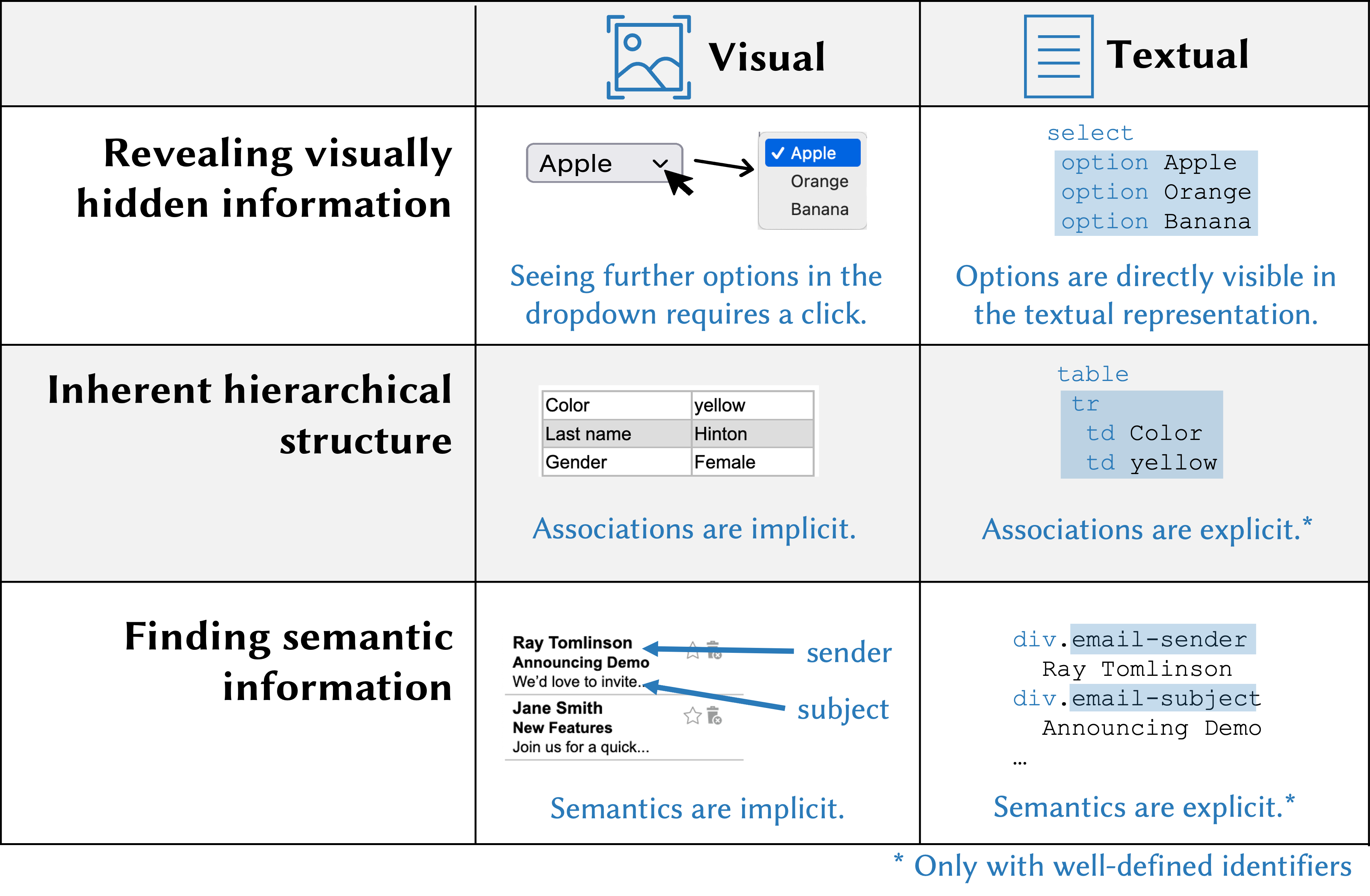}
\caption{Advantages of textual screen representations.}
\Description{
The figure compares visual and textual interfaces across three axes. 
1. Revealing visually hidden information: Textual representation displays all dropdown options directly in code, unlike the visual interface, which requires user interaction. 
2. Inherent hierarchical structure: Textual HTML code explicitly defines structure using tags like <table>, <tr>, and <td>, while visual layouts only imply structure. 
3. Finding semantic information: Textual markup contains identifiers like ``email-sender'' and ``email-subject'', whereas visual cues are inferred from layout and formatting.
}
\label{fig:text-repr-advantages}
\end{figure}
\begin{figure}[htbp]
\centering
\includegraphics[width=0.6\linewidth]{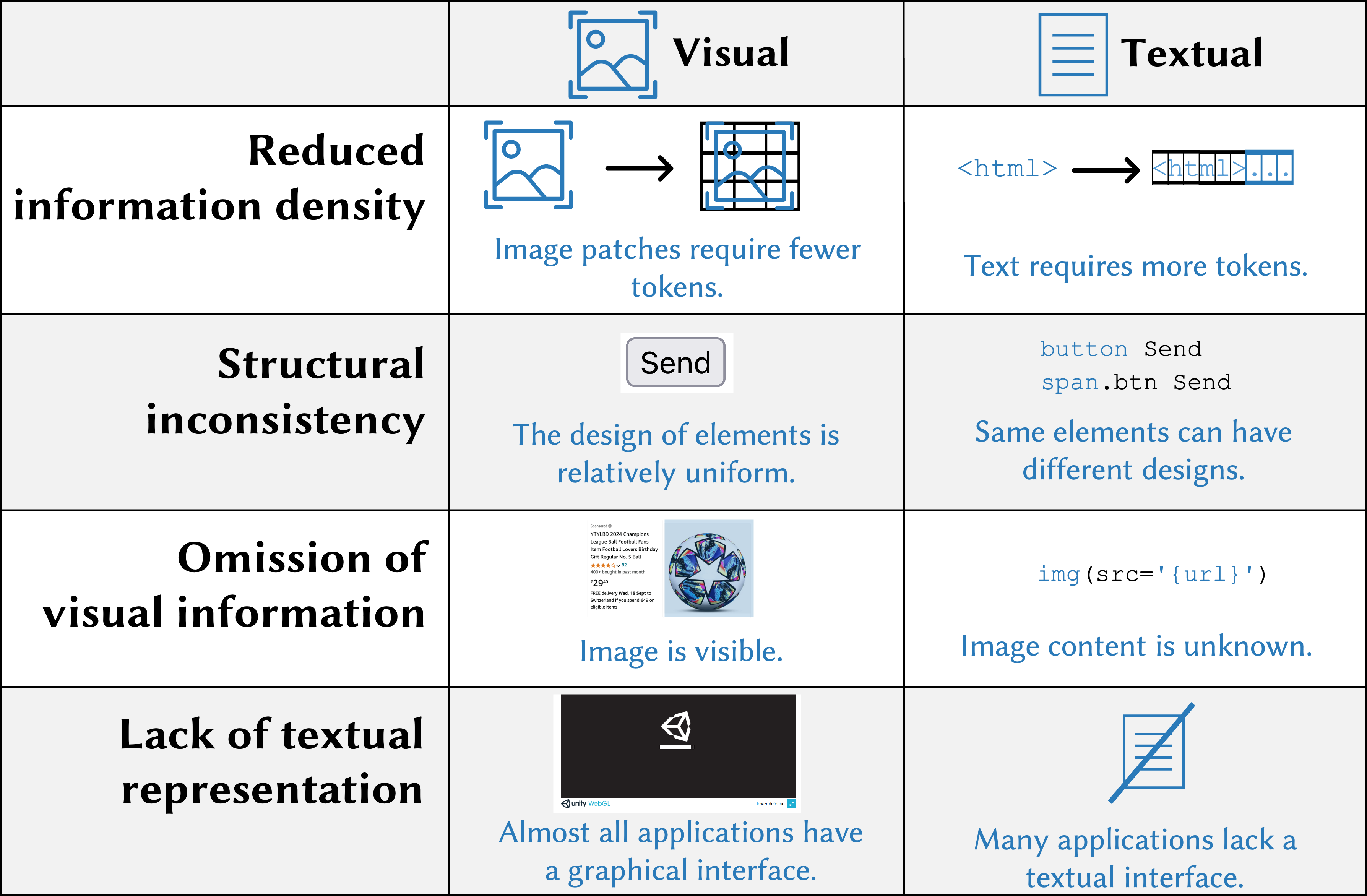}
\caption{Disadvantages of textual screen representations.}
\Description{
The figure compares limitations of visual and textual representations across four categories.
1. Reduced information density: Visual image patches can represent more with fewer tokens than textual equivalents.
2. Structural inconsistency: In HTML, the same function (e.g., a button) may be represented differently across applications, whereas in visuals, the rendering is relatively uniform.
3. Omission of visual information: Textual code like img(src='{url}') lacks embedded image content, while visuals show actual images.
4. Lack of textual representation: Some applications only have visual interfaces with no textual HTML representation.
}
\label{fig:text-repr-disadvantages}
\end{figure}
\begin{description}
    \item[Advantages] of textual screen representations are: \hfill
    \begin{description}
        \item[Revealing visually hidden information:] 
            Textual representations can explicitly show information that may be visually hidden in images, such as items within a collapsed drop-down menu.
        \item[Inherent hierarchical structure:] 
            Textual representations, like the Document Object Model (DOM) tree, are structured in a hierarchical tree, facilitating a clearer understanding of relationships between elements \citepeg{jia_dom-q-net_2019}.
        \item[Explicit semantic information:] 
            Textual representations often include semantic information in element attributes that are not visible in images, such as \texttt{id} tags. For example, the \texttt{id} attribute in \texttt{\(<\)input id="flight-from"\(>\)} indicates that the input field corresponds to the flight departure location (example taken from the MiniWoB++ benchmark \citep{shi_world_2017}).
    \end{description}
    \item[Disadvantages] of textual screen representations are: \hfill
    \begin{description}
        \item[Reduced information density:] 
            Some text formats, particularly raw HTML, can introduce verbosity that reduces the overall information density.
        \item[Structural inconsistency:] 
            Visually similar content can be rendered using different underlying structures. For example, a button might be implemented with either a \texttt{<button>} or a \texttt{<span>} tag. Similarly, visually similar components can have vastly different underlying code due to different implementation choices, such as the selected styling framework (e.g., Bootstrap\footnote{\url{https://getbootstrap.com/}} vs. Tailwind CSS\footnote{\url{https://tailwindcss.com/}}) and HTML-generating framework (e.g., Angular\footnote{\url{https://angular.dev/}} vs. React\footnote{\url{https://react.dev/}}).
        \item[Omission of visual information:] 
            Textual representations often lack information about spatial relationships and positioning that can be critical in understanding the screen's layout.
        \item[Lack of textual representation:] 
            Some screen components, such as embedded plugins, may not have an alternative textual screen representation. Certain applications may entirely lack any alternative textual screen representation.
    \end{description}
\end{description}

Some of these disadvantages can be mitigated through engineering solutions. For instance, the absence of visual positioning can be addressed by incorporating absolute or relative screen coordinates into the textual screen representation \citepeg{shi_world_2017, liu_reinforcement_2018}, or by embedding elements with information from nearby neighboring elements \citep{liu_reinforcement_2018}. 
Additionally, the verbosity inherent in raw text can be reduced by simplifying the observations $o_t \rightarrow o_t^*$.
However, these mitigation strategies usually do not fully overcome the inherent limitations observed in practice.

\section{Code Generation Example} \label{sec:code_generation_examples}

\begin{figure}[htbp]
    \centering
    \includegraphics[width=0.6\linewidth]{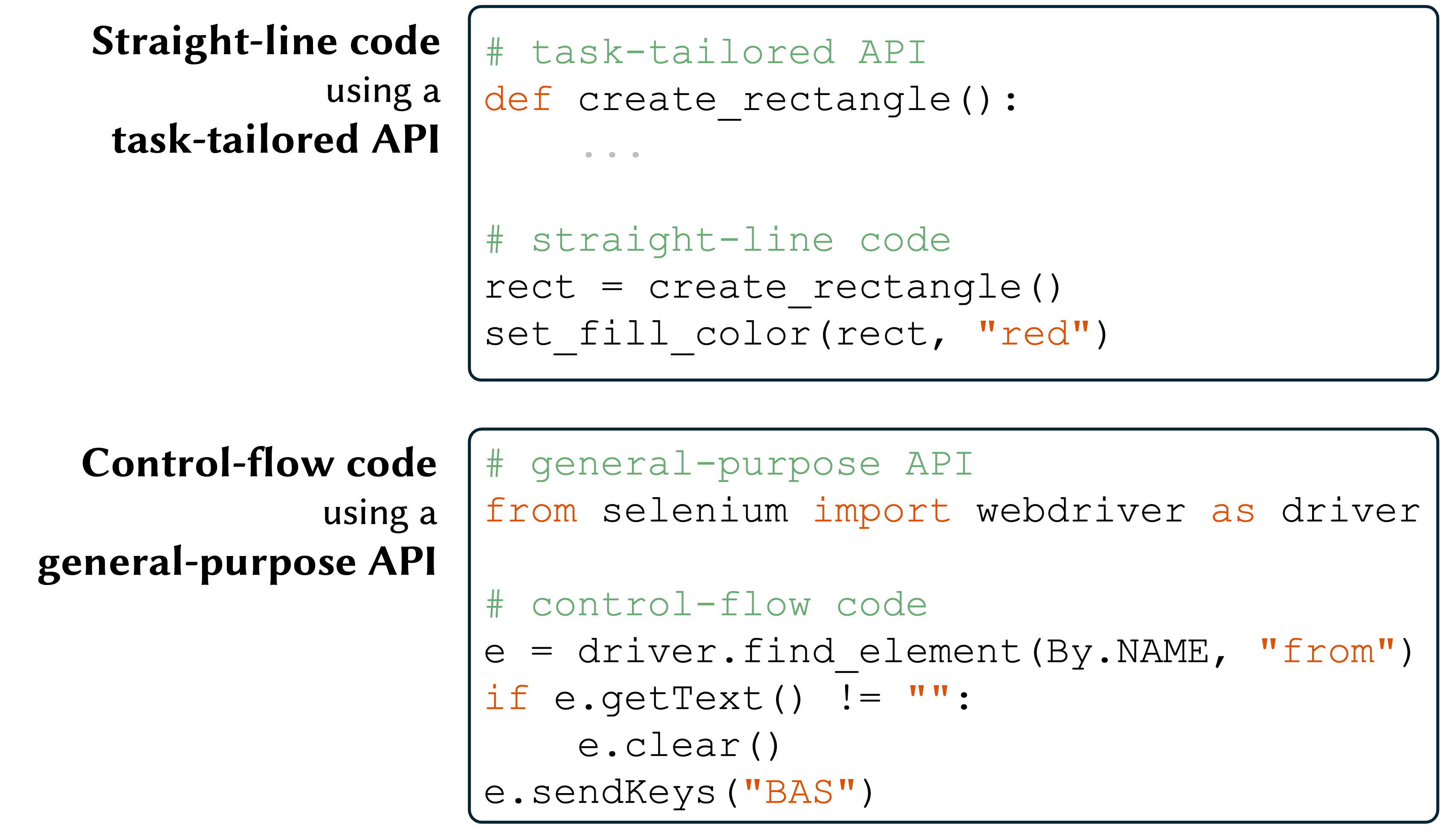}
    \caption{Two examples of executable Python code actions. Two different structures (straight-line or with control flow) and API abstraction levels (task-tailored or general-purpose) are shown.}
    \Description{
    The image compares two coding paradigms: straight-line code using a task-tailored API, and control-flow code using a general-purpose API.
    The top half shows straight-line code for a graphics task using a task-specific API. The function \texttt{create\_rectangle} is defined and then invoked, followed by setting the fill color to red using \texttt{set\_fill\_color}. The code has a custom function and is sequential.
    The bottom half shows control-flow code using Selenium, a general-purpose web automation library. It imports the webdriver, finds an element by its name attribute, checks if the element contains text, clears it if so, and then sends keys ``BAS'' to the element. The code includes an explicit if-statement and imports a library function, demonstrating more complex logic.
    }
    \label{fig:action-space-code}
\end{figure}

As illustrated in Figure~\ref{fig:action-space-code}, executable code actions generated by agents can vary significantly in both structural complexity and abstraction level. Specifically, we distinguish between straight-line execution versus control-flow logic, and the use of task-tailored APIs versus general-purpose APIs. 

\section{The Nature of Computer Environments}\label{sec:domain_view:nature}

\begin{table*}
\begin{threeparttable}
    \caption{Properties of common computer environments, assembled from \citet[Chapter~2]{russell_artificial_2022} and \citet[Chapter~2.3]{sutton_reinforcement_2018}. The middle and right columns compare common assumptions in research to actual computer environments.}
    \label{table:environment-property}

    \begin{tabular}{>{\raggedright\arraybackslash}p{4.3cm} >{\raggedright\arraybackslash}p{4cm} >{\raggedright\arraybackslash}p{4cm}}
    \toprule
    \textbf{Property}   & \textbf{Research computer \newline environment}   & \textbf{Actual computer \newline environment}     \\ 
    \midrule
    Observability                & Partially observable                                  & Partially observable      \\ 
    Number of agents             & Single-agent                                          & Single-agent\tnote{a}     \\ 
    Determinism                  & Deterministic                                         & Primarily deterministic\tnote{b}   \\ 
    Episodicity                 & Episodic                                              & Sequential                \\ 
    Dynamism                     & Static\tnote{c}                                       & Dynamic                   \\ 
    Stationarity                & Stationary                                            & Non-stationary            \\ 
    Environment knowledge       & Initially unknown                                    & Initially unknown          \\ 
    \bottomrule
    \end{tabular}
    \begin{tablenotes}
        \footnotesize
        \item[a] Assuming the user hands control to the agent and does not intervene.
        \item[b] Computer use is primarily deterministic due to user-friendly design principles, but can be stochastic.
        \item[c] \citet{toyama_androidenv_2021} is an exception providing a dynamic Android environment.
    \end{tablenotes}
\end{threeparttable}
\end{table*}

In \cref{table:environment-property}, we classify computer environments according to the framework established by \citet[Chapter~2.3]{russell_artificial_2022}.
We distinguish between the computer environment typically found in research (middle column) and the actual computer environment in a productive setting (right column).

Determinism is often assumed, meaning that for a state $s_t$ and action $a_t$, only one possible outcome $s_{t+1}$ exists. 
While this holds for many interface-driven tasks, real environments may contain stochastic elements, such as randomized content (e.g., shuffle button in a music app) or latency effects, that introduce variability.
The assumption of episodicity simplifies credit assignment, but computer environments are inherently sequential. States may depend on long-term history across sessions, requiring agents to model extended temporal dependencies beyond the task-specific trajectories.
Research environments are often considered static, where only agent actions cause changes. In contrast, real environments are dynamic—background processes, user actions, or updates can alter state independently, requiring robustness to asynchronous events \citep{humble_continuous_2011}.
Stationarity, another common assumption, implies stable dynamics over time. Yet actual environments are non-stationary due to software updates, configuration changes, or shifting data, which challenges long-term generalization.
Lastly, computer environments are typically assumed to have unknown dynamics, meaning an agent does not initially know the effect of an action. While technically true, some agents leverage pre-training to learn conventions and begin with anticipatory knowledge (see \cref{sec:learning_strategy_pretraining}). For example, they might learn that clicking a 'submit' button typically submits a form.

\section{Challenges for Deployment and Application} \label{sec:considerations_production}

Current research in agents for computer use focuses on enhancing their autonomous capabilities across various domains and benchmarks. However, deploying these agents in production introduces several additional challenges. 

\subsection{Technical Challenges and Considerations} \label{sec:cp-technical-challenges}

A production setting entails a specific environment, such as a business application, that the agent must be able to control.
However, effectively adapting an agent to a production environment remains an open research question.
Besides efficient environment learning, a production setting holds additional challenges, including diverse user hardware. For instance, ACUs must scope with different screen resolutions, multi-monitor setups, as well as different device configurations, including a wide range of Android distributions, home screen setups, or color schemes \citep{lee_benchmarking_2024}.
Additionally, a production environment is \emph{non-stationary} as applications undergo continuous enhancement \citep{humble_continuous_2011}, changing their interfaces and behavior.
A production-ready agent must be able to handle those ever-changing circumstances, either autonomously or through continuous updates implemented by its developers.

\paragraph{Speed, Cost, and Availability}
While current research primarily focuses on an agent's autonomous capabilities, practical deployment demands careful consideration of \emph{prediction speed}, \emph{operational costs}, and \emph{availability}.
Faster prediction time leads to less latency and a better user experience.
Costs can be monetary through API calls to third-party foundation models or hardware considerations for local agents.
In terms of potential monetary costs, solving a single task costs roughly \$$0.28$ when assuming to use a state-of-the-art foundation model, processing $765$ image tokens (high-resolution screenshot), $600$ text tokens (agent prompt and user instruction), $1000$ text output tokens (reasoning and action prediction), and $7$ actions per task \citep[as in][]{deng_mind2web_2023} and current API pricing (December 2024).
Furthermore, reliance on external resources introduces dependencies that can impact availability, such as requiring a stable internet connection and the reliable operation of third-party services.

\paragraph{Privacy}
While LLMs can run on local machines \citep{tuggener_so_2024}, many state-of-the-art models such as GPT-4V \citet{openai_gpt-4_2024} are only available through an API.
Agents relying on external resources, such as proprietary foundation models, introduce privacy concerns.
Individuals and companies may be reluctant to send screenshots of their applications, which may show sensitive data, to an external server streamed over the internet.
This raises similar data privacy challenges observed in other foundation model applications \citep{neel_privacy_2024}.
However, a crucial difference emerges with agents: traditional user education on data-sharing practices becomes insufficient, as users cannot fully control an agent's access to information when it operates autonomously on their devices.
For example, an agent in financial reporting might inadvertently open, observe, and thus transmit sensitive financial documents without the user's explicit consent and in contradiction to contractual or legal requirements.

\subsection{Safety Considerations} \label{sec:safety}

Despite advances in autonomous agent development, current systems often lack the reliability and comprehensiveness required for safe real-world deployment.
The consequences of an agent's unintentional, erroneous actions can differ depending on the domain, ranging from minor disruptions, such as playing the wrong music video, to more severe issues, like the unauthorized disclosure of confidential medical records.
For production, the risk of erroneous actions must be balanced with the agent's capabilities and the benefits of automation.
This balance can be achieved by adjusting design parameters: The agent's \emph{level of autonomy} and the \emph{scope of its deployment}.


\paragraph{Reducing Automation}

Most ACU research is about \emph{full automation}, meaning the agent is in control, and it is assumed no human is in the loop.
To decrease the risk of erroneous actions, agents can operate in \emph{conditional automation}, meaning the agent is in control, but it can hand back control to the user for critical actions.
For example, \citet{li_appagent_2024} let their agent determine critical actions, such as validating payments. However, this approach still risks the agent overlooking critical actions, which can be avoided in use cases like payment by requiring external validation through a separate payment processing system inaccessible to the agent.
In contrast, \citet{wang_enabling_2023} also allows agent-initiated conversations, allowing them to solicit information.
A further restriction would be running the agent in \emph{partial automation}, meaning the human is in control and hands it to an agent only to fulfill a straightforward sub-task.
For example, web browsers providing auto-fill functions for typical web forms can be considered partial, non-instruction-based agents for computer use.
An even further automation restriction is agents only \emph{assisting} users, meaning the human stays in control the whole time while the agent provides only suggestions.
This design is typical for non-instruction-based agents for computer use like GitHub CoPilot\footnote{\url{https://github.com/features/copilot}} or Grammarly\footnote{\url{https://grammarly.com/}}.

\paragraph{Managing the Scope of the Production Environment} 
To decrease the risk of erroneous actions, the scope of the production environment can be constrained.
For a given use case, the action space $\mathcal{A}$ can be restricted by removing high-risk actions, such as disabling critical deletion operations. This can be achieved, for instance, by limiting the agent's file system permissions. Additionally, safety checks can be implemented to autonomously verify the feasibility and safety of actions prior to execution, effectively providing guardrails for the agent \citep{liang_taskmatrixai_2023}.
Similarly, the state space $\mathcal{S}$ can be reduced to simplify the operational environment. For example, a web agent’s access could be restricted to a predefined set of curated websites instead of granting access to the entire web. In the context of personal computers, the operational domain could be narrowed to specific applications, such as those within an office productivity suite.
These constraints not only limit the agent’s potential behaviors but also simplify environment learning and enable more accurate assessments of the agent's capabilities.

\subsection{Adapting Generally Capable Agents}

Leading AI companies, such as Anthropic, have begun advancing into the realm of ACUs, offering generally capable, out-of-the-box solutions \citep{hu_dawn_2024}. However, we anticipate that truly \emph{general} autonomous instruction-based ACUs -- defined as those with capabilities, resilience, and safety comparable to highly skilled human computer users across most domains -- are unlikely to emerge in the next two years, given the current state-of-the-art, for example, the unavailability of massive and challenging training data.

This projection highlights a critical research question: \textit{How can generally capable agents be effectively adapted to address specific organizational use cases?} For example, enabling an agent to autonomously, safely, and reliably control a unique business application currently requires comprehensive customization. It involves tailoring pre-trained, capable agents to meet the precise needs of a given use case, thereby warranting extensive on-task training experience.

For pure text-based agents, the parallel challenge of adopting a generalist model to organizational needs and know-how is currently approached using retrieval-augmented generation (RAG) strategies, where foundation models are equipped with use-case-specific knowledge by grounding them in internal documents \citep{lewis2020retrieval}. Similarly, the focus in adapting ACUs would lie in achieving robust, organization-specific adaptation starting from a general-purpose, pre-trained agent -- yet a similar process or framework has yet to be developed.

\section{Structured Overview of Existing Work}\label{sec:appendix_tables}

For this review, we identified $\numagents$ ACUs and $\numdatasets$ datasets and categorized them according to the introduced taxonomy. Here, we present a detailed list of the identified literature and their classification.
A more detailed version of the tables presented in this section are available on our project page at \url{https://sagerpascal.github.io/agents-for-computer-use}.

\definecolor{observation_space_color_bg_l}{RGB}{230,241,252} 

\subsection{Environment and Interaction Perspective}
\label{appendix_A1}

{\footnotesize
\begin{longtable}{%
    m{6.5cm}
    >{\centering\arraybackslash\columncolor{domain_color_bg}}m{1.8cm}
    >{\columncolor{observation_space_color_bg}}m{0.3cm}
    >{\columncolor{observation_space_color_bg}}m{0.3cm}
    >{\columncolor{observation_space_color_bg_l}}m{0.3cm}
    >{\columncolor{observation_space_color_bg_l}}m{0.3cm}
    >{\columncolor{observation_space_color_bg_l}}m{0.3cm}
    >{\columncolor{observation_space_color_bg}}m{0.3cm}
    >{\columncolor{observation_space_color_bg}}m{0.3cm}
    >{\columncolor{action_space_color_bg}}m{0.3cm}
    >{\columncolor{action_space_color_bg}}m{0.3cm}
    >{\columncolor{action_space_color_bg}}m{0.3cm}
    >{\columncolor{action_space_color_bg}}m{0.3cm}
}

\caption{
    Literature overview: Domain and interaction types. \T\ indicates full presence of an aspect; \halfTT\ indicates partial presence; empty means absence.
}
\label{tab:interaction_literature}
\\

\toprule
\textbf{Paper} & \textbf{Domain} & \multicolumn{7}{>{\columncolor{observation_space_color_bg}}c}{\textbf{Observation Space}} & \multicolumn{4}{>{\columncolor{action_space_color_bg}}c}{\textbf{Action Space}} \\
\midrule
 &  & \rotatebox{90}{\textbf{Image}} & \rotatebox{90}{\textbf{Image to Textual}} & \rotatebox{90}{\textbf{HTML}} & \rotatebox{90}{\textbf{Android View Hierarchy\ }} & \rotatebox{90}{\textbf{UI Automation Tree}} & \rotatebox{90}{\textbf{Accessibility Tree}} & \rotatebox{90}{\textbf{Indirect}} & \rotatebox{90}{\textbf{Mouse Keyboard}} & \rotatebox{90}{\textbf{Direct UI Access}} & \rotatebox{90}{\textbf{Tailored}} & \rotatebox{90}{\textbf{Executable Code}} \\
\midrule
\endfirsthead

\multicolumn{13}{c}
{{\bfseries \tablename\ \thetable{} -- continued from previous page}} \\
\toprule
\textbf{Paper} & \textbf{Domain} & \multicolumn{7}{>{\columncolor{observation_space_color_bg}}c}{\textbf{Observation Space}} & \multicolumn{4}{>{\columncolor{action_space_color_bg}}c}{\textbf{Action Space}} \\
\midrule
 &  & \rotatebox{90}{\textbf{Image}} & \rotatebox{90}{\textbf{Image to Textual}} & \rotatebox{90}{\textbf{HTML}} & \rotatebox{90}{\textbf{Android View Hierarchy\ }} & \rotatebox{90}{\textbf{UI Automation Tree}} & \rotatebox{90}{\textbf{Accessibility Tree}} & \rotatebox{90}{\textbf{Indirect}} & \rotatebox{90}{\textbf{Mouse Keyboard}} & \rotatebox{90}{\textbf{Direct UI Access}} & \rotatebox{90}{\textbf{Tailored}} & \rotatebox{90}{\textbf{Executable Code}} \\
\midrule
\endhead

\endfoot

\bottomrule
\endlastfoot
\citet{shaw_pixels_2023,niu_screenagent_2024,he_webvoyager_2024} & Web & \T &  &  &  &  &  &  & \T &  &  &  \\
\midrule
\citet{pan_autonomous_2024,koh_tree_2024} & Web & \T &  &  &  &  &  &  &  & \T &  &  \\
\midrule
\citet{iki_berts_2022} & Web &  & \T &  &  &  &  &  & \T &  &  &  \\
\midrule
\citet{lo_hierarchical_2023,fereidouni_search_2024,guan_intelligent_2023} & Web &  & \T &  &  &  &  &  &  & \T &  &  \\
\midrule
\citet{cho_caap_2024} & Web &  & \T &  &  &  &  &  & \T & \T &  &  \\
\midrule
\citet{kim_language_2023,li_zero-shot_2023,liu_reinforcement_2018,deng_multi-turn_2024,sodhi_heap_2023,gur_learning_2019,ma_laser_2024,gur_environment_2021,jia_dom-q-net_2019,zheng_synapse_2024,li_glider_2021,murty_bagel_2024,deng_mind2web_2023,gur_understanding_2023,lutz_wilbur_2024,lai_autowebglm_2024} & Web &  &  & \T &  &  &  &  &  & \T &  &  \\
\midrule
\citet{putta_agent_2024,xu_grounding_2021} & Web &  &  & \T &  &  &  &  &  &  & \T &  \\
\midrule
\citet{furuta_exposing_2023,sun_adaplanner_2023,tao_webwise_2023,gur_real-world_2024} & Web &  &  & \T &  &  &  &  &  &  &  & \T \\
\midrule
\citet{nakano_webgpt_2022} & Web &  &  & \T &  &  &  &  &  & \T & \T &  \\
\midrule
\citet{zaheer_learning_2022} & Web &  &  & \halfT &  &  &  &  &  & \T &  &  \\
\midrule
\citet{zhou_webarena_2024} & Web &  &  &  &  &  & \T &  &  & \T &  &  \\
\midrule
\citet{zhang_webpilot_2024} & Web &  &  &  &  &  & \T &  &  & \T &  & \T \\
\midrule
\citet{humphreys_data-driven_2022,lin_automating_2021,shi_world_2017} & Web & \T &  & \T &  &  &  &  & \T &  &  &  \\
\midrule
\citet{furuta_multimodal_2024,mazumder_flin_2021,lu_weblinx_2024,kil_dual-view_2024,zheng_gpt-4vision_2024} & Web & \T &  & \T &  &  &  &  &  & \T &  &  \\
\midrule
\citet{chae_web_2024} & Web &  &  & \T &  &  & \T &  &  & \T &  &  \\
\midrule
\citet{wang_mobile-agent_2024,zhang_you_2024,zhang_android_2024,lu_gui_2024} & Android & \T &  &  &  &  &  &  & \T &  &  &  \\
\midrule
\citet{wen_autodroid_2024,sun_meta-gui_2022,wu_mobilevlm_2024,ding_mobileagent_2024,li_interactive_2020,nong_mobileflow_2024} & Android & \T &  &  &  &  &  &  &  & \T &  &  \\
\midrule
\citet{dorka_training_2024} & Android & \T &  &  &  &  &  &  & \T & \T &  &  \\
\midrule
\citet{abukadah_mapping_2024,song_navigating_2023,song_visiontasker_2024,li_learning_2021,ma_coco-agent_2024} & Android &  & \T &  &  &  &  &  &  & \T &  &  \\
\midrule
\citet{rawles_android_2023} & Android &  & \T &  &  &  &  &  & \T & \T &  &  \\
\midrule
\citet{wen_droidbot-gpt_2024,li_mapping_2020} & Android &  &  &  & \T &  &  &  &  & \T &  &  \\
\midrule
\citet{bishop_latent_2024,li_effects_2024} & Android &  &  &  &  &  & \T &  & \T &  &  &  \\
\midrule
\citet{li_uinav_2024,lee_explore_2023} & Android &  &  &  &  &  & \T &  &  & \T &  &  \\
\midrule
\citet{zhang_appagent_2023,li_appagent_2024} & Android & \T &  &  & \T &  &  &  &  & \T &  &  \\
\midrule
\citet{wang_enabling_2023} & Android &  &  & \halfT & \T &  &  &  &  & \T &  &  \\
\midrule
\citet{deng_mobile-bench_2024} & Android &  &  & \halfT & \T &  &  &  &  & \T &  & \T \\
\midrule
\citet{cheng_seeclick_2024,hong_cogagent_2024} & Web, Android & \T &  &  &  &  &  &  & \T &  &  &  \\
\midrule
\citet{lu_omniparser_2024} & Web, Android &  & \T &  &  &  &  &  &  & \T &  &  \\
\midrule
\citet{rahman_v-zen_2024} & PC & \T &  &  &  &  &  &  & \T &  &  &  \\
\midrule
\citet{gao_assistgui_2024} & PC & \T &  &  &  &  &  &  & \T &  &  & \T \\
\midrule
\citet{song_mmac-copilot_2024} & PC & \T &  &  &  &  &  &  &  & \T &  & \T \\
\midrule
\citet{wang_officebench_2024} & PC &  &  &  &  &  &  & \T &  &  & \T &  \\
\midrule
\citet{wu_os-copilot_2024,guo_pptc_2024} & PC &  &  &  &  &  &  & \T &  &  &  & \T \\
\midrule
\citet{zhang_ufo_2024} & PC & \T &  &  &  & \T &  &  &  & \T & \T &  \\
\midrule
\citet{bonatti_windows_2024} & Web, PC & \T &  & \T &  & \T &  &  &  & \T & \T & \T \\
\midrule
\citet{yan_gpt-4v_2023} & Android, iOS & \T &  &  &  &  &  &  &  & \T &  &  \\
\midrule
\citet{song_restgpt_2023} & API &  &  &  &  &  &  & \T &  &  &  & \T \\
\end{longtable} 
}

\newpage
\subsection{Agent Perspective}
\label{appendix_A2}
{\footnotesize
\begin{longtable}{ m{7cm} >{\centering\arraybackslash\columncolor{agent_color_bg}}m{0.25cm} >{\centering\arraybackslash\columncolor{agent_color_bg}}m{0.25cm} >{\centering\arraybackslash\columncolor{policy_color_bg}}m{0.25cm} >{\centering\arraybackslash\columncolor{policy_color_bg}}m{0.25cm} >{\centering\arraybackslash\columncolor{policy_color_bg}}m{0.25cm} >{\centering\arraybackslash\columncolor{learning_strategy_color_General_Pre_training_bg}}m{0.25cm} >{\centering\arraybackslash\columncolor{learning_strategy_color_General_Pre_training_bg}}m{0.25cm} >{\centering\arraybackslash\columncolor{learning_strategy_color_Environment_Adaption_bg}}m{0.25cm} >{\centering\arraybackslash\columncolor{learning_strategy_color_Environment_Adaption_bg}}m{0.25cm} >{\centering\arraybackslash\columncolor{learning_strategy_color_Environment_Adaption_bg}}m{0.25cm} >{\centering\arraybackslash\columncolor{learning_strategy_color_Episodic_improvement_bg}}m{0.25cm} >{\centering\arraybackslash\columncolor{learning_strategy_color_Episodic_improvement_bg}}m{0.25cm} >{\centering\arraybackslash\columncolor{learning_strategy_color_Episodic_improvement_bg}}m{0.25cm} >{\centering\arraybackslash\columncolor{agent_color_Reinforce_Agent_bg}}m{0.25cm} }
\caption{
Literature overview: Core agent design principles. PT = general pre-training; EL = environment learning; EI = episodic improvement;
BC = behavioral cloning; RL = reinforcement learning; LTM = long-term memory;
\T\ indicates the full presence of an aspect; \halfTT\ indicates the presence of an aspect with variations; empty means the aspect is absent.
} \label{tab:agent_literature}
\\
\toprule
\textbf{Paper} & \multicolumn{2}{>{\centering\arraybackslash\columncolor{agent_color_bg}}c }{\textbf{Type}} & \multicolumn{3}{>{\centering\arraybackslash\columncolor{policy_color_bg}}c }{\textbf{Policy}} & \multicolumn{2}{>{\centering\arraybackslash\columncolor{learning_strategy_color_General_Pre_training_bg}}c }{\textbf{PT}} & \multicolumn{3}{>{\centering\arraybackslash\columncolor{learning_strategy_color_Environment_Adaption_bg}}c }{\textbf{EL}} & \multicolumn{4}{>{\centering\arraybackslash\columncolor{learning_strategy_color_Episodic_improvement_bg}}c }{\textbf{EI}} \\
\midrule
& \rotatebox{90}{\textbf{Foundation agent}} & \rotatebox{90}{\textbf{Specialized agent}} & \rotatebox{90}{\textbf{Memoryless}} & \rotatebox{90}{\textbf{History-based}} & \rotatebox{90}{\textbf{State-based}} & \rotatebox{90}{\textbf{Foundation model\ }} & \rotatebox{90}{\textbf{Backbone}} & \rotatebox{90}{\textbf{BC}} & \rotatebox{90}{\textbf{RL}} & \rotatebox{90}{\textbf{LTM}} & \rotatebox{90}{\textbf{Instruction tuning\ }} & \rotatebox{90}{\textbf{Few-shot}} & \rotatebox{90}{\textbf{Planning}} \\
\midrule
\endfirsthead

\multicolumn{11}{c}
{{\bfseries \tablename\ \thetable{} -- continued from previous page}} \\
\toprule
\textbf{Paper} & \multicolumn{2}{>{\centering\arraybackslash\columncolor{agent_color_bg}}c }{\textbf{Type}} & \multicolumn{3}{>{\centering\arraybackslash\columncolor{policy_color_bg}}c }{\textbf{Policy}} & \multicolumn{2}{>{\centering\arraybackslash\columncolor{learning_strategy_color_General_Pre_training_bg}}c }{\textbf{PT}} & \multicolumn{3}{>{\centering\arraybackslash\columncolor{learning_strategy_color_Environment_Adaption_bg}}c }{\textbf{EL}} & \multicolumn{4}{>{\centering\arraybackslash\columncolor{learning_strategy_color_Episodic_improvement_bg}}c }{\textbf{EI}} \\
\midrule
& \rotatebox{90}{\textbf{Foundation agent}} & \rotatebox{90}{\textbf{Specialized agent}} & \rotatebox{90}{\textbf{Memoryless}} & \rotatebox{90}{\textbf{History-based}} & \rotatebox{90}{\textbf{State-based}} & \rotatebox{90}{\textbf{Foundation model\ }} & \rotatebox{90}{\textbf{Backbone}} & \rotatebox{90}{\textbf{BC}} & \rotatebox{90}{\textbf{RL}} & \rotatebox{90}{\textbf{LTM}} & \rotatebox{90}{\textbf{Instruction tuning\ }} & \rotatebox{90}{\textbf{Demonstrations}} & \rotatebox{90}{\textbf{Planning}} \\
\midrule
\endhead

\endfoot

\bottomrule 
\endlastfoot
\citet{wang_enabling_2023} & \T &  & \T &  &  & \T &  &  &  &  & \T & \T &  \\
\midrule
\citet{niu_screenagent_2024} & \T &  & \T &  &  & \T &  &  &  &  & \T &  & \T \\
\midrule
\citet{ding_mobileagent_2024} & \T &  & \T &  &  & \T &  &  &  &  & \T &  &  \\
\midrule
\citet{sun_adaplanner_2023,lee_explore_2023} & \T &  & \T &  &  & \T &  &  &  & \T & \T & \T & \T \\
\midrule
\citet{tao_webwise_2023} & \T &  & \T &  &  & \T &  &  &  & \T & \T & \T &  \\
\midrule
\citet{wu_os-copilot_2024} & \T &  & \T &  &  & \T &  &  &  & \T & \T &  &  \\
\midrule
\citet{nong_mobileflow_2024} & \T &  & \T &  &  & \T & \T &  &  &  & \T &  & \T \\
\midrule
\citet{kim_language_2023,zhang_webpilot_2024,zhou_webarena_2024,sodhi_heap_2023,cho_caap_2024,koh_tree_2024,deng_mobile-bench_2024,tan_towards_2024} & \T &  &  & \T &  & \T &  &  &  &  & \T & \T & \T \\
\midrule
\citet{zheng_synapse_2024,bishop_latent_2024} & \T &  &  & \T &  & \T &  &  &  &  & \T & \T &  \\
\midrule
\citet{chae_web_2024,song_restgpt_2023} & \T &  &  & \T &  & \T &  &  &  &  & \T &  & \T \\
\midrule
\citet{li_zero-shot_2023,ma_laser_2024,zheng_gpt-4vision_2024,wang_mobile-agent_2024,wen_droidbot-gpt_2024,cheng_seeclick_2024,wang_officebench_2024,guo_pptc_2024} & \T &  &  & \T &  & \T &  &  &  &  & \T &  &  \\ \midrule
\citet{murty_bagel_2024,deng_mind2web_2023,lu_weblinx_2024,zhang_android_2024,li_effects_2024} & \T &  &  & \T &  & \T &  & \T &  &  & \T & \T &  \\
\midrule
\citet{lai_autowebglm_2024,ma_coco-agent_2024} & \T &  &  & \T &  & \T &  & \T &  &  & \T &  &  \\
\midrule
\citet{deng_multi-turn_2024,lutz_wilbur_2024,wen_autodroid_2024,li_appagent_2024} & \T &  &  & \T &  & \T &  &  &  & \T & \T & \T &  \\
\midrule
\citet{gao_assistgui_2024} & \T &  &  & \T &  & \T & \T &  &  &  & \T & \T & \T \\
\midrule
\citet{furuta_exposing_2023,gur_real-world_2024} & \T &  &  & \T &  & \T & \T &  &  &  & \T & \T &  \\
\midrule
\citet{guan_intelligent_2023} & \T &  &  & \T &  & \T & \T &  &  &  & \T &  & \T \\
\midrule
\citet{song_navigating_2023,lu_omniparser_2024} & \T &  &  & \T &  & \T & \T &  &  &  & \T &  &  \\
\midrule
\citet{rawles_android_2023} & \T &  &  & \T &  & \T & \T & \T &  &  & \T & \T & \T \\
\midrule
\citet{song_visiontasker_2024} & \T &  &  & \T &  & \T & \T &  &  & \T & \T & \T &  \\
\midrule
\citet{pan_autonomous_2024} & \T &  &  &  & \T & \T &  & \T &  &  & \T &  &  \\
\midrule
\citet{zhang_appagent_2023} & \T &  &  &  & \T & \T &  &  &  & \T & \T & \T &  \\
\midrule
\citet{zhang_ufo_2024} & \T &  &  & \T & \T & \T &  &  &  &  & \T & \T & \T \\
\midrule
\citet{bonatti_windows_2024} & \T &  &  & \T & \T & \T &  &  &  &  & \T &  &  \\
\midrule
\citet{xu_grounding_2021} & \halfT &  & \T &  &  & \T &  &  &  &  &  &  &  \\
\midrule
\citet{song_mmac-copilot_2024} & \halfT &  & \T &  &  & \T &  & \T &  &  &  &  & \T \\
\midrule
\citet{abukadah_mapping_2024} & \halfT &  & \T &  &  & \T & \T & \T &  &  &  &  &  \\
\midrule
\citet{zhang_you_2024} & \halfT &  &  & \T &  & \T &  & \T &  &  &  &  & \T \\
\midrule
\citet{gur_understanding_2023,he_webvoyager_2024,wu_mobilevlm_2024,lu_gui_2024,hong_cogagent_2024,rahman_v-zen_2024} & \halfT &  &  & \T &  & \T &  & \T &  &  &  &  &  \\
\midrule
\citet{putta_agent_2024} & \halfT &  &  & \T &  & \T &  &  & \T &  &  &  & \T \\
\midrule
\citet{lo_hierarchical_2023} & \halfT &  &  & \T &  & \T &  &  &  & \T &  &  &  \\
\midrule
\citet{nakano_webgpt_2022,fereidouni_search_2024} & \halfT &  &  & \T &  & \T &  & \T & \T &  &  &  &  \\
\midrule
\citet{furuta_multimodal_2024,kil_dual-view_2024,dorka_training_2024} & \halfT &  &  & \T &  & \T & \T & \T &  &  &  &  &  \\
\midrule
\citet{liu_reinforcement_2018} &  & \T & \T &  &  &  &  &  &  &  &  &  &  \\
\midrule
\citet{zaheer_learning_2022,li_interactive_2020,li_mapping_2020} &  & \T & \T &  &  &  &  & \T &  &  &  &  &  \\
\midrule
\citet{gur_learning_2019,gur_environment_2021,jia_dom-q-net_2019,li_glider_2021} &  & \T & \T &  &  &  &  &  & \T &  &  &  &  \\
\midrule
\citet{shi_world_2017,li_learning_2021} &  & \T & \T &  &  &  &  & \T & \T &  &  &  &  \\
\midrule
\citet{mazumder_flin_2021} &  & \T & \T &  &  &  & \T &  &  &  &  &  &  \\
\midrule
\citet{li_uinav_2024} &  & \T & \T &  &  &  & \T & \T &  &  &  &  &  \\
\midrule
\citet{shaw_pixels_2023} &  & \T & \T &  &  &  & \T & \T & \T &  &  &  &  \\
\midrule
\citet{lin_automating_2021,sun_meta-gui_2022} &  & \T &  & \T &  &  &  & \T &  &  &  &  &  \\
\midrule
\citet{yan_gpt-4v_2023} &  & \T &  & \T &  &  &  & \T &  & \T &  &  &  \\
\midrule
\citet{humphreys_data-driven_2022} &  & \T &  &  & \T &  &  & \T & \T &  &  &  &  \\
\midrule
\citet{iki_berts_2022} &  & \T &  & \T & \T &  & \T & \T &  &  &  &  &  \\
\end{longtable} 
}

\newpage
\subsection{Datasets}
\label{appendix_A3}

{\footnotesize
\begin{longtable}{ p{4cm} >{\centering\arraybackslash\columncolor{domain_color_bg}}p{2cm} c c >{\centering\arraybackslash\columncolor{observation_space_color_bg}}c >{\centering\arraybackslash\columncolor{observation_space_color_bg}}c >{\centering\arraybackslash\columncolor{action_space_color_bg}}c >{\centering\arraybackslash\columncolor{action_space_color_bg}}c >{\centering\arraybackslash\columncolor{action_space_color_bg}}c >{\centering\arraybackslash\columncolor{action_space_color_bg}}c }
\caption{Literature overview: Datasets. OS = observation space; AS = action space. \T\ indicates the presence of an aspect; empty means the aspect is absent.} \label{tab:overview_datasets} \\
\toprule 
\textbf{Paper} & \textbf{Domain} & \multicolumn{2}{c }{\textbf{Type}} & \multicolumn{2}{>{\centering\arraybackslash\columncolor{observation_space_color_bg}}c }{\textbf{OS}} & \multicolumn{4}{>{\centering\arraybackslash\columncolor{action_space_color_bg}}c }{\textbf{AS}} \\
\midrule
 &  & \rotatebox{90}{\textbf{Controlled Environment\ }} & \rotatebox{90}{\textbf{Offline Dataset}} & \rotatebox{90}{\textbf{Image}} & \rotatebox{90}{\textbf{Textual}} & \rotatebox{90}{\textbf{Mouse}} & \rotatebox{90}{\textbf{Direct}} & \rotatebox{90}{\textbf{Tailored}} & \rotatebox{90}{\textbf{Code}} \\
\midrule
\endfirsthead

\multicolumn{10}{c}
{{\bfseries \tablename\ \thetable{} -- continued from previous page}} \\
\toprule 
\textbf{Paper} & \textbf{Domain} & \multicolumn{2}{c }{\textbf{Type}} & \multicolumn{2}{>{\centering\arraybackslash\columncolor{observation_space_color_bg}}c }{\textbf{OS}} & \multicolumn{4}{>{\centering\arraybackslash\columncolor{action_space_color_bg}}c }{\textbf{AS}} \\
\midrule
 &  & \rotatebox{90}{\textbf{Controlled Environment\ }} & \rotatebox{90}{\textbf{Offline Dataset}} & \rotatebox{90}{\textbf{Image}} & \rotatebox{90}{\textbf{Textual}} & \rotatebox{90}{\textbf{Mouse}} & \rotatebox{90}{\textbf{Direct}} & \rotatebox{90}{\textbf{Tailored}} & \rotatebox{90}{\textbf{Code}} \\
\midrule
\endhead

\endfoot

\bottomrule 
\endlastfoot

\multicolumn{10}{ c }{\textbf{Established benchmarks}} \\
\midrule
MiniWoB \citep{shi_world_2017} & Web & \T &  & \T & \T & \T &  &  &  \\
MiniWoB++ \citep{liu_reinforcement_2018} & Web & \T &  &  & \T &  & \T &  &  \\
WebShop \citep{yao_webshop_2022} & Web & \T & \T &  & \T &  & \T & \T &  \\
Mind2Web \citep{deng_mind2web_2023} & Web &  & \T &  & \T & \T &  &  &  \\
WebArena \citep{zhou_webarena_2024} & Web &  & \T & \T & \T & \T &  &  &  \\
VisualWebArena \citep{koh_visualwebarena_2024} & Web &  & \T & \T & \T &  & \T &  &  \\
PixelHelp \citep{li_mapping_2020} & Android &  & \T & \T & \T & \T & \T &  &  \\
AndroidEnv \citep{toyama_androidenv_2021} & Android & \T &  &  & \T &  & \T &  &  \\
MoTIF \citep{avidan_dataset_2022} & Android &  & \T & \T & \T & \T &  &  &  \\
Android in the Wild \citep{rawles_android_2023} & Android &  & \T & \T &  & \T &  &  &  \\
AgentBench \citep{liu_agentbench_2023-1} & PC & \T &  &  &  &  &  & \T &  \\
OmniACT \citep{kapoor_omniact_2024} & PC &  & \T & \T & \T & \T &  &  &  \\
\midrule
\multicolumn{10}{ c }{\textbf{Other datasets}} \\
\midrule
RUSS \citep{xu_grounding_2021} & Web & \T & \T &  & \T &  & \T &  &  \\
gMiniWoB \citep{gur_environment_2021} & Web & \T &  & \T & \T &  & \T &  &  \\
WebVLN \citep{chen_webvln_2024} & Web &  & \T & \T & \T &  & \T &  &  \\
MT-Mind2Web \citep{deng_multi-turn_2024} & Web & \T &  &  & \T &  & \T &  &  \\
WorkArena \citep{drouin_workarena_2024} & Web &  & \T & \T & \T & \T & \T &  &  \\
AutoWebBench \citep{lai_autowebglm_2024} & Web &  & \T & \T & \T &  & \T &  &  \\
QBE-F-Droid \citep{koroglu_qbe_2018} & Android &  & \T &  & \T &  & \T &  &  \\
AppBuddy \citep{shvo_appbuddy_2021} & Android & \T &  &  & \T &  & \T &  &  \\
Meta-GUI \citep{sun_meta-gui_2022} & Android &  & \T & \T & \T &  & \T &  &  \\
UGIF \citep{venkatesh_ugif_2023} & Android &  & \T & \T & \T &  & \T &  &  \\
Mobile-Env \citep{zhang_mobile-env_2024} & Android &  & \T & \T & \T & \T &  &  &  \\
DroidTask \citep{wen_empowering_2023} & Android &  & \T & \T &  &  & \T &  &  \\
Android in the zoo \citep{zhang_android_2024} & Android &  & \T & \T &  &  & \T &  &  \\
GUIAct \citep{chen_guicourse_2024} & Android &  & \T & \T &  &  & \T &  &  \\
AssistGUI \citep{gao_assistgui_2024} & PC &  & \T & \T & \T & \T &  &  &  \\
ScreenAgent \citep{niu_screenagent_2024} & PC & \T &  & \T &  & \T &  &  &  \\
OSWorld \citep{xie_osworld_2024} & PC &  & \T & \T & \T & \T &  &  &  \\
AgentStudio \citep{zheng_agentstudio_2024} & PC &  & \T &  &  & \T &  &  & \T \\
PPTC \citep{guo_pptc_2024} & PC &  & \T &  & \T &  &  &  & \T \\
RestBench \citep{song_restgpt_2023} & API &  & \T &  &  &  &  &  & \T \\
GUI-World \citep{chen_gui-world_2024} & Multi &  & \T & \T &  & \T &  &  &  \\

\end{longtable} 
}

%
%
%
%
%
%
%
%

\end{document}